\documentclass[sigconf]{acmart}

\setcopyright{none}
\renewcommand\footnotetextcopyrightpermission[1]{}
\AtBeginDocument{%
  \providecommand\BibTeX{{%
    \normalfont B\kern-0.5em{\scshape i\kern-0.25em b}\kern-0.8em\TeX}}}

\usepackage{amsmath}

\usepackage{graphicx}
\usepackage{subcaption}
\usepackage[export]{adjustbox}
\usepackage[justification=justified]{caption}
\usepackage[nameinlink]{cleveref}
\usepackage{todonotes}
\usepackage[noend,linesnumbered,ruled]{algorithm2e}
\usepackage{microtype}
\usepackage{titlesec}
\usepackage{makecell}

\begin{document}

\title[Quality with Just Enough Diversity in Evolutionary Policy Search]{Quality with Just Enough Diversity\\in Evolutionary Policy Search}

\newcommand{\isae}{ISAE-SUPAERO, Univ. de Toulouse}

\author{Paul Templier}
\email{paul.templier@isae.fr}
\affiliation{%
  \institution{\isae}
  \city{Toulouse}
  \country{France}
}

\author{Luca Grillotti}
\email{luca.grillotti16@imperial.ac.uk}
\affiliation{%
  \institution{Imperial College London}
  \city{London}
  \country{United Kingdom}
}

\author{Emmanuel Rachelson}
\email{emmanuel.rachelson@isae.fr}
\affiliation{%
  \institution{\isae}
  \city{Toulouse}
  \country{France}
}

\author{Dennis G. Wilson}
\email{dennis.wilson@isae.fr}
\affiliation{%
  \institution{\isae}
  \city{Toulouse}
  \country{France}
}
\author{Antoine Cully}
\email{a.cully@imperial.ac.uk}
\affiliation{%
  \institution{Imperial College London}
  \city{London}
  \country{United Kingdom}
}

\renewcommand{\shortauthors}{Templier et al.}
\newcommand{\fulltitle}{Harry Potter and the Use of Structures in\\Evolution Strategies for Neural Policy Search}

\newcommand{\tocite}[1]{ \textcolor{red}{[#1]} }
\newcommand{\addlink}[2]{\href{#1}{#2}\footnote{\url{#1}}}
\def\code#1{\texttt{#1}}

\newcommand{\pt}[1]{\todo[backgroundcolor=blue]{PT}\textcolor{blue}{[\textbf{PT}: #1]}}
\newcommand{\er}[1]{\todo[backgroundcolor=red]{ER}\textcolor{red}{[\textbf{ER}: #1]}}
\newcommand{\dgw}[1]{\todo[backgroundcolor=green]{DGW}\textcolor{green}{[\textbf{DGW}: #1]}}

\newcommand{\addref}[1]{\todo[backgroundcolor=orange]{Ref}\textcolor{orange}{[\textbf{Add ref}: #1]}}

\newcommand{\censor}[1]{#1}

\newcommand{\neat}{NEAT}
\newcommand{\hyperneat}{HyperNEAT}
\newcommand{\es}{evolution strategies}
\newcommand{\snes}{SNES}
\newcommand{\xnes}{XNES}
\newcommand{\enes}{eNES}
\newcommand{\cmaes}{CMA-ES}
\newcommand{\sepcmaes}{Sep-CMA-ES}
\newcommand{\lmmaes}{LM-MA-ES}
\newcommand{\openaies}{OpenAI ES}
\newcommand{\canonical}{Canonical ES}
\newcommand{\direct}{direct encoding}
\newcommand{\ars}{Augmented Random Search}
\newcommand{\mapelites}{MAP-Elites}
\newcommand{\cmame}{CMA-ME}
\newcommand{\pgame}{PGA-ME}

\newcommand{\gene}{GENE}
\newcommand{\longgene}{Geometric Encoding for Neural network Evolution}
\newcommand{\coordsnet}{GENE}
\newcommand{\xdgene}{XD-GENE}
\newcommand{\ltwogene}{L2-GENE}
\newcommand{\grngene}{tag-GENE}
\newcommand{\pltwogene}{pL2-GENE}
\newcommand{\coordinates}{\textit{coordinates}}
\newcommand{\distfunc}{\textit{distance function}}

\newcommand{\dqnes}{DQNES}
\newcommand{\elitES}{ElitES}
\newcommand{\customes}{Custom ES}

\newcommand{\berl}{\code{BERL}}
\newcommand{\cambrian}{\code{Cambrian.jl} }

\newcommand{\openaipaper}{\citep{salimansEvolutionStrategiesScalable2017}}
\newcommand{\canonicalpaper}{\citep{chrabaszczBackBasicsBenchmarking2018} }
\newcommand{\nespaper}{\citep{wierstraNaturalEvolutionStrategies2014}}
\newcommand{\aleguidelines}{\citep{machadoRevisitingArcadeLearning2017}}

\newcommand{\iif}{\Leftrightarrow}

\newcommand{\norm}[1]{\|#1\|}
\newcommand{\carlies}{CARLIES}
\newcommand{\jedi}{JEDi}
\newcommand{\longjedi}{Quality with Just Enough Diversity}
\newcommand{\boldjedi}{Quality with \textbf{J}ust \textbf{E}nough \textbf{Di}versity}

\newcommand{\wtf}{WTFS}
\newcommand{\longwtf}{Weighted Target Fitness Score}

\newcommand{\gdr}{GDR}
\newcommand{\longgdr}{Genetic Drift Regularization}

\newcommand{\sgdr}{GDR²}
\newcommand{\longsgdr}{Squared Genetic Drift Regularization}

\newcommand{\trgdr}{TR-GDR}
\newcommand{\longtrgdr}{Trust Region Genetic Drift Regularization}

\newcommand{\mujoco}{\textsc{MuJoCo}}

\newcommand{\halfcheetah}{\textsc{HalfCheetah}}
\newcommand{\walker}{\textsc{Walker2D}}
\newcommand{\hopper}{\textsc{Hopper}}
\newcommand{\humanoid}{\textsc{Humanoid}}
\newcommand{\swimmer}{\textsc{Swimmer}}
\newcommand{\ant}{\textsc{Ant}}
\newcommand{\antmaze}{\textsc{AntMaze}}
\newcommand{\pointmaze}{\textsc{PointMaze}}

\newcommand{\ie}{\emph{i.e.}}
\newcommand{\eg}{\emph{e.g.}}
\newcommand{\wrt}{\emph{w.r.t.}}
\newcommand{\etc}{\emph{e.t.c.}}
\newcommand{\onemax}{\textsc{OneMax}}
\newcommand{\leadingones}{\textsc{LeadingOnes}}
\newcommand{\procgen}{\textsc{ProcGen}}
\newcommand{\cartpole}{\textsc{CartPole}}
\newcommand{\acrobot}{\textsc{Acrobot}}
\newcommand{\pendulum}{\textsc{Pendulum}}
\newcommand{\qm}[1]{``#1''}
\newcommand{\note}[1]{{\color{blue}#1}}
\newcommand{\honemax}{H2}
\newcommand{\hleadingones}{H5}
\newcommand{\red}[1]{{\color{red}#1}}

\begin{abstract}

    Evolution Strategies (ES) are effective gradient-free optimization methods that can be competitive with gradient-based approaches for policy search. ES only rely on the total episodic scores of solutions in their population, from which they estimate fitness gradients for their update with no access to true gradient information. However this makes them sensitive to deceptive fitness landscapes, and they tend to only explore one way to solve a problem. Quality-Diversity methods such as \mapelites{} introduced additional information with behavior descriptors (BD) to return a population of diverse solutions, which helps exploration but leads to a large part of the evaluation budget not being focused on finding the best performing solution. Here we show that behavior information can also be leveraged to find the best policy by identifying promising search areas which can then be efficiently explored with ES. We introduce the framework of \boldjedi{} (\jedi{}) which learns the relationship between behavior and fitness to focus evaluations on solutions that matter. When trying to reach higher fitness values, \jedi{} outperforms both QD and ES methods on hard exploration tasks like mazes and on complex control problems with large policies. 

\end{abstract}


    \maketitle

\def\figwidth{0.23\textwidth}
\begin{figure}[htp]
    \begingroup
    \centering

    \subfloat[ME fitness (max: 2118) \label{fig:budget_baselines:ME:fit}]{{\includegraphics[width=\figwidth]{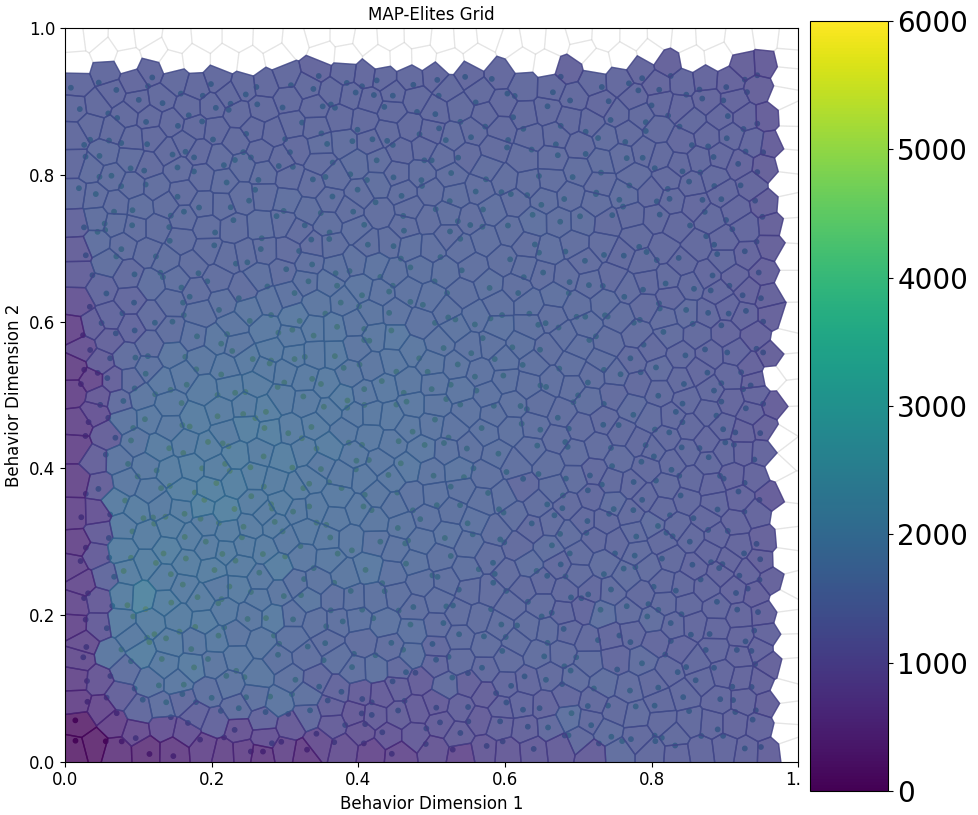} }}%
    \subfloat[ME budget (log) \label{fig:budget_baselines:ME:budget}]{{\includegraphics[width=\figwidth]{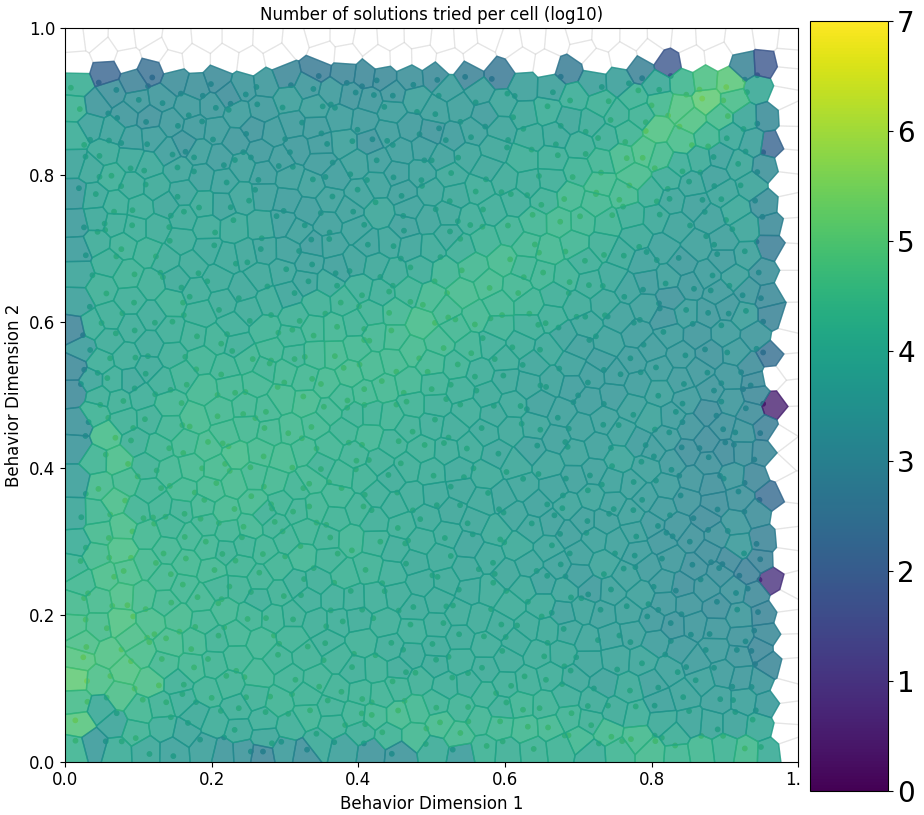} }}%

    \subfloat[ES fitness (max: 3988) \label{fig:budget_baselines:ES:fit}]{{\includegraphics[width=\figwidth]{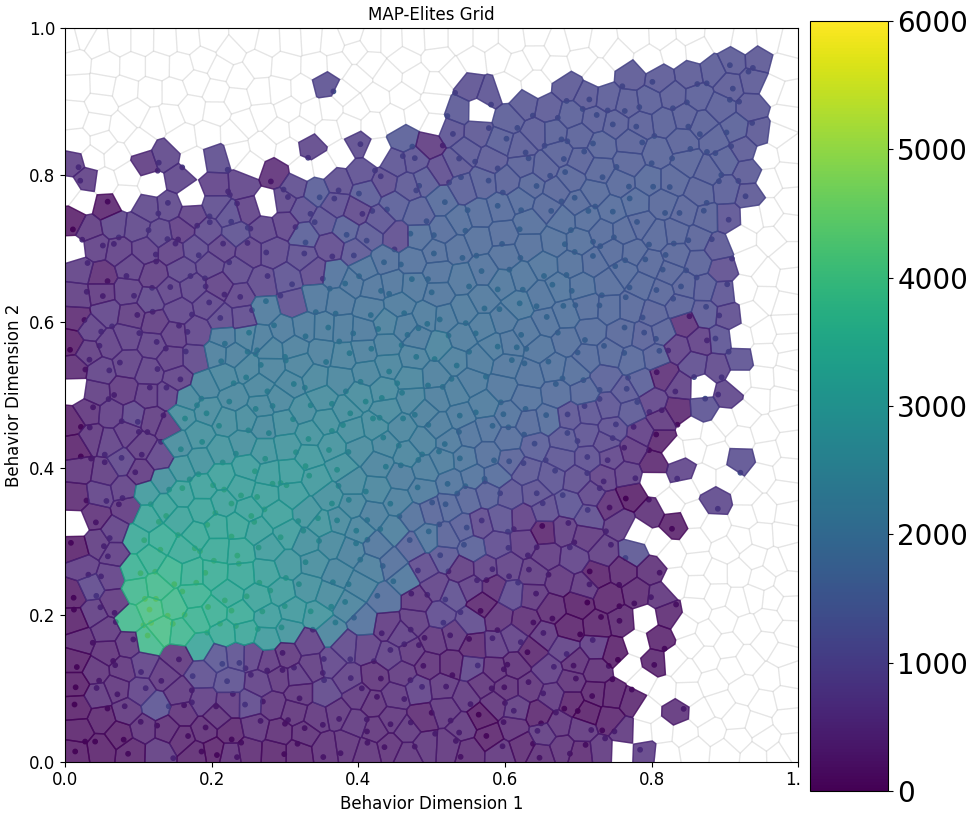} }}
    \subfloat[ES budget (log) \label{fig:budget_baselines:ES:budget}]{{\includegraphics[width=\figwidth]{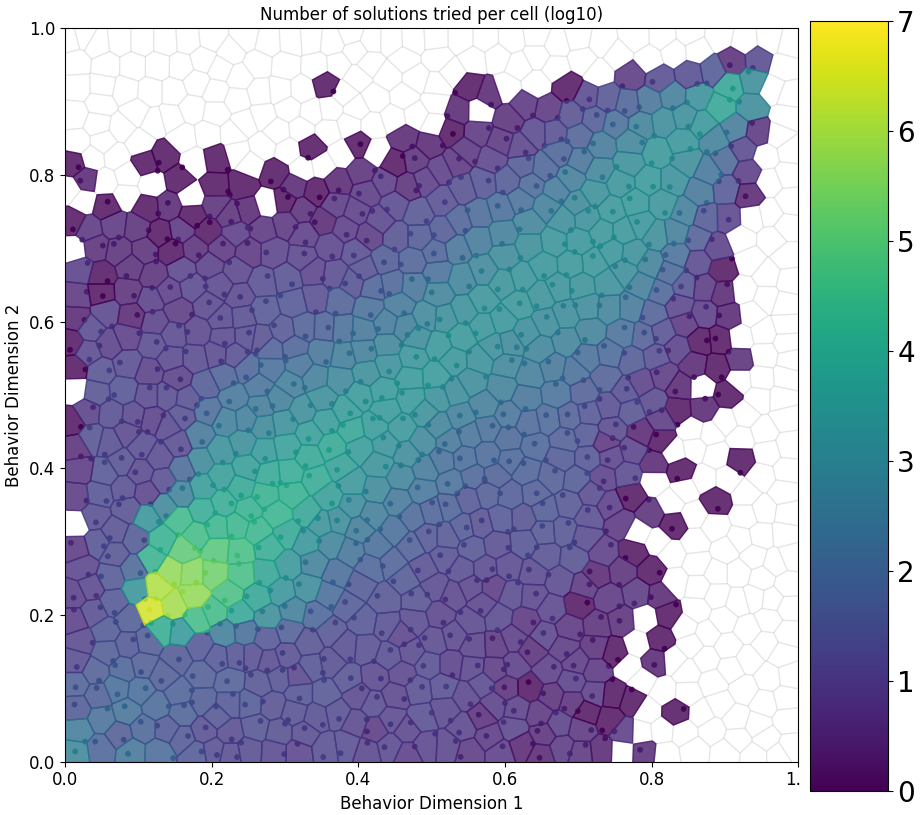} }}

    \subfloat[\jedi{} fitness (max: 5535) \label{fig:budget_jedi:fitness}]{{\includegraphics[width=\figwidth]{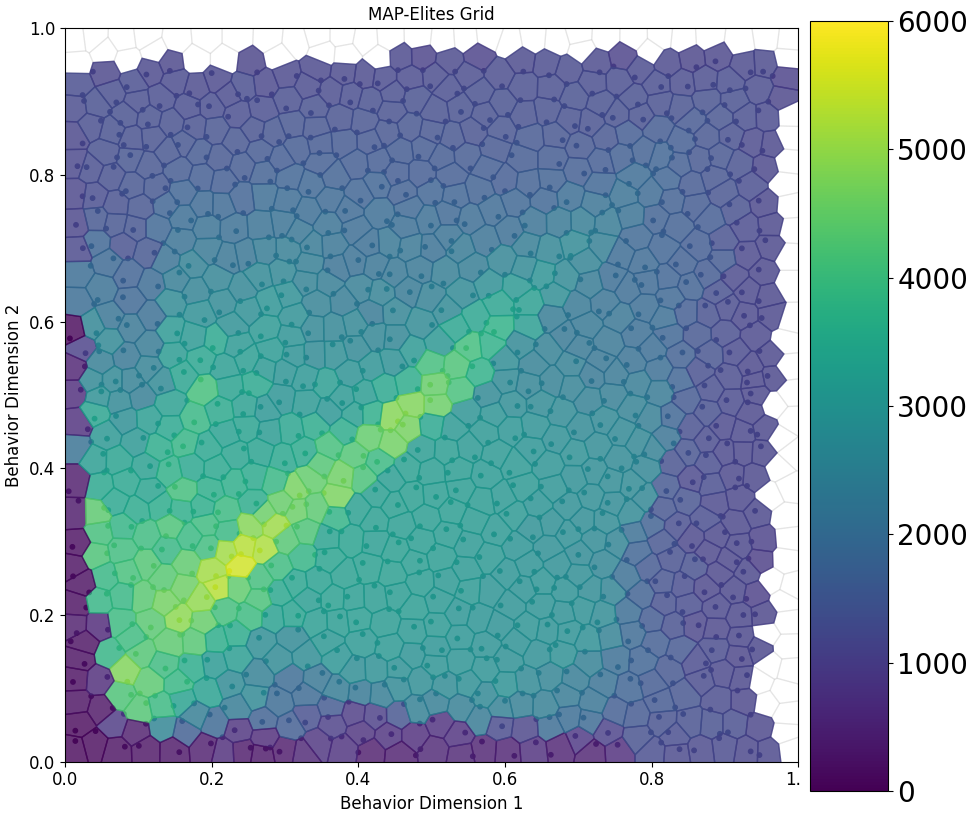} }}%
    \subfloat[\jedi{} budget (log) \label{fig:budget_jedi:budget}]
    {{\includegraphics[width=\figwidth]{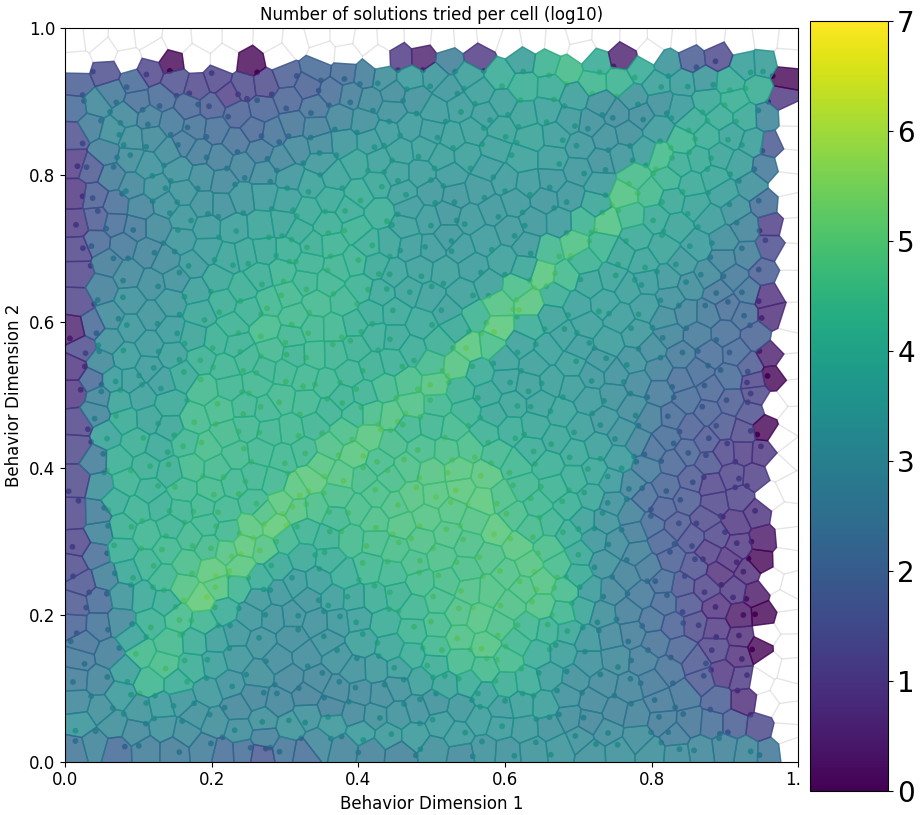} }}

    \caption{Fitness and number of solutions tried in each behavior cell for \mapelites{} (ME), Evolution Strategies (ES), and \jedi{} on \walker{}.}
    \vspace{-1.5em}
    \label{fig:budget_baselines}
    \endgroup
\end{figure}

\section{Introduction}
\label{sec:intro}

When learning a new skill like walking or playing a game, humans rarely try to find and perfect all possible ways to solve a task. They instead usually focus on the most promising approaches and try to improve on them, testing one for a while and pivoting if a different way seems to work better. While learning to limp fast is useful to adapt to a sprained ankle, it is not the most efficient way to learn to walk normally. If the final goal is only to learn to walk fast, finding the best way to walk and refining it is more efficient than finding all possible ways to walk and picking the fastest one.

In evolutionary optimization, Quality-Diversity (QD) methods such as \mapelites{} \citep{mouretIlluminatingSearchSpaces2015} keep a population of policies with diverse behaviors, using different solutions as intermediary steps ("stepping stones") to find ones that reach higher scores without getting stuck in local optima. The final archive of solutions can be used to adapt to damaged robots without training again \citep{cullyRobotsThatCan2015}, justifying the need for a full repertoire of good policies. However a large portion of the evaluation budget is used to improve non-best individuals in order to have a globally well-performing archive. 

On the other hand, Evolution Strategies (ES) focus purely on reaching the highest possible fitness on one individual, with no information from the behavior of policies. Since these methods do not keep a diversity of policies, they cannot benefit from the stepping stones of QD and tend to focus all evaluations on the same behavior (see \Cref{fig:budget_baselines}).

To illustrate the difference in behavior space exploration we compare in \Cref{fig:budget_baselines} the highest fitness (left) and the budget distribution (right) of \mapelites{} and an ES (\lmmaes{} \cite{loshchilov2017limited} here) on the \walker{} control task. As expected, coverage is better with \mapelites{} which explicitly tries to cover the behavior space while ES do not (\Cref{fig:budget_baselines}). \mapelites{} finds a large variety of good solutions, with close fitness values across the archive. This is complemented by the budget distribution, which shows that \mapelites{} has a homogeneous distribution of evaluations across the behavior space. The ES on the other hand focuses on a few behavior cells that perform well, with a very large number of evaluations spent in these cells and furthest behaviors never explored. The budget distribution shows a few cells with a large number of evaluations and a lot of cells with few to no evaluations. On this task the ES finds a better solution than \mapelites{}, but all its good solutions solve the task in a similar way and very few of the evaluations yielded different behaviors.

In this work we look at a way to use behavior information in an ES approach by introducing the framework of \boldjedi{} (or \jedi{}), which explores the behavior space enough to gain an idea of which behaviors might perform well while still focusing only on reaching the highest possible maximal fitness like an ES. \jedi{} learns the relationship between behavior and fitness with a Gaussian Process (GP) to select target behaviors that could lead to high fitness. It then uses Evolution Strategies to find high-performing solutions close to each target behavior, and uses the evaluated solutions to update the GP. This allows \jedi{} to focus on reaching high fitness values while still exploring the behavior space enough to look for potential important stepping stones. We show with \jedi{} that deliberately exploring the behavior space allows to focus evaluation budget on promising behaviors and to reach higher fitness values faster than QD methods, while using behavior information also helps solve problems where an ES would get stuck.

\section{Related work}
\label{sec:related}

\subsection{Evolution Strategies}

Evolution Strategies (ES, \citep{rechenberg1978evolutionsstrategien}) are single-objective gradient-free continuous optimization methods that use a distribution over the search space to generate new solutions. The distribution is updated at each iteration to favor the best solutions, and the process is repeated until convergence. ES have been used in policy search for a long time and have been shown to be competitive with deep reinforcement learning methods \citep{chrabaszczBackBasicsBenchmarking2018,salimansEvolutionStrategiesScalable2017}. Covariance Matrix Adaptation ES (\cmaes{}, \citep{hansenCMAEvolutionStrategy2016}) is a well-known high-performing ES that exploits the covariance matrix of the distribution to adapt its shape, which yields high performance to the cost of a higher computational complexity. To tackle this issue, variants such as Separable \cmaes{} (\sepcmaes{}, \citep{rosSimpleModificationCMAES2008}) and Limited Memory MA-ES (\lmmaes{}, \citep{loshchilov2017limited}) have been proposed.

To improve search and get out of local optima, restart methods such as Local-Restart \cmaes{} (LR-\cmaes{}) and Restart \cmaes{} with Increasing Population (IPOP-\cmaes{}, both from \citep{augerRestartCMAEvolution2005}) restarts the ES from a random genome when convergence is observed.

\subsection{Quality-Diversity Algorithms}

Novelty Search (NS, \citep{lehmanAbandoningObjectivesEvolution2011}) showed that finding a diversity of policies which behave differently can help find better solutions than directly optimizing for the task, notably in hard exploration problems. It defines behaviors as variables of interest to the user, such as the final position of a robot in a maze or the percentage of timesteps each robot foot touches the ground. NS then optimizes for novelty, which is the distance between the behavior of a solution and the behavior of solutions seen before.

Quality-Diversity (QD) methods build on NS by adding fitness-based competition to the search for novelty. Diverse high-performing solutions provide both stepping stones for the optimization process and a repertoire of solutions that can be used to adapt to new situations after training.
Multi-dimensional Archive of Phenotypic Elites (\mapelites{}, \citep{mouretIlluminatingSearchSpaces2015}) keeps an archive (also called \textit{repertoire}) of solutions by discretizing the behavior space into cells and keeping the best solution found in each behavior cell. At each generation a solution is sampled from the archive, mutated and evaluated. If the new solution is novel or better than the one already in the archive for its behavior, it replaces it.

\subsection{Quality-Diversity with ES}
\label{sec:qd_es}

Multiple methods have joined the efficient search of Evolution Strategies with Quality-Diversity, such as NSR-ES \citep{contiImprovingExplorationEvolution2018} for Novelty Search. Similarly, ES and \mapelites{} have been combined to improve the search of \mapelites{} with ES emitters.

To scale \mapelites{} to policy search with higher dimensions, where genetic algorithm mutation and crossover can struggle, \mapelites{}-ES (ME-ES,  \citep{colasScalingMAPElitesDeep2020}) performs an ES step on a solution from the archive to generate new solutions, based on either fitness or novelty. While it improves the mutation operator, it is not sample efficient as it requires many evaluations to add one solution to the archive.

Covariance Matrix Adaptation \mapelites{} (\cmame{}, \citep{fontaineCovarianceMatrixAdaptation2020}) takes advantage of the \cmaes{} \citep{hansenCMAEvolutionStrategy2016} algorithm to improve on \mapelites{} by adding local search with ES emitters. The \textit{Improvement} variant defines the scoring function as the improvement in fitness brought by each solution compared to the solution already present in the archive: \cmame{} then aims to improve the quality of the whole archive by looking for solutions that are either newer of better. In more recent work, CMA-MAP-Annealing (CMA-MAE, \citep{fontaineCovarianceMatrixAdaptation2023}) raises the addition threshold in the \cmame{} repertoire smoothly to allow the ES emitters to exploit before moving to further exploration, also making \cmame{} more robust to flat objectives and low-resolution archives. Its learning rate $\alpha$ controls the annealing, allowing for a smooth transition between \cmame{} and \cmaes{}.

In the domain of Differentiable QD, \cmame{} via Gradient Arborescence (CMA-MEGA, \citep{tjanakaApproximatingGradientsDifferentiable2022}) approximates the gradient for each behavior descriptor with an ES, and estimates the fitness gradient with an ES or reinforcement learning methods methods. A new population is then sampled using randomly weighted combinations of the behavior and fitness gradients, effectively searching in the subspace of solutions that have an impact on the fitness or behavior of the solutions and can hence have a better chance of being added to the archive.

\subsection{Learning relationships}

Bayesian optimization has been used in evolutionary optimization and in QD to learn surrogate models of the fitness and the behavior from genomes \citep{kentBOPElitesBayesianOptimisation2023,kentBayesianQualityDiversity2023,gaier2018data,haggDesigningAirFlow2020}. Gaussian Processes (GP) produce probabilistic predictions by modeling a distribution over functions, defined by a mean function $\mu$ and a covariance function $k$: $f \sim \mathcal{GP}(\mu, k)$. $\mu$ and $k$ are defined by a set of hyperparameters $\theta$ that can be optimized to fit data points, and next points to be evaluated can be sampled from the GP with a chosen acquisition function. Neural networks can also be used as surrogate models in QD \citep{zhangDeepSurrogateAssisted2022}, and when mixed with deep reinforcement learning the critic can compare policies \citep{wangSurrogateAssistedControllerExpensive2022}. %

The relationship between behavior and fitness has been learned by using a bandit to select which parents to mutate \cite{sfikasMonteCarloElites2021} in order to find useful offspring. While we also select behavior cells in an archive, their work selects starting points while we use the GP to get mutation targets, which can be unexplored yet. Gaussian processes to learn the behavior-fitness relationship have been also used in an adaptation phase from a complete \mapelites{} archive \citep{cullyRobotsThatCan2015}, but we use them to inform the optimization run. 

\subsection{ES in behavior space}

ES have been introduced in QD to follow fitness gradients (see \Cref{sec:qd_es}), but they can also be used to explore the behavior space. 
Differentiable QD (DQD) algorithms such as CMA-MEGA approximate behavior gradients with an ES, using them to explore the behavior space. The \textit{rnd} emitter of \cmame{} also picks a random direction in behavior space to be explored, the improvement in that direction being then used in the \cmaes{} scoring function. This shows how the behavior space can be explored with an ES.

When trying to reach a specific behavior, ARIA \citep{grillottiDonBetLuck2023} uses ES to minimize Euclidean distances in behavior space. Since ARIA's aim is to improve reproducibility in an archive already filled, it is not a baseline we compare to but it shows using an ES to reach a specific behavior is a valid approach.

\section{Method}
\label{sec:method}

Using the definition of \citet{mouretIlluminatingSearchSpaces2015}, \jedi{} is an \textit{optimization} method and not an \textit{illumination} algorithm like \mapelites{}: it aims to find the best policy and not a set of good solutions with different behaviors. However, like QD, it uses the behavior space to guide the search of the best policy. \jedi{} is composed of two main parts: a repertoire that stores the behavior of the solutions and a loop that alternates between selecting target behaviors and running ES to reach them. 

\subsection{\longjedi{}}

The approach of \jedi{} is to iteratively select behaviors that seem interesting ("\textit{targets}") and then optimize policies with these behaviors to reach high fitness values. Based on the information gathered at each step, \jedi{} learns which areas of the behavior space lead to high performing policies, to then focus on them and better select the next targets.

The main loop is presented in Algorithm \ref{algo:jedi}. Like \mapelites{}, the archive is initialized with a population of random solutions evaluated and added to the repertoire. Each step of \jedi{} learns the behavior-fitness mapping with a Gaussian Process to select target behaviors (see \Cref{sec:target_selection}), then for each of them runs an ES which finds the best solutions with a behavior close to the target (see \Cref{sec:target_reach}). After all ES are run for a fixed number of iterations $N$, the archive is updated with all the evaluated solutions like a standard \mapelites{} archive. This allows for a simple parallelization of the ES loops, and setting $n_{ES}$ to 1 is equivalent to running them sequentially and updating the Gaussian Process after each ES.

\subsection{Selecting target behaviors}
\label{sec:target_selection}

For each selected target behavior \jedi{} will run a full ES, which is expensive. To select targets in a sample-efficient way we turn to Bayesian Optimization, using Gaussian Processes (GP). A GP learns the mapping between behavior and fitness, as considered in the adaptation phase of \citet{cullyRobotsThatCan2015}, then target behaviors are selected based on potential and uncertainty using a batch sampling method for parallelization. 

\subsubsection{Behavior-fitness mapping}

We use a GP to learn the relationship between behaviors and fitness to get a smoother approximation of the landscape, reducing the need to explore each cell by locally propagating information through the GP. Data points are behavior coordinates in the repertoire, with the fitness of that point in the archive as their value. The archive stores a finite number of solutions, which limits the training cost of the GP. Policies with the same behavior can reach different fitness values, so we use the GP to learn the highest fitness reached by a policy in each cell. This trains the GP to predict the potential of a behavior (its highest fitness) through its average, and we use a weighted GP (see \Cref{sec:wgp}) to make its variance reflect the uncertainty on this highest value. As ES in \jedi{} optimize for fitness around a behavior, more evaluations in a cell increase the chance of finding good policies. 

\subsubsection{Weighted Gaussian Process}
\label{sec:wgp}
In a \mapelites{} repertoire, the value of each cell is actually the result of multiple evaluations where the highest fitness is kept. If we count each cell as 1 point in the GP, a full archive will be considered as a homogeneously explored space no matter how many evaluations were made in each cell and hence the uncertainty landscape will be flat. In order to consider \textit{how much} we explored an area instead of \textit{if} we explored it, we use a weighted GP \citep{calandrielloScalingGaussianProcess2022} which counts each point as if it appeared multiple times in the dataset (\Cref{app:wgp}).
With a weighted GP, the uncertainty landscape can then still evolve once the archive is fully covered since it depends on the number of evaluations, allowing to focus on unexplored areas of the behavior space.

\subsubsection{Target selection}
Once the GP is trained on the current repertoire, we can use it to select target behaviors for the next ES loop. We select multiple targets at once and optimize for them in parallel in order to improve the speed of training. We want these targets to balance between exploration by selecting behaviors with high fitness uncertainty, and exploitation by selecting behaviors with high predicted fitness.

In GP regression, the GP-UCB (GP Upper Confidence Bound, \citep{srinivasGaussianProcessOptimization2012}) acquisition function is commonly used to select the next point to evaluate from a weighted sum of the mean and the standard deviation predicted by the GP. To select multiple points at once GP-UCB has been extended to batch sampling with GP-BUBC (GP Batch UCB, \citep{desautels2014parallelizing}) by updating the standard deviation after each selection to amount for the expected evaluation of the target picked before. However, we use an ES to reach each target behavior and then store all evaluated solutions in the archive, so simulating the evaluation of a target with GP-BUCB is not representative of the reality. 

We instead consider target selection as a multi-objective optimization problem where we try to pick targets that maximize both the mean and standard deviation of the GP. Based on this approach we use a Pareto front of the GP predictions: the mean and variance of the GP are computed for each centroid in the archive, then target behaviors are randomly picked in the Pareto front of solutions maximizing these two values. This method yields a good balance between exploration (maximizing variance) and exploitation (maximizing the mean), with a natural hyperparameter-free trade-off and independent batch target sampling.

\subsection{Reaching target behaviors}
\label{sec:target_reach}

Once a target behavior $d_{target}$ has been picked, we want to solve the problem of maximizing the fitness $f$ under the constraint of staying close to the target behavior:
\begin{equation} \label{eq:jedi_problem}
  \max_{\phi} f(\phi) \quad \text{ s.t. } \quad \| d(\phi) - d_{target} \| \leq \epsilon
\end{equation}

In order to balance the two goals, we use the Lagrangian relaxation of the problem and we introduce the \longwtf{} (\wtf{}, \Cref{eq:wtf}) as a way to combine the goals of reaching a target and maximizing the fitness in the environment. We define a scoring function $\mathcal{S}$ which gives a utility score to each individual based on its fitness and behavior. This utility score is then used by the rank-based ES update, so the value of the score only matters relative to the other scores, not its absolute value. To make it task agnostic and independent from fitness and behavior ranges, we normalize the fitness and target scores $\mathcal{S}_{fitness}$ and $\mathcal{S}_{target}$ to be between 0 and 1, both being maximized. With $f_{min}$ and $f_{max}$ the minimum and maximum fitness values in the current ES population, the normalized fitness score for a solution $\phi_i$ is:
\begin{equation}
  \mathcal{S}_{fitness}(\phi_i) = \frac{f(\phi_i) - f_{min}}{f_{max} - f_{min}}
\end{equation}

To reach a target behavior, we minimize the Euclidean distance between the behavior descriptor $d(\phi_i)$ of an individual $\phi_i$ and the target behavior descriptor $d_{target}$:
\begin{equation}
  \mathcal{S}_{target}(\phi_i) = 1 - \frac{\| d(\phi_i) - d_{target} \| - d_{min}}{d_{max} - d_{min}}
\end{equation}
with $d_{min}$ and $d_{max}$ the minimum and maximum distance between the target behavior and behavior descriptors of the current ES population. The final \longwtf{} is then:
\begin{equation}\label{eq:wtf}
  \mathcal{S}_{\wtf{}}^\alpha(\phi_i) = \alpha \mathcal{S}_{target}(\phi_i) + (1 - \alpha) \mathcal{S}_{fitness}(\phi_i)
\end{equation}

with $\alpha$ the weight of the target score in the global scoring function. $\alpha = 0$ corresponds to pure exploitation, $\alpha = 1$ to pure exploration and $\alpha = 0.5$ to a balance between the two. The value of $\alpha$ can also be annealed during the run to first focus on exploration and then on exploitation.

Each ES center is initialized with the genome of the solution in the archive which has the closest behavior to the target, since we intend to minimize the Euclidean distance to it. It runs for $N$ generations with a population of $\lambda$ individuals and uses the \jedi{} score function $\mathcal{S}^\alpha_{\wtf{}}$. In this paper we use \sepcmaes{} or \lmmaes{}, but any ES method can be used.

\begin{algorithm}[th]
    \caption{JEDi}
    \label{algo:jedi}
    
    \KwData{
    JEDi scoring function $ \mathcal{S} $, 
    JEDi target selection function $ \mathcal{T} $, 
    Number of ES emitters $n_{ES}$, 
    Number of JEDi loops $L$, 
    Repertoire $R$, 
    Rollout function $r$, 
    Number of ES generations $N$, 
    ES population size $\lambda$, 
    ES parameters $\theta$, 
    ES update function $U$,
    $\alpha$ the \wtf{} parameter 
    }
    
    \BlankLine
    
    \For{$l \leftarrow 1$ \KwTo $L$}{
        Initialize the repertoire $R$ with $n_{init}$ random genomes\;
        Sample $n_{ES}$ behavior targets: $(t_i)_{1..n_{ES}} =  \mathcal{T} (R, n_{ES})$\;
        Update $\alpha$ in case of annealing\;  
        \For{$i \leftarrow 1$ \KwTo $n_{ES}$}{
            Select the starting genome from the repertoire: $\phi_0, d_0 = \arg\min_{\phi, d \in R} \| d - t_i \|$\;
            Initialize the ES center: $\theta = \phi_0$\;
            \For{$g \leftarrow 1$ \KwTo $N$}{
                Sample $\lambda$ genomes: $\{\phi_k\}_{k \in \{1, ..., \lambda\}} \sim \theta$\;
                Get fitness values and descriptors: $\forall k, (f_k, d_k) = r(\phi_k)$\;
                Compute scores: $\forall k, s_k = \mathcal{S}_{\wtf{}}^\alpha (\phi_k)$\;
                Update $\theta $ using scores\;
            }
        Update the repertoire $R$ with all evaluated genomes\;
        }
    }
  \end{algorithm}

\section{Experiments and Results}
\label{sec:results}

\subsection{Control tasks}

In many standard QD benchmarks, like \textsc{AntOmni}, the main target of the optimization is the behavior and the fitness is a cost metric to minimize, \eg{} the energy needed to reach a position. Focusing on the fitness alone will lead to not moving and not having any useful solution. We hence focus here on tasks where the fitness is the main target of the optimization (a higher fitness means more a useful solution) and the behavior is a side effect. We devise two sets of tasks to test \jedi{} on environments presenting increasing levels of difficulty.

The first set includes 3 mazes A, B and C presented in \Cref{fig:mazes} and increasingly difficult to explore. Based on Kheperax \cite{grillottiKheperaxLightweightJAXbased2023}, the agent is a robot with 3 lidars and 2 bumpers, moving in a 2D maze with a fixed target. Actions are the speed for its 2 wheels. There is no real physical model used, and the control of the robot can be done with small policies. An episode ends when the agent reaches the target or after 250 steps. If the agent reaches the target, its fitness is minus the number of steps used to reach the target. If the agent does not reach the target, its fitness is -250 minus the Euclidean distance to the target scaled by 100. Averaging fitness values between a run that reached the target and one that did not would make the scaling factor of the distance bias the result so we use the median to aggregate results. This fitness incentivizes the agent to reach the target as fast as possible, but makes the environment deceptive since getting higher fitness values requires going through the maze and momentarily going away from the target. Behavior descriptors (BD) are the final position of the robot at the end of the episode. To make exploration harder, we create a fourth maze Quad B based on Maze B, where the maze is quadrupled in X and Y directions as presented in \Cref{fig:quad_maze}. Robots start at the center of the new maze and only one quadrant has a target to reach.

\def\figwidth{0.12\textwidth}
\begin{figure}[th]
	\begingroup
	\centering

	\subfloat[Maze A]{{\includegraphics[width=\figwidth]{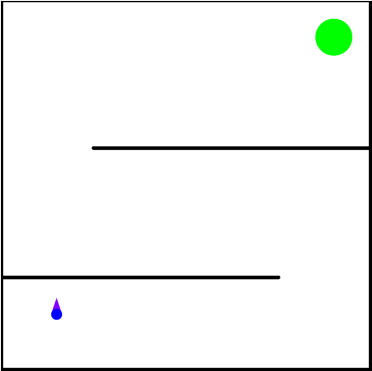}}}
	\qquad
	\subfloat[Maze B]{{\includegraphics[width=\figwidth]{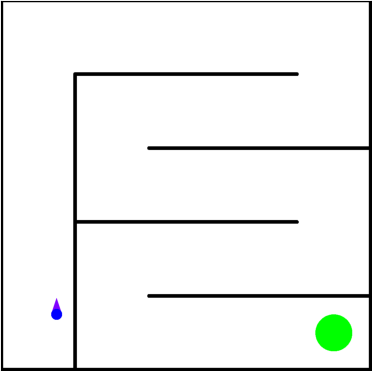}}}
	\qquad
	\subfloat[Maze C]{{\includegraphics[width=\figwidth]{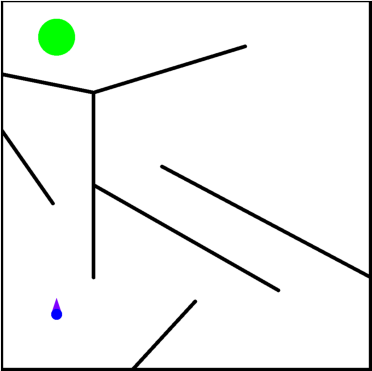}}}

	\caption{Mazes for robot exploration. Robots start at the blue point and the target is the green circle. }
	\label{fig:mazes}

	\endgroup
\end{figure}

The other set of problems is based on Brax control tasks \citep{brax2021github}: \halfcheetah{} and \walker{} which require to make a cheetah-like and a bipedal robot walk faster in 2D, and \antmaze{} where a 3D ant-like robot 4-legged robot needs to reach a target in a 3D maze. 
We reduce the length of \halfcheetah{} episodes to 500 steps (from standard 1000) to allow the ES to reach evaluation limits in 24h without using very large populations which would impact its sample efficiency. This still allows the cheetah to start running but explains the difference in scores with other works. Behavior descriptors for \halfcheetah{} and \walker{} are the percentage of timesteps where each foot was in contact with the floor, moving away from tasks where the BD and fitness are highly correlated like mazes. Behavior descriptors for \antmaze{} are the final position of the robot at the end of the episode. \antmaze{} requires large policies to be solved \citep{cideronQDRLEfficientMixing2020, chalumeauAssessingQualityDiversityNeuroEvolution2023} so use policies with 2 hidden layers of 256 neurons for all Brax tasks. This is a larger architecture than the one used in \citep{langeEvosaxJAXbasedEvolution2022}, with genomes an order of magnitude larger making the optimization harder.
Implementation details are described in \Cref{app:implem} and research code will be made available.

\subsection{Baselines}
The ES baselines presented here are two cheaper \cmaes{} variants: \sepcmaes{} for maze tasks and \lmmaes{} for Brax tasks to reduce memory footprint. They are presented in both standard and restart versions, with a simple restart criterion based on fitness convergence in the population since the full internal state of \cmaes{} is not available for the BIPOP criterion. 
As QD baselines we show \mapelites{}, \cmame{} with \textit{Imp} and \textit{Opt} emitters and CMA-MAE. As \citet{tjanakaTrainingDiverseHighDimensional2023} show their ES version of CMA-MEGA is outperformed by Policy Gradient Assisted \mapelites{} (\pgame{}, \citep{nilssonPolicyGradientAssisted2021a}) we use \pgame{} with their hyperparameters as a baseline instead. \pgame{} uses reinforcement learning gradients contrary to \jedi{} and all other methods presented here, making it a strong baseline for policy search. A more fair comparison would be to use the same RL gradients in \jedi{}, but we leave this to future work.

\subsection{Results}

\def\figwidth{0.3\linewidth}
\def\xleg{Evaluations}
\def\yleg{Max fitness}
\begin{table*}[htp]
	\centering
	\begin{tabular}{cccc}
		                                                                                         & Maze A         & Maze B    & Maze C     \\
		\raisebox{5\normalbaselineskip}[1cm][0cm]{\rotatebox[origin=c]{90}{\vspace{1cm}\yleg{}}} &
		\includegraphics[width=\figwidth]{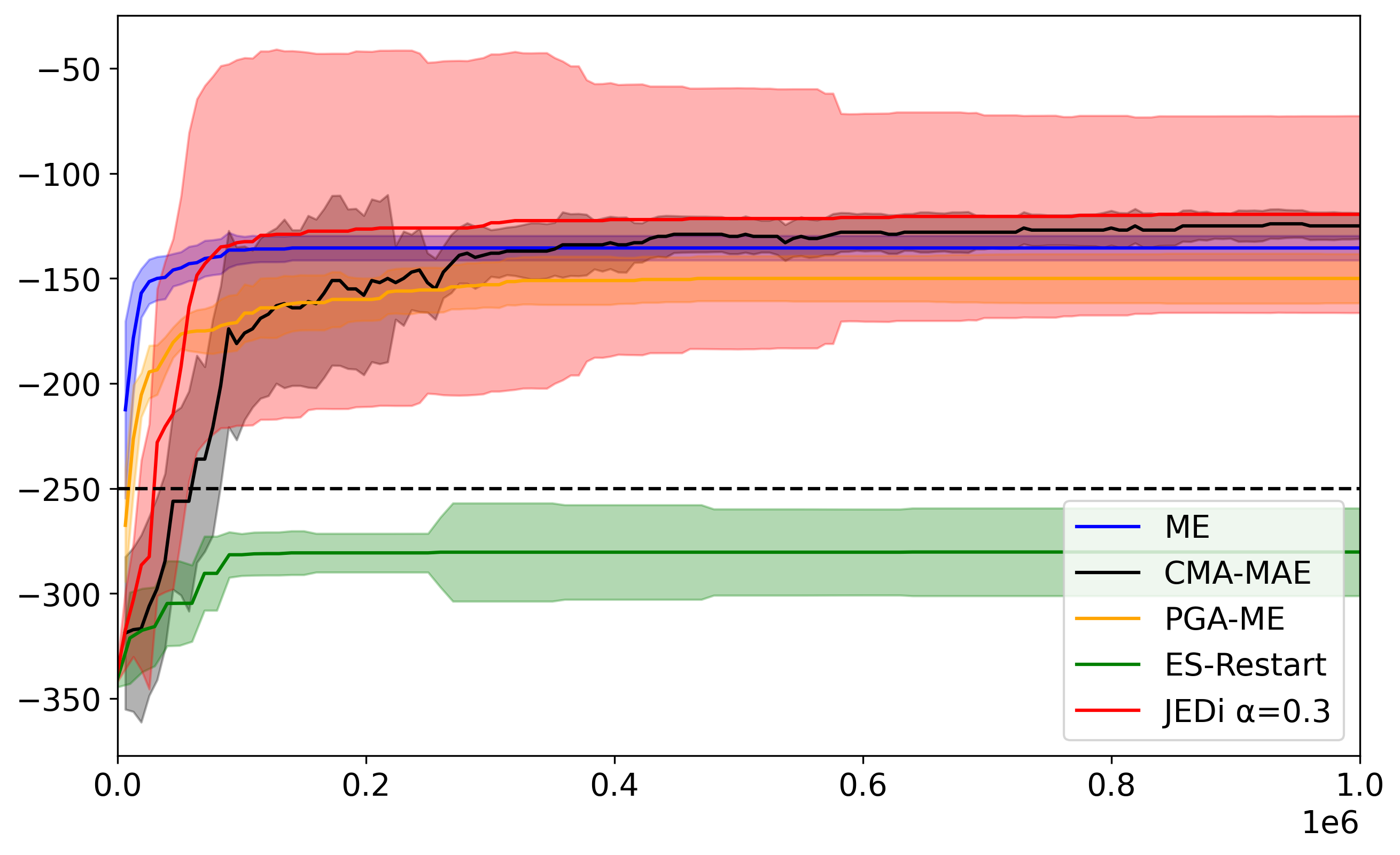}                &
		\includegraphics[width=\figwidth]{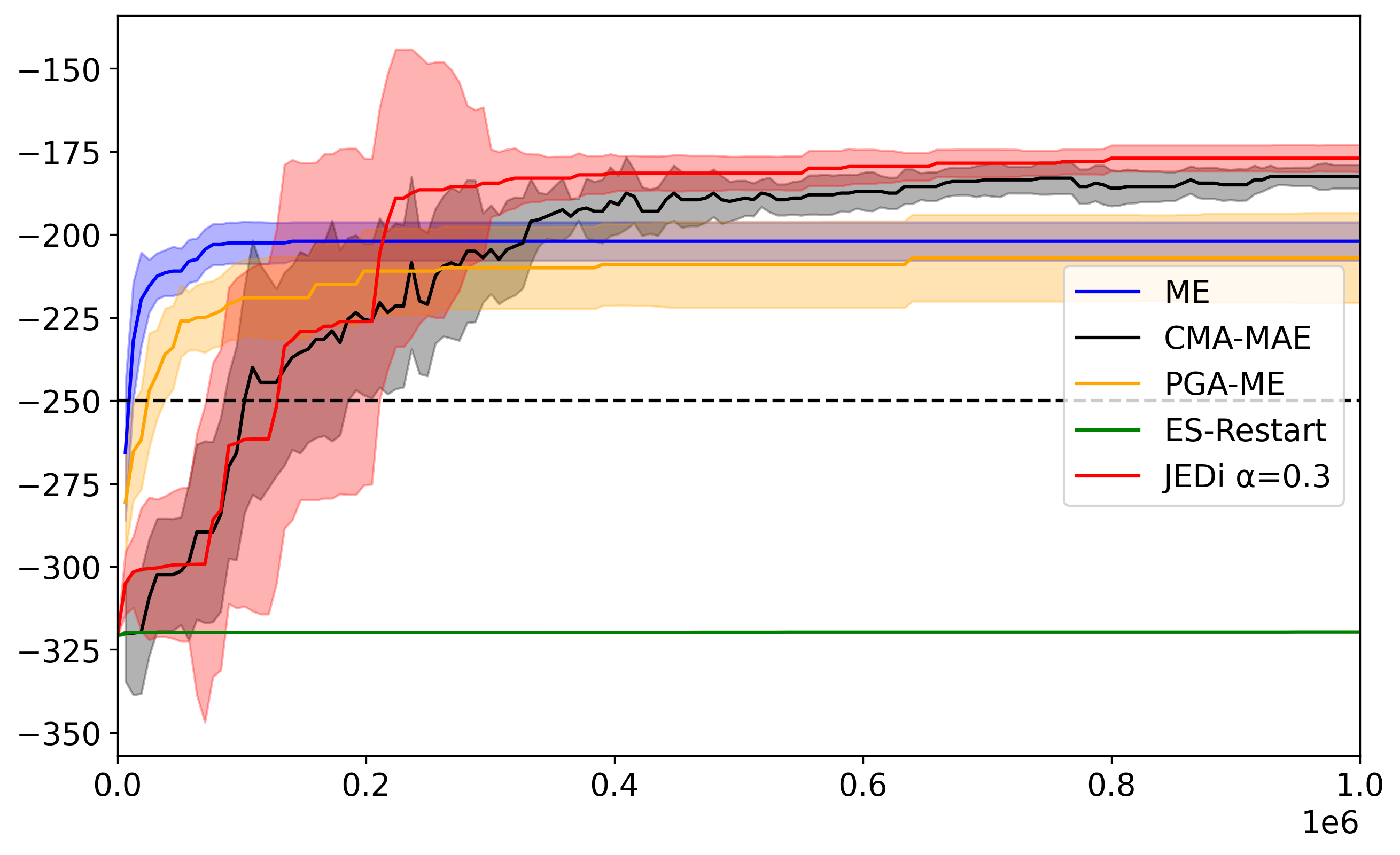}                    &
		\includegraphics[width=\figwidth]{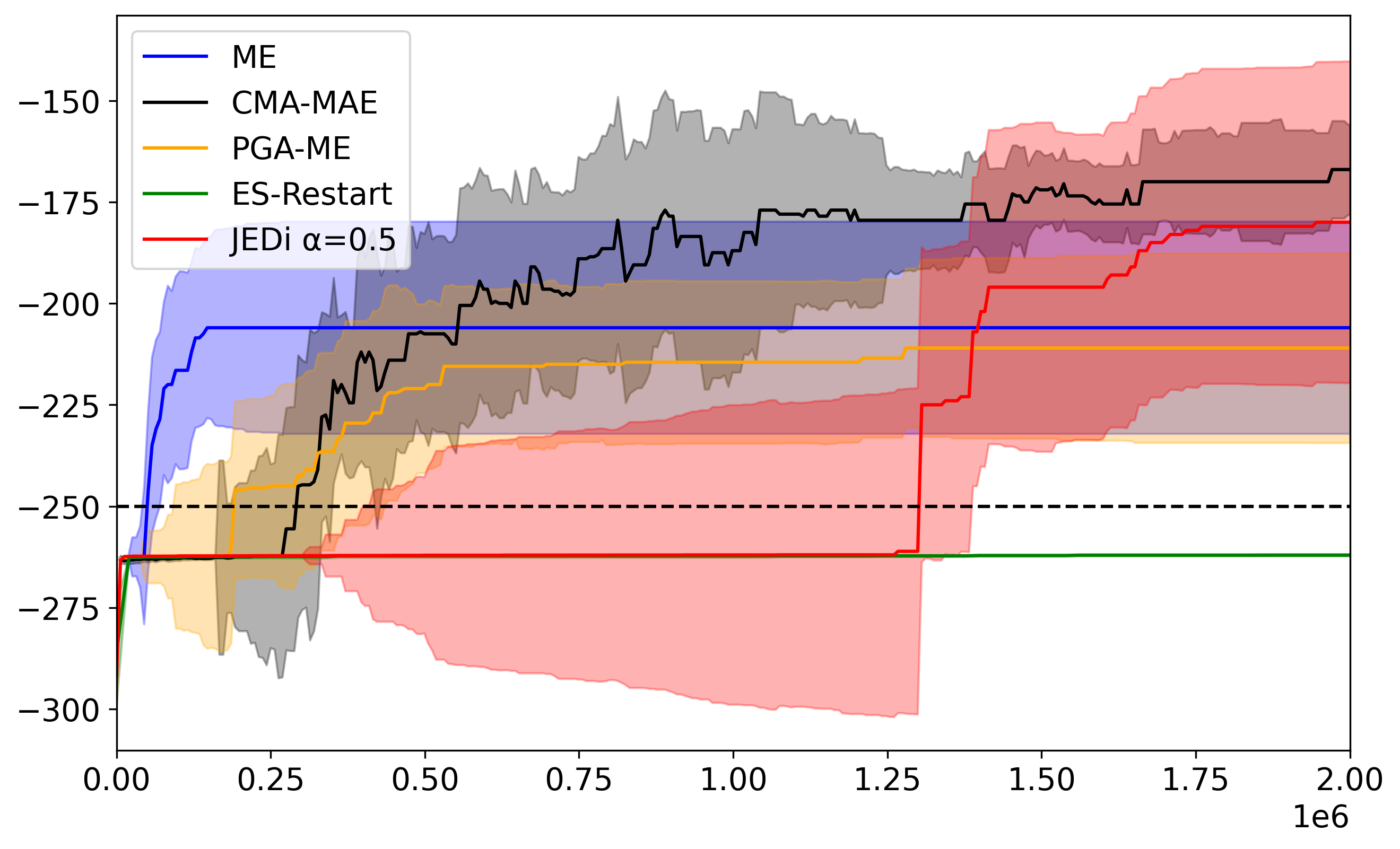}                                                           \\

		                                                                                         & \halfcheetah{} & \walker{} & \antmaze{} \\
		\raisebox{5\normalbaselineskip}[1cm][0cm]{\rotatebox[origin=c]{90}{\vspace{1cm}\yleg{}}} &
		\includegraphics[width=\figwidth]{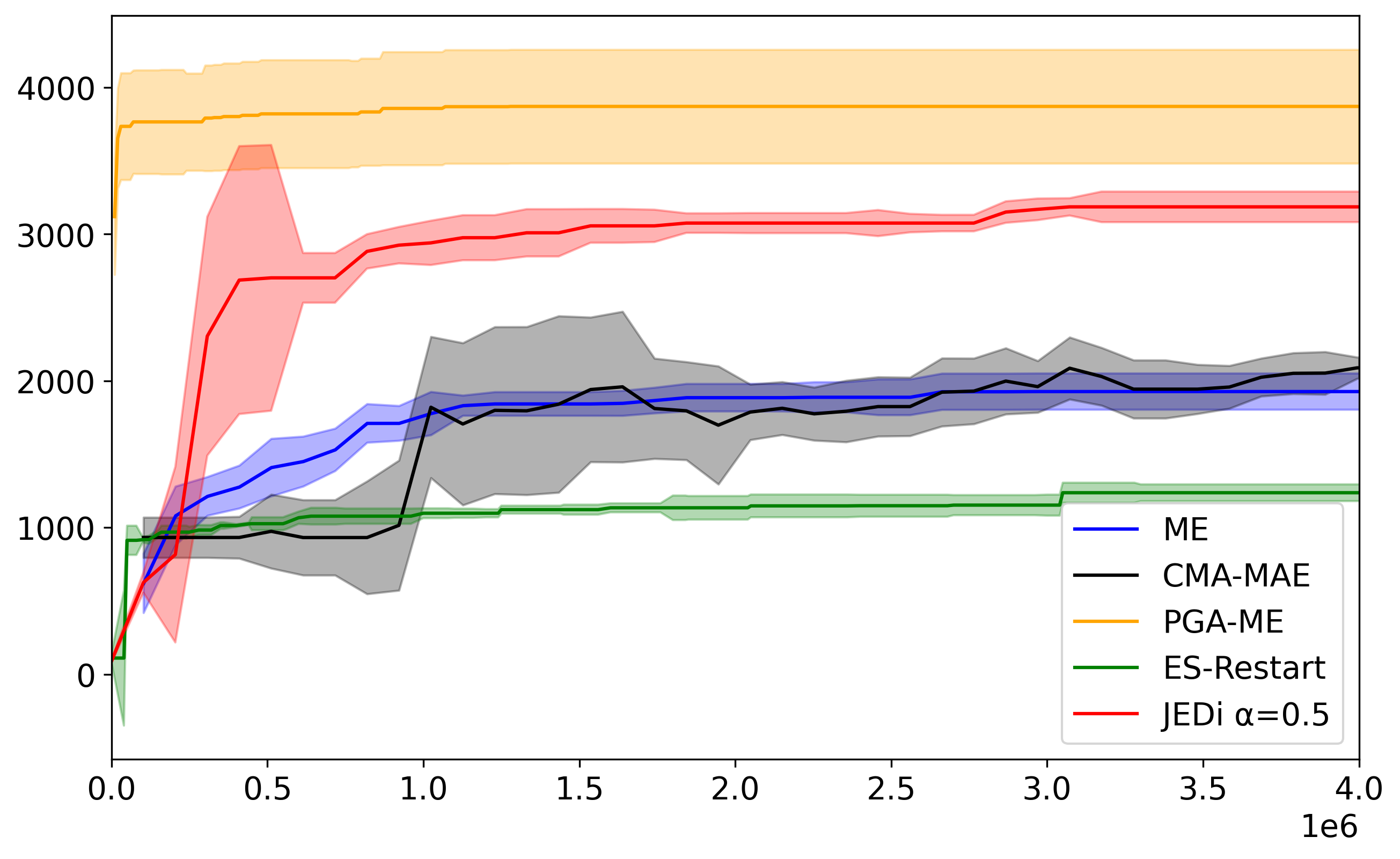}             &
		\includegraphics[width=\figwidth]{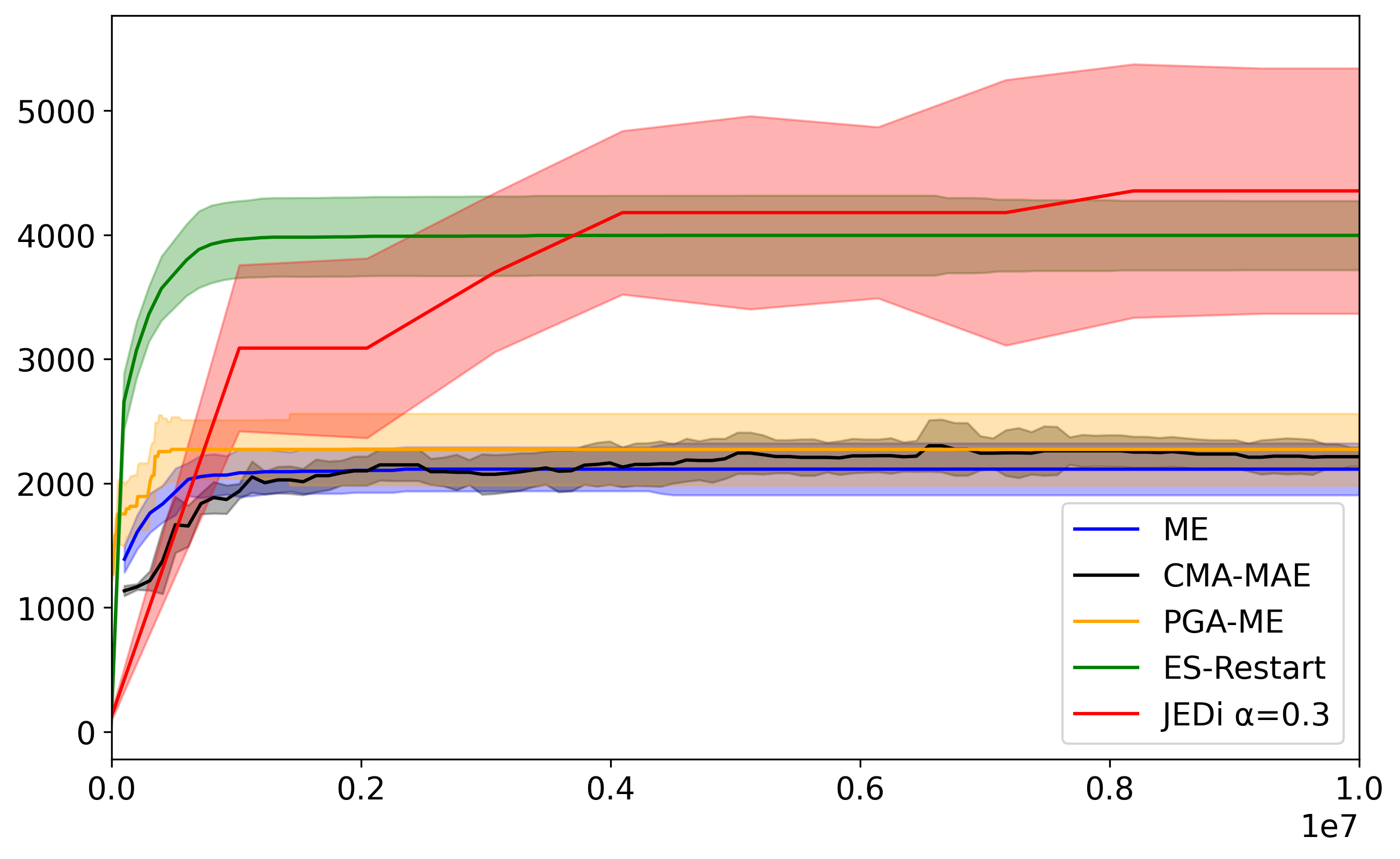}               &
		\includegraphics[width=\figwidth]{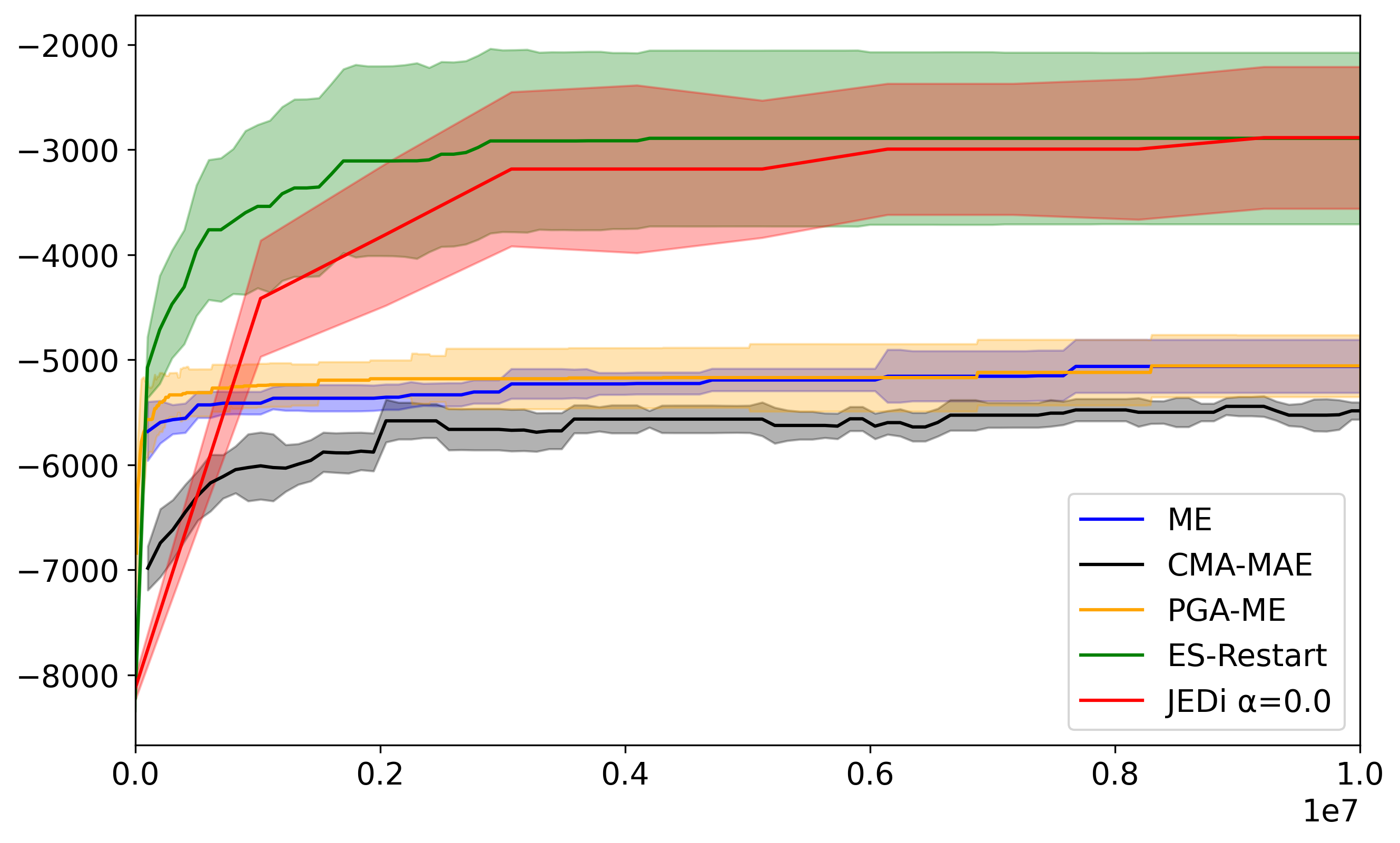}                                                               \\

		                                                                                         & \xleg          & \xleg     & \xleg
	\end{tabular}
	\captionsetup{type=figure}
	\caption{Fitness results for maze exploration (row 1) and robotics control tasks (row 2). Reaching the dotted line at -250 in a maze means an agent has reached the target. Strong line is the median over all runs, with colored areas showing $\pm 1$ standard deviation over and under the median.}
	\label{plot:results}
\end{table*}

\def\figwidth{0.3\linewidth}
\def\xleg{Evaluations}
\def\yleg{Max fitness}
\begin{table*}[h!]
	\centering
	\begin{tabular}{cccc}
		                                                                                         & Maze A         & Maze B    & Maze C     \\
		\raisebox{3\normalbaselineskip}[1cm][0cm]{\rotatebox[origin=c]{90}{\vspace{1cm}\yleg{}}} &
		\includegraphics[width=\figwidth]{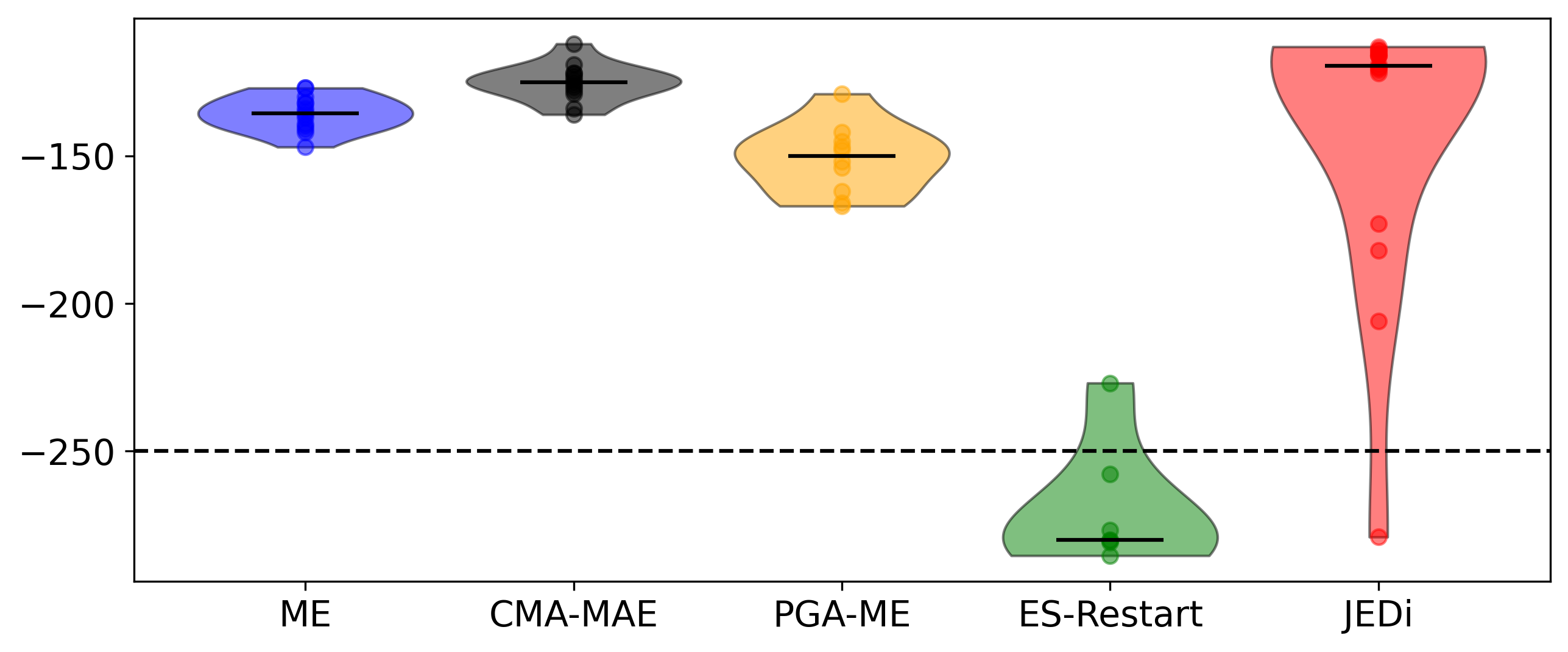}         &
		\includegraphics[width=\figwidth]{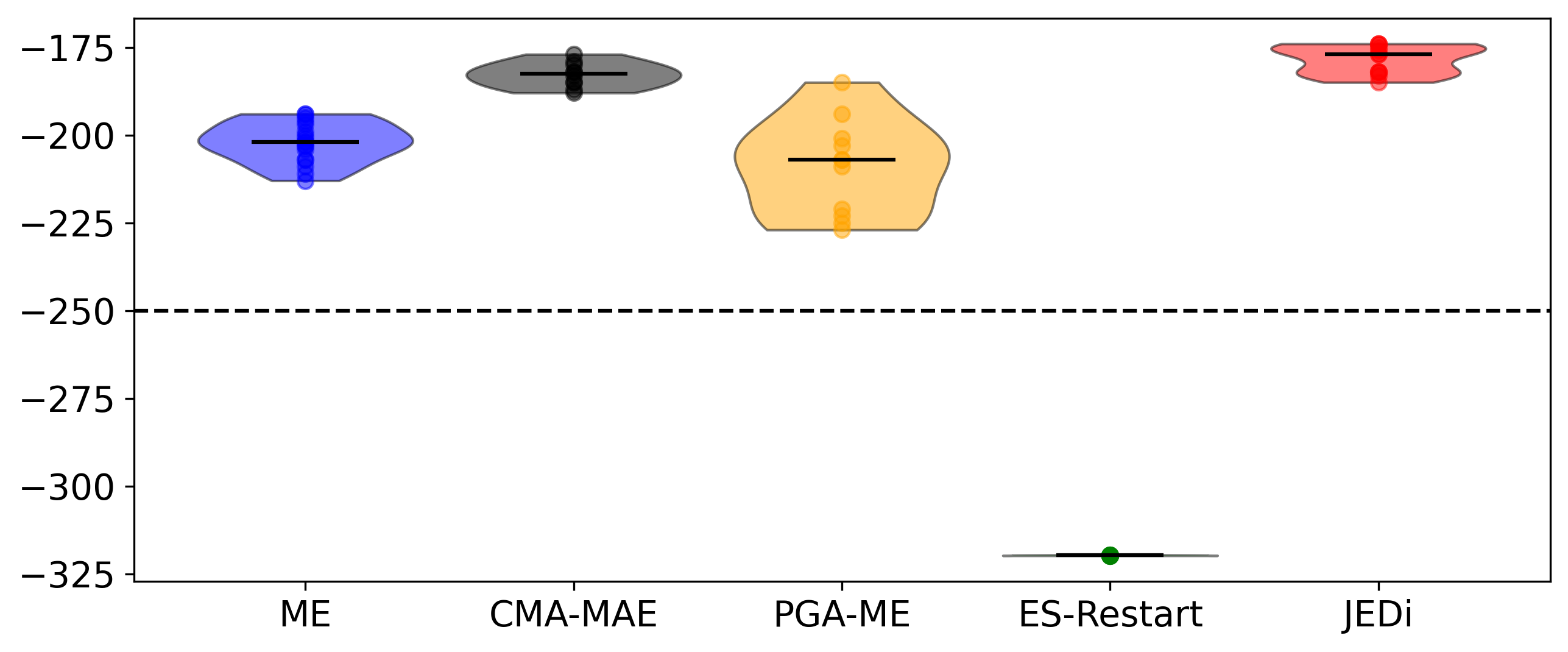}             &
		\includegraphics[width=\figwidth]{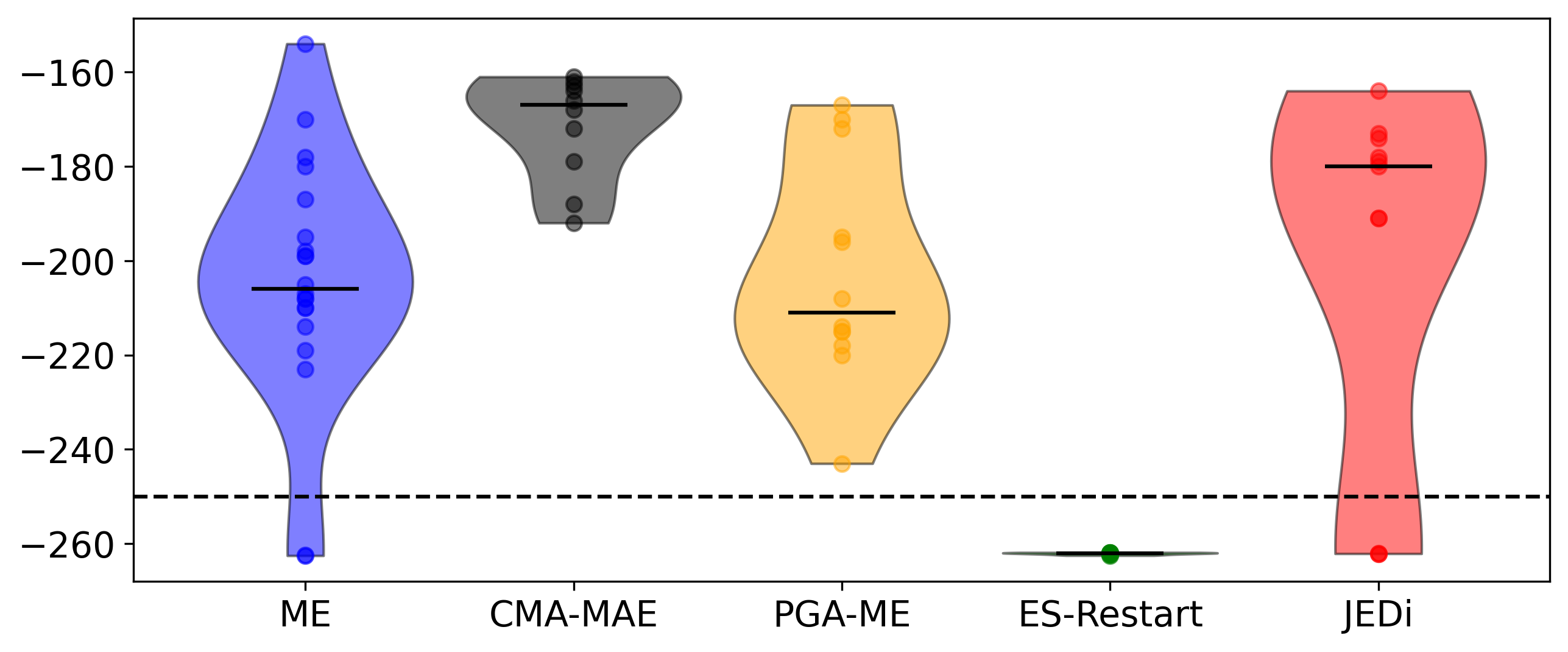}                                                    \\

		                                                                                         & \halfcheetah{} & \walker{} & \antmaze{} \\
		\raisebox{3\normalbaselineskip}[1cm][0cm]{\rotatebox[origin=c]{90}{\vspace{1cm}\yleg{}}} &
		\includegraphics[width=\figwidth]{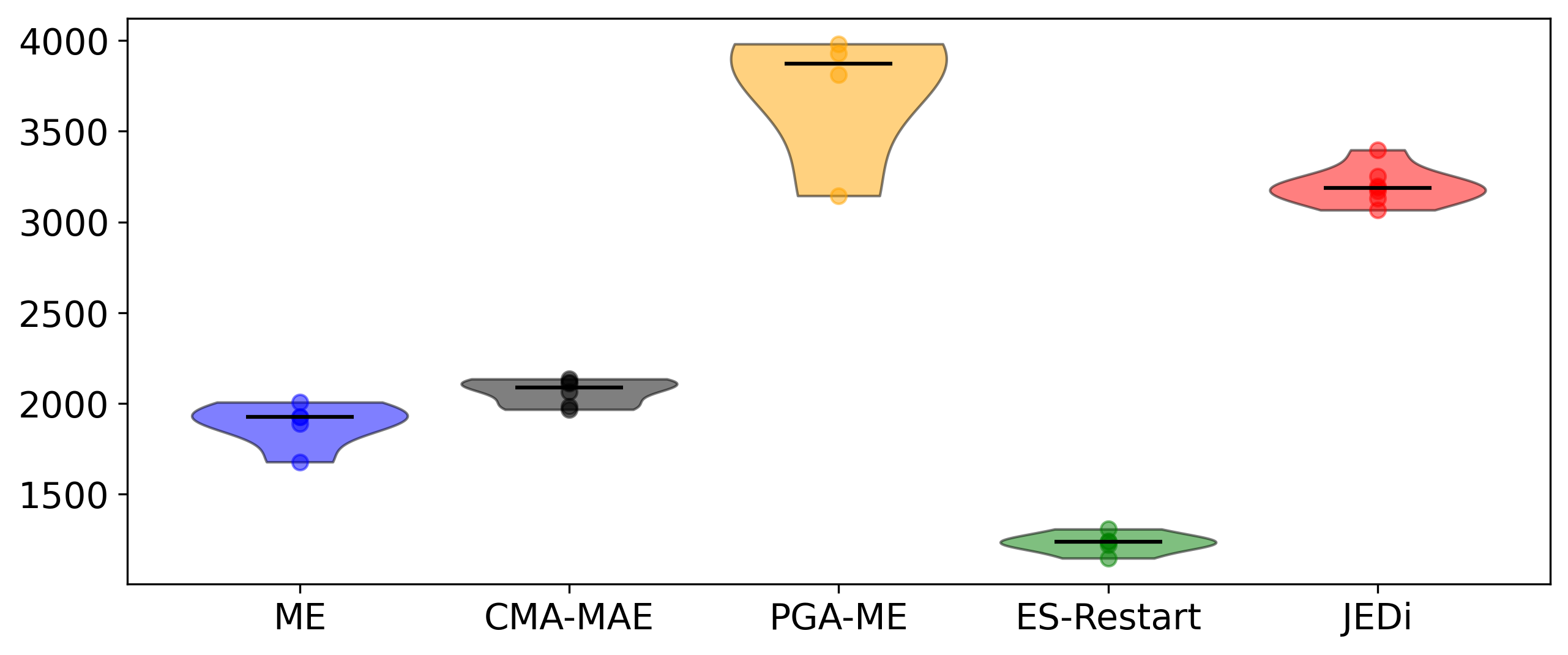}      &
		\includegraphics[width=\figwidth]{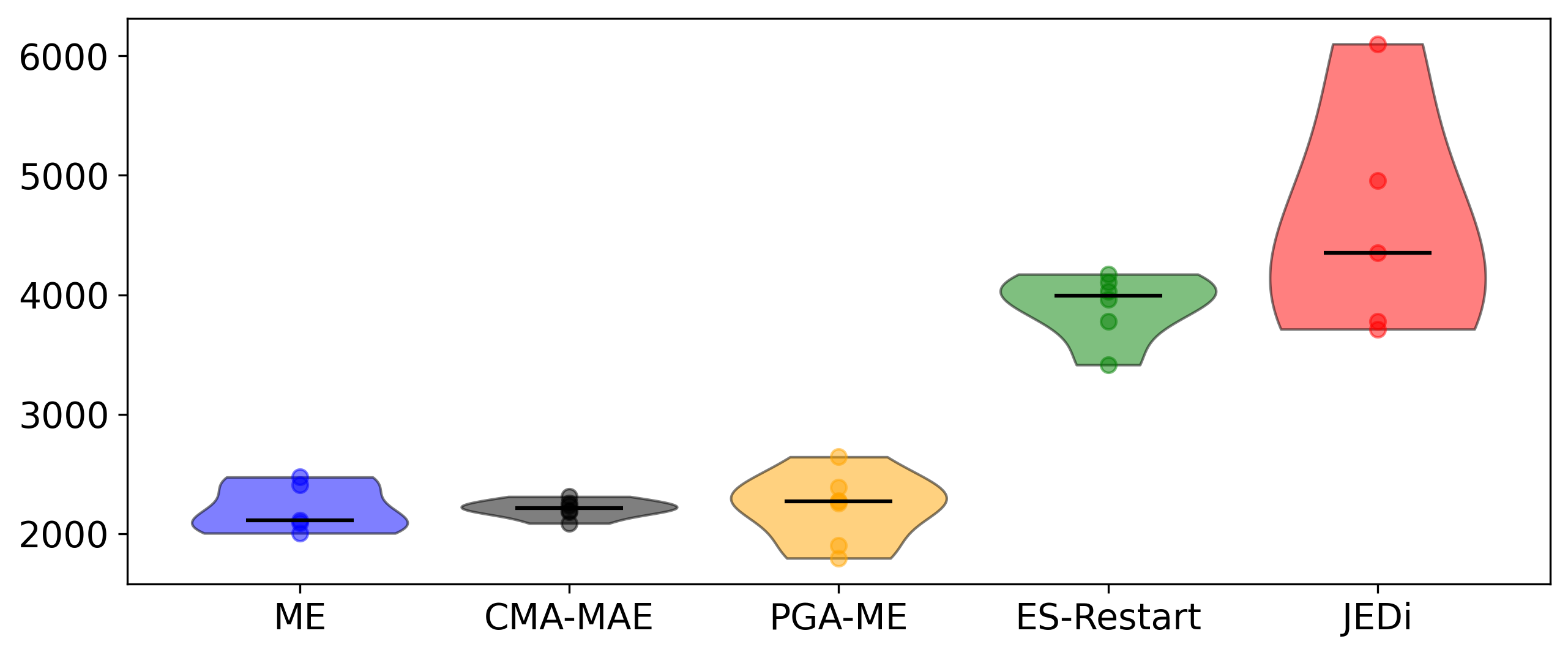}        &
		\includegraphics[width=\figwidth]{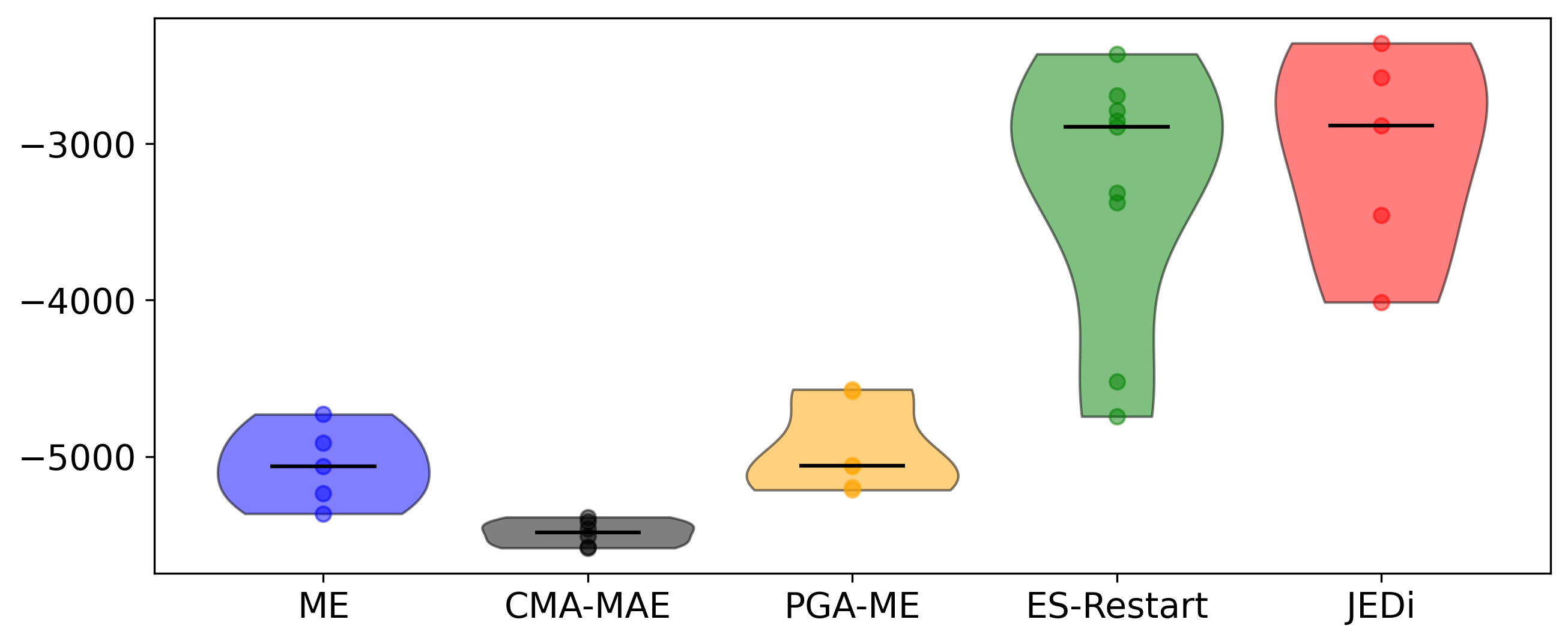}                                                        \\
	\end{tabular}
	\captionsetup{type=figure}
	\caption{Violin plots of final max fitness results for maze exploration (row 1) and robotics control tasks (row 2). The solid line in each violin is the median. Reaching the dotted line at -250 in a maze means an agent has reached the target.}

	\label{plot:violin_results}
\end{table*}

\begin{table*}[htp]
	\centering
	\begin{tabular}{l|c|c|cc|ccc|c}
\toprule
          Task &             \makecell{\jedi{}\\Tuned $\alpha$} &      \makecell{\jedi{}\\Decay} &                    ES &                        \makecell{ES\\Restart} &                 \mapelites{} &        \cmame{}  &              CMA-MAE &                       \pgame{} \\
\midrule
        Maze A & \underline{-120} &    \bfseries{-117} &           -281 (0.00) &                       -280 (0.00) & -136 (0.06) $^\star$ & -138 (0.10) $^\star$ & -125 (0.31) $^\star$ &          -150 (0.06) $^\star$ \\
        Maze B & \underline{-177} &    \bfseries{-176} &           -319 (0.00) &                       -320 (0.00) &          -202 (0.00) &          -190 (0.00) &          -182 (0.00) &                   -207 (0.00) \\
        Maze C & \underline{-180} &             -186 &           -262 (0.00) &                       -262 (0.01) & -206 (0.19) $^\star$ &          -236 (0.01) & \bfseries{-167 (0.02)} &          -211 (0.60) $^\star$ \\
   Maze Quad B & \underline{-194} &    \bfseries{-189} &           -319 (0.00) &                       -319 (0.01) &          -210 (0.03) &          -219 (0.05) & -196 (0.23) $^\star$ &          -213 (0.09) $^\star$ \\
\halfcheetah{} &             3186 & \underline{3248} &           1239 (0.00) &                       1237 (0.00) &          1927 (0.00) &          1382 (0.00) &          2089 (0.00) & \bfseries{3871 (0.26) $^\star$} \\
     \walker{} &    \bfseries{4353} &             3916 &  3967 (0.22) $^\star$ &  \underline{3995 (0.43) $^\star$} &          2114 (0.01) &          2159 (0.01) &          2214 (0.00) &                   2273 (0.00) \\
    \antmaze{} &   \bfseries{-2886} &            -2958 & -3575 (0.22) $^\star$ & \underline{-2892 (0.70) $^\star$} &         -5064 (0.01) &         -5145 (0.02) &         -5487 (0.00) &                  -5059 (0.00) \\
\bottomrule
\end{tabular}

	\caption{Statistical results: median final max fitness and p-values for the Mann–Whitney U test comparing \jedi{} with baselines. Highest fitness is in bold, second best is underlined. p-values $> 0.05$ (not significantly different) are highlighted with a star $\star$.}
	\label{tab:u_test}
\end{table*}

\def\figwidth{0.3\textwidth}
\begin{figure*}[htp]
	\begingroup
	\centering

	\subfloat[Maze Quad B]{{\includegraphics[width=0.15\textwidth,valign=t]{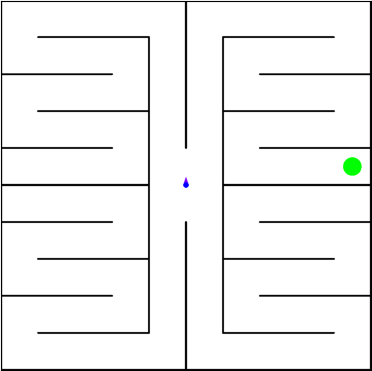} }}%
	\qquad
	\subfloat[Median fitness]{{\includegraphics[width=0.25\textwidth,valign=t]{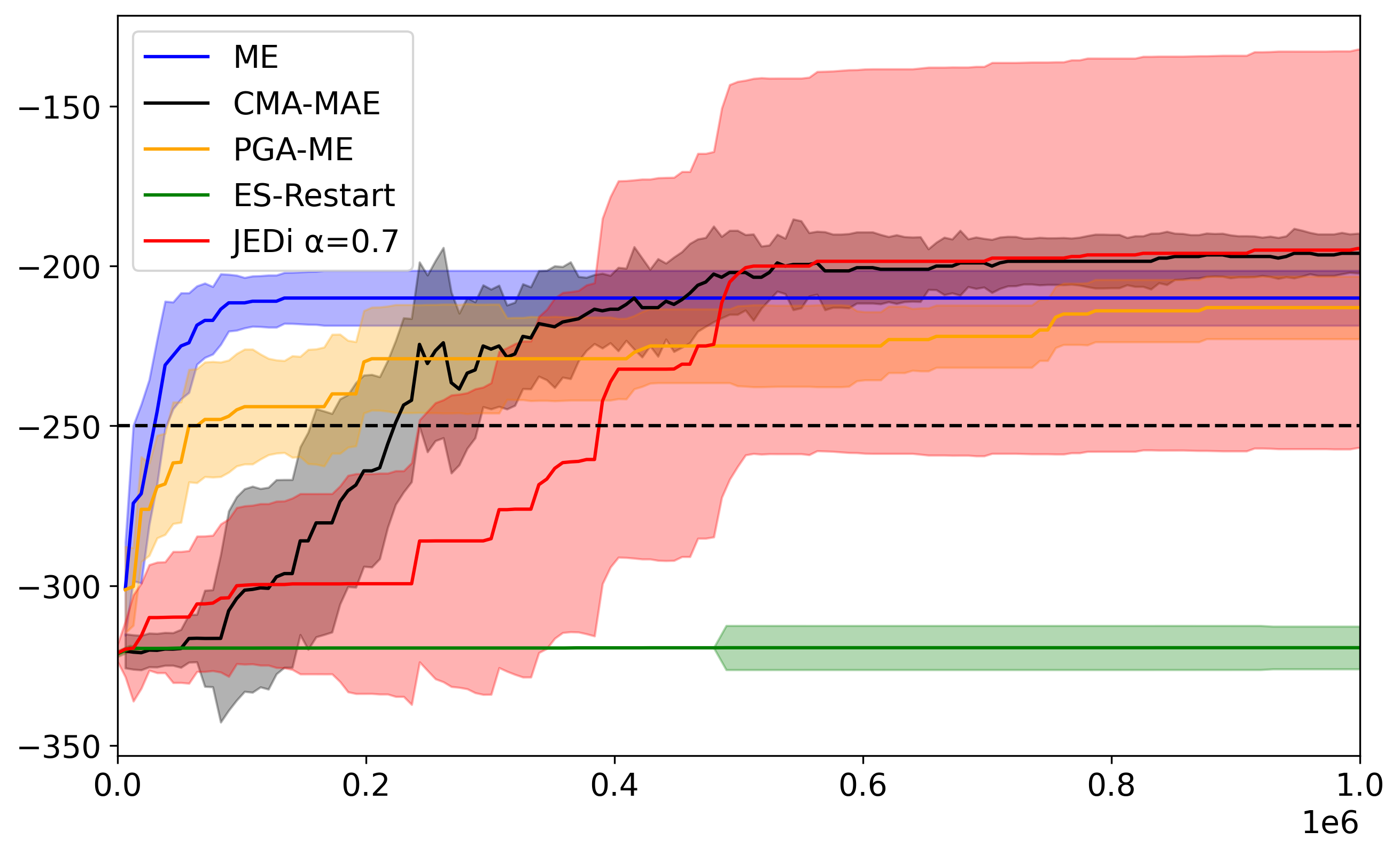} }}%
	\qquad
	\subfloat[Final fitness distribution]{{\includegraphics[width=\figwidth,valign=t]{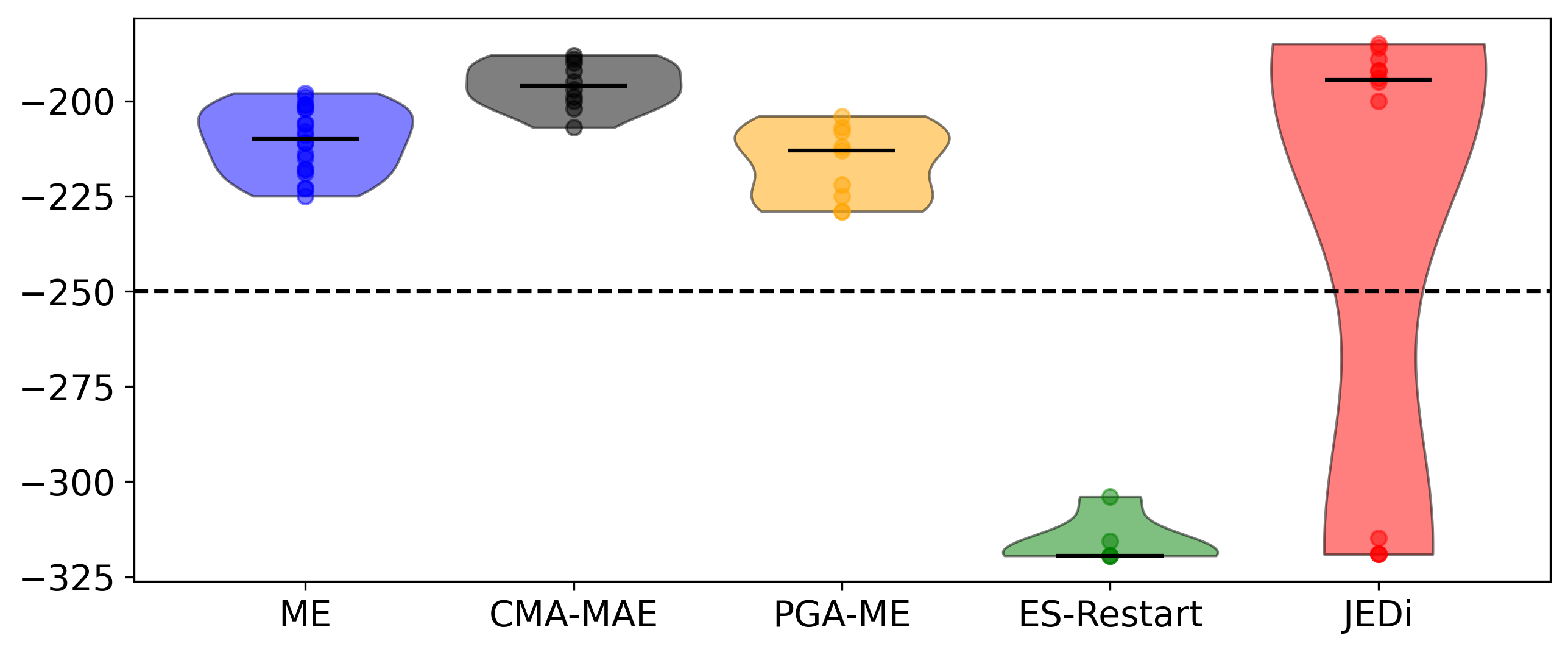} }}%

	\caption{Convergence and final fitness distribution on the quadrupled maze problem.}
        \vspace{-1.5em}
	\label{fig:quad_maze}
	\endgroup
\end{figure*}

We track the maximum fitness reached by each method: median convergence plots are presented in \Cref{plot:results}, with violin plots of final values in \Cref{plot:violin_results} and coverage evolution in \Cref{plot:coverage}. 
Figures present results for ES with restart, \mapelites{}, CMA-MAE, \pgame{} and \jedi{} with a fixed $\alpha$ value. 
To avoid crowding figures with 7 baselines we provide convergence plots for \cmame{} (outperformed by CMA-MAE) and ES without restart (equivalent or outperformed by ES with restart) in \Cref{app:sec:additional}, with final fitness values in \Cref{tab:u_test} and \Cref{tab:additional_u_test}. \Cref{tab:u_test} includes with Mann–Whitney U tests \cite{mwu_test} comparing final fitness distributions between the best \jedi{} version and each baseline. 

\subsubsection{Maze exploration}
\jedi{} with $\alpha$ both fixed and decaying outperforms all other baselines on mazes A, B and Quad B (\Cref{plot:results}, \Cref{fig:quad_maze}). On maze C with hard exploration only CMA-MAE reaches higher fitness values, but \jedi{} can still solve it. On all mazes \mapelites{} finds the target earlier, but the fitness plateaus as it struggles to optimize the policy to reach it faster while \jedi{} keeps improving. This is also highlighted by the coverage metric (\Cref{plot:coverage}) where QD methods cover the behavior space faster: \jedi{} explores as needed and focuses on behaviors that help get high fitness scores. 
U tests (\Cref{tab:u_test}) show small p-values when comparing \jedi{} to QD methods. 

Even with restarts the ES cannot solve mazes, stuck in local optima. With very low coverage it lacks the exploration capabilities of QD as the task requires to go around walls instead of trying to go through them to follow fitness gradients. 

\jedi{} shows a more bimodal distribution in final fitness values on mazes A, C and Quad B (\Cref{plot:violin_results}), which explains its high variance in convergence plots. In mazes, the \wtf{} function is also often deceptive for an ES when trying to reach a behavior on the other side of a wall and \jedi{} takes more time to explore the maze than QD methods. The targets however act as stepping stones, which leads \jedi{} to solve tasks a pure ES cannot solve. Learning the type of behaviors that lead to high fitness values then allows \jedi{} to constrain the search to policies that reach them, focusing the search and finding better policies. 
Using a linear scheduler to decay $\alpha$ for \jedi{} during the optimization (here from 0.8 to 0) makes it explore more at the beginning and then focus on fitness gradients, which is very effective in mazes which require exploration. This version outperforms fixed values of $\alpha$ on 3 of the 4 mazes while also removing an important hyperparameter.

\subsubsection{Robotics control}
On Brax control tasks \jedi{} largely outperforms gradient-free QD methods ($p \leq 0.02$) but also reaches higher scores than the ES, showing how using behavior information can help the search in problems where the behavior and the fitness are not naturally correlated (contrary to the maze). 
When tested on Brax tasks with large networks, \mapelites{}, \cmame{} and CMA-MAE struggle in the high-dimensional optimization problem. The RL method in \pgame{} helps it reach very high scores on \halfcheetah{} but not on the other tasks, while ES struggle on \halfcheetah{} but perform well on \walker{} and \antmaze{} (\Cref{plot:results}). 

QD methods get higher coverage than the ES on \halfcheetah{} and \walker{}, as expected. Similarly to what is observed in mazes, \jedi{} covers the behavior space more progressively as it focuses on specific search areas (\Cref{plot:coverage}). 
It is interesting to note that on \antmaze{} both ES and \jedi{} find policies that make the ant robot jump over the maze wall to reach the target without going through the maze, highlighting how evolution can find surprising solutions. This leads to higher fitness values than what QD can reach, and explains how they also reach higher coverage. 

The final archive of \jedi{} (\Cref{fig:budget_jedi:fitness}) can be qualitatively compared to \mapelites{} and ES (\Cref{fig:budget_baselines}). \mapelites{} finds low performing solutions (\ref{fig:budget_baselines:ME:fit}) but covers the behavior space uniformly with evaluations (\ref{fig:budget_baselines:ME:budget}, right) while the ES focuses most evaluations on one behavior but reaches higher scores (\ref{fig:budget_baselines:ES:fit}, \ref{fig:budget_baselines:ES:budget}). The \jedi{} archive shows a better coverage than ES, with budget focused on multiple areas including the one with the best scores (\ref{fig:budget_jedi:budget}). Behaviors with best fitnesses are slightly different from the ones mostly explored by the ES, showing exploration helped.

\def\figwidth{0.3\linewidth}
\def\xleg{Evaluations}
\def\yleg{Coverage (\%)}
\begin{table*}[h!]
	\centering
	\begin{tabular}{cccc}
		                                                                                         & Maze A         & Maze B    & Maze C     \\
		\raisebox{5\normalbaselineskip}[1cm][0cm]{\rotatebox[origin=c]{90}{\vspace{1cm}\yleg{}}} &
		\includegraphics[width=\figwidth]{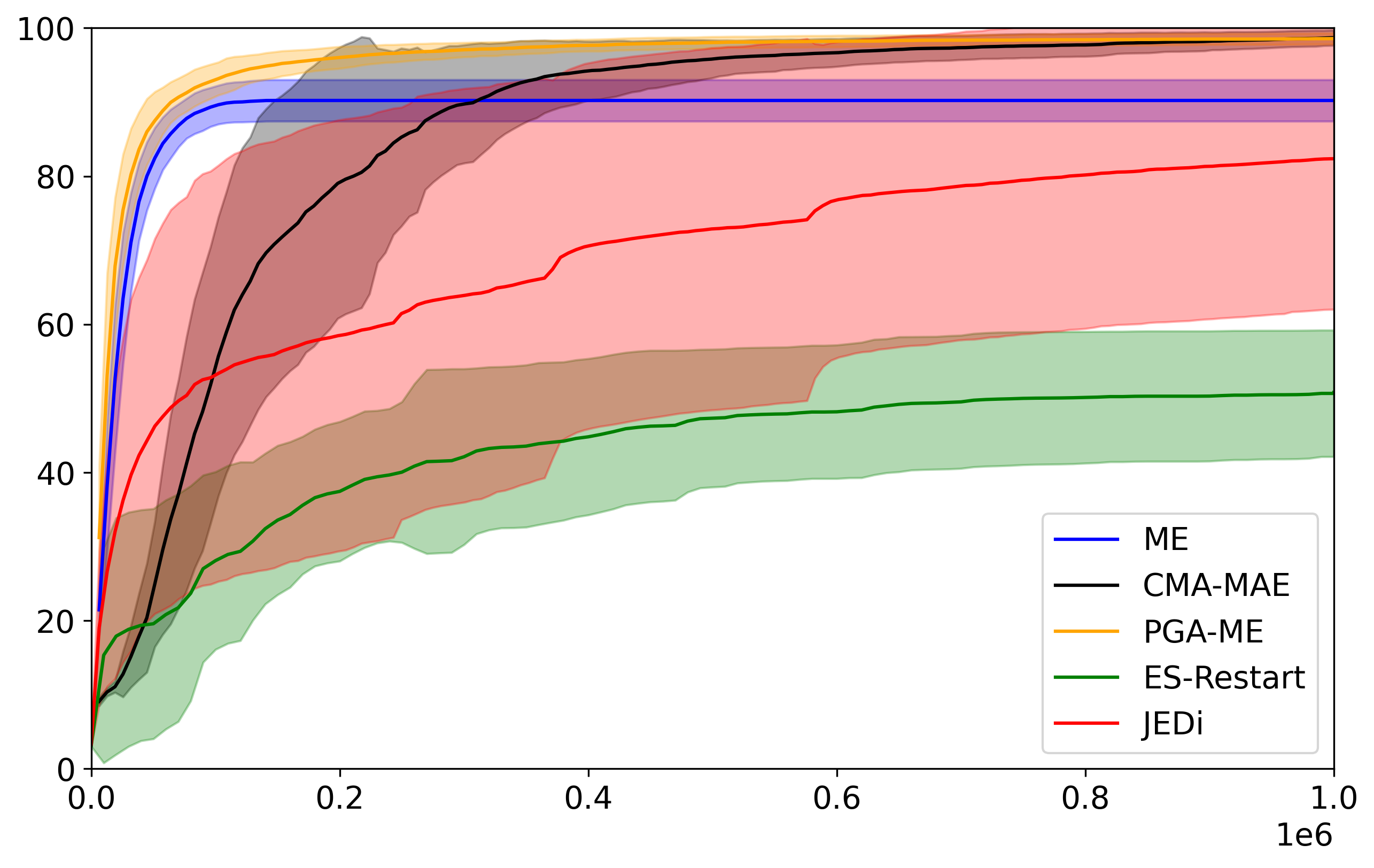}                 &
		\includegraphics[width=\figwidth]{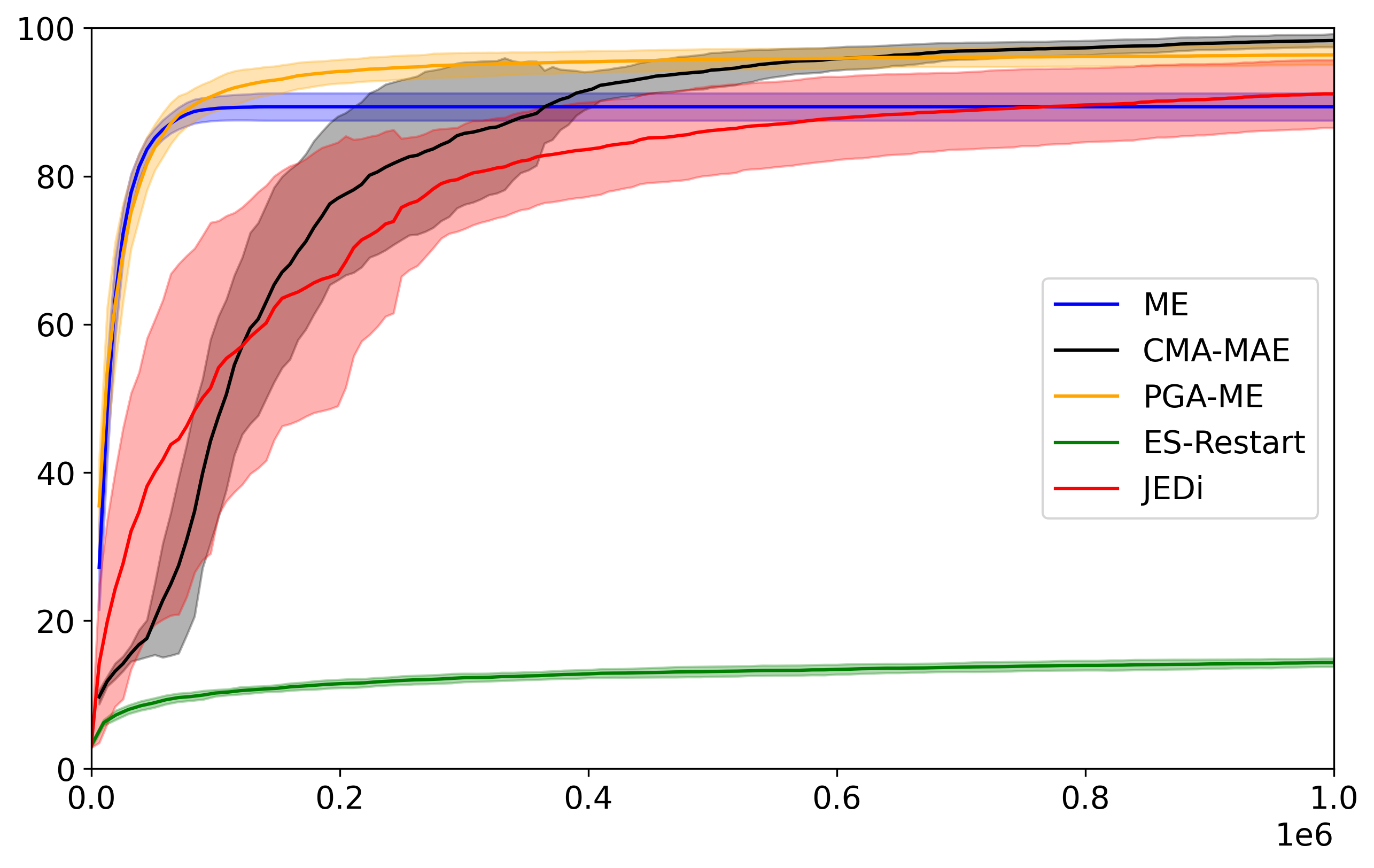}                     &
		\includegraphics[width=\figwidth]{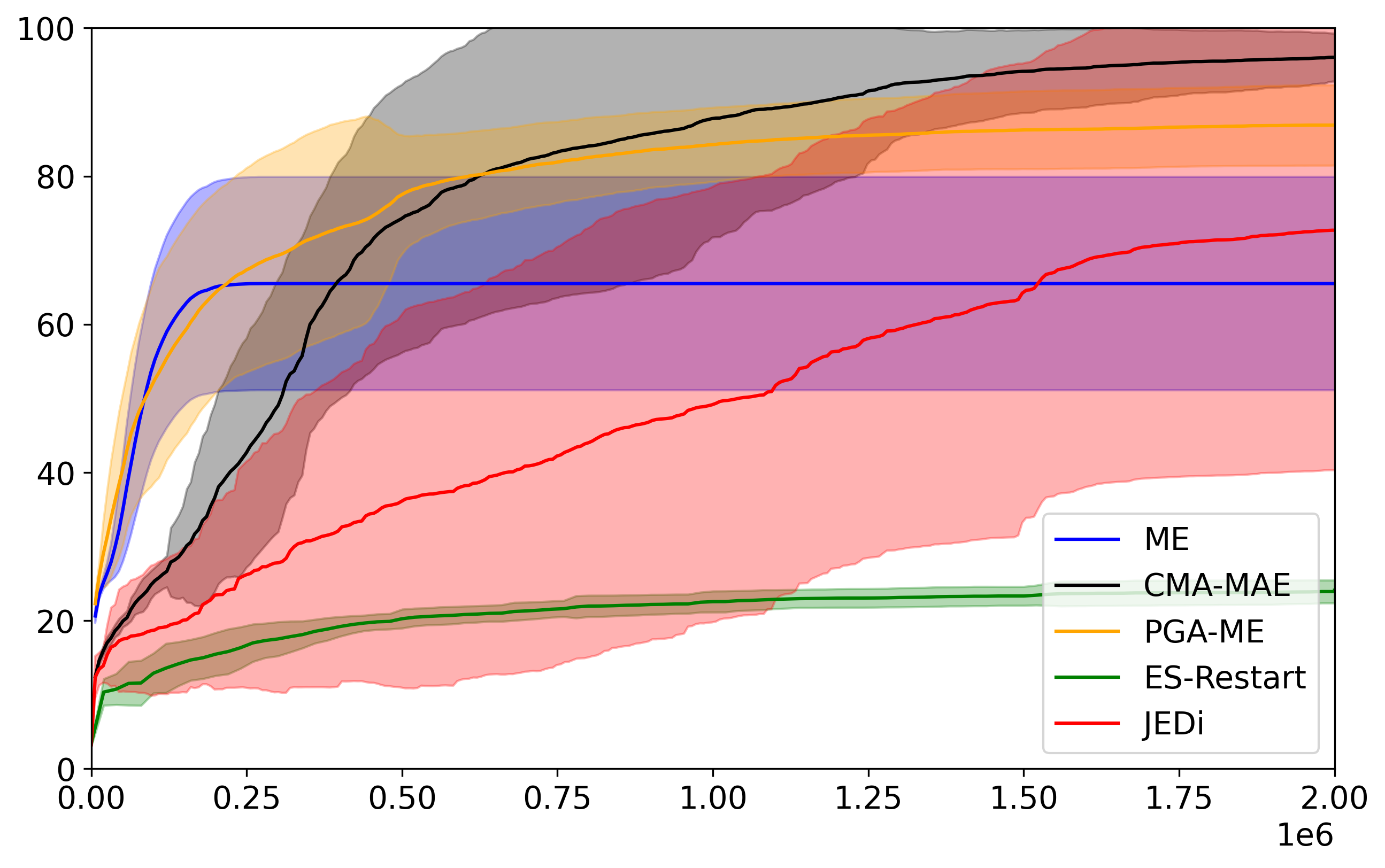}                                                            \\

		                                                                                         & \halfcheetah{} & \walker{} & \antmaze{} \\
		\raisebox{5\normalbaselineskip}[1cm][0cm]{\rotatebox[origin=c]{90}{\vspace{1cm}\yleg{}}} &
		\includegraphics[width=\figwidth]{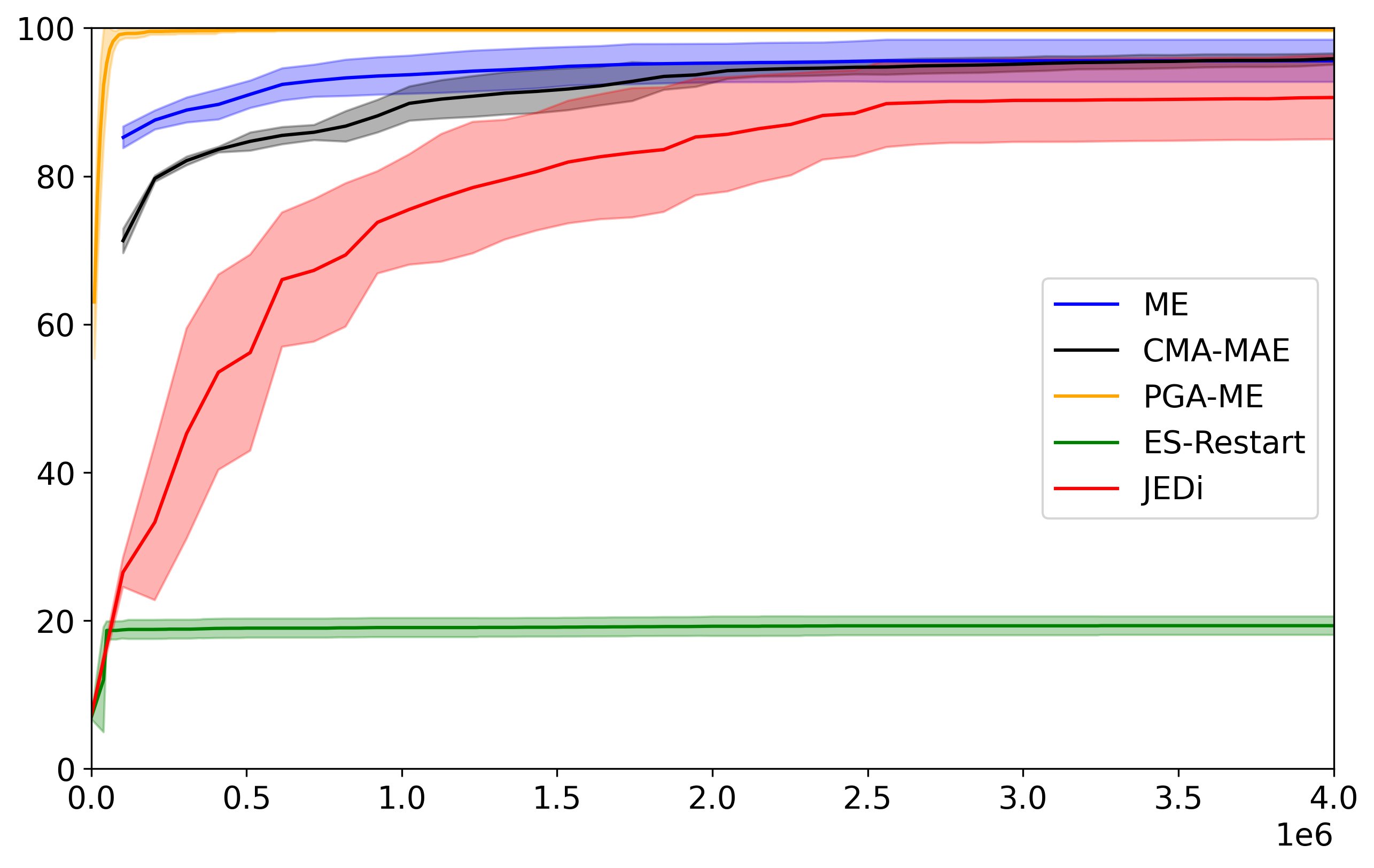}              &
		\includegraphics[width=\figwidth]{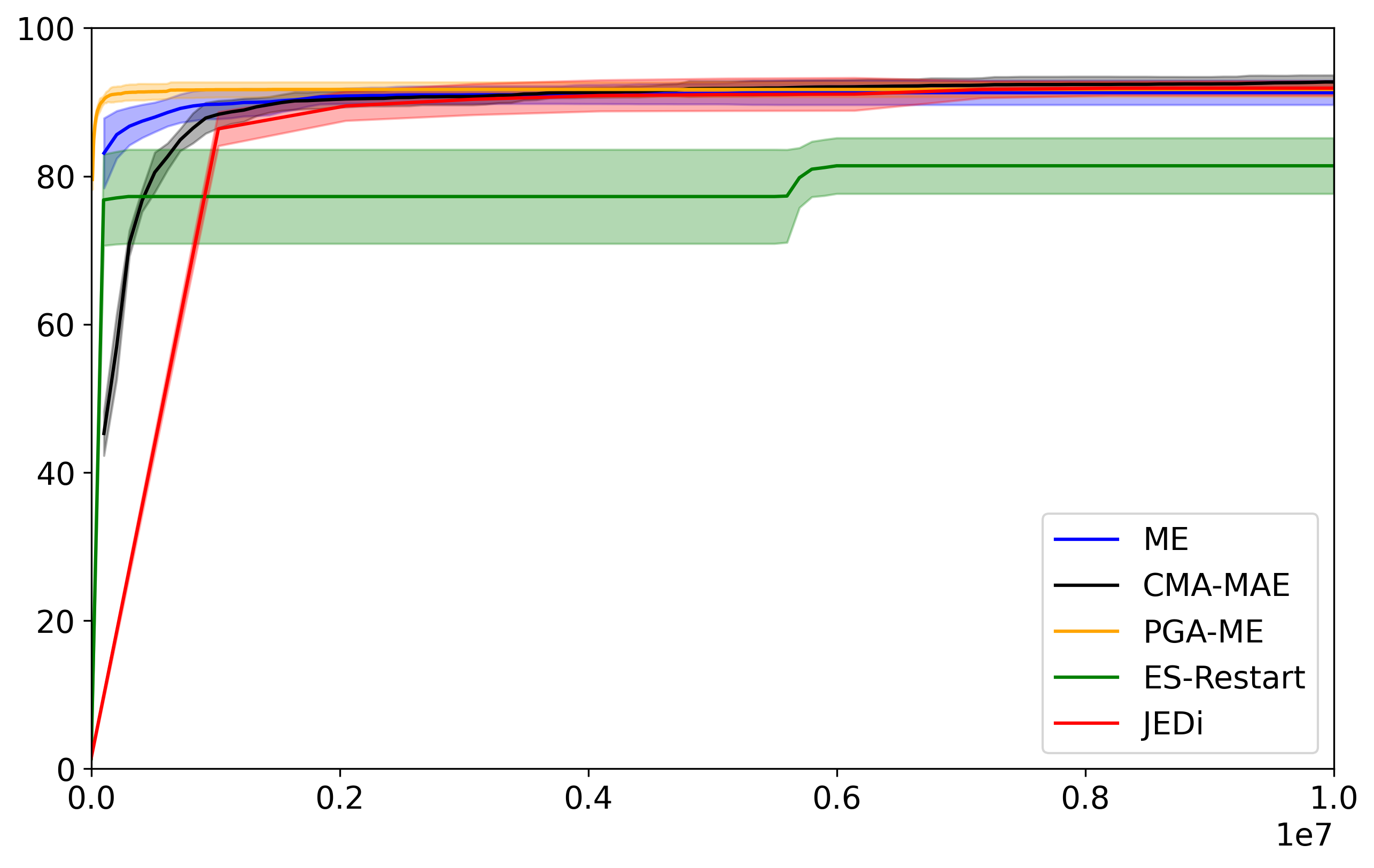}                &
		\includegraphics[width=\figwidth]{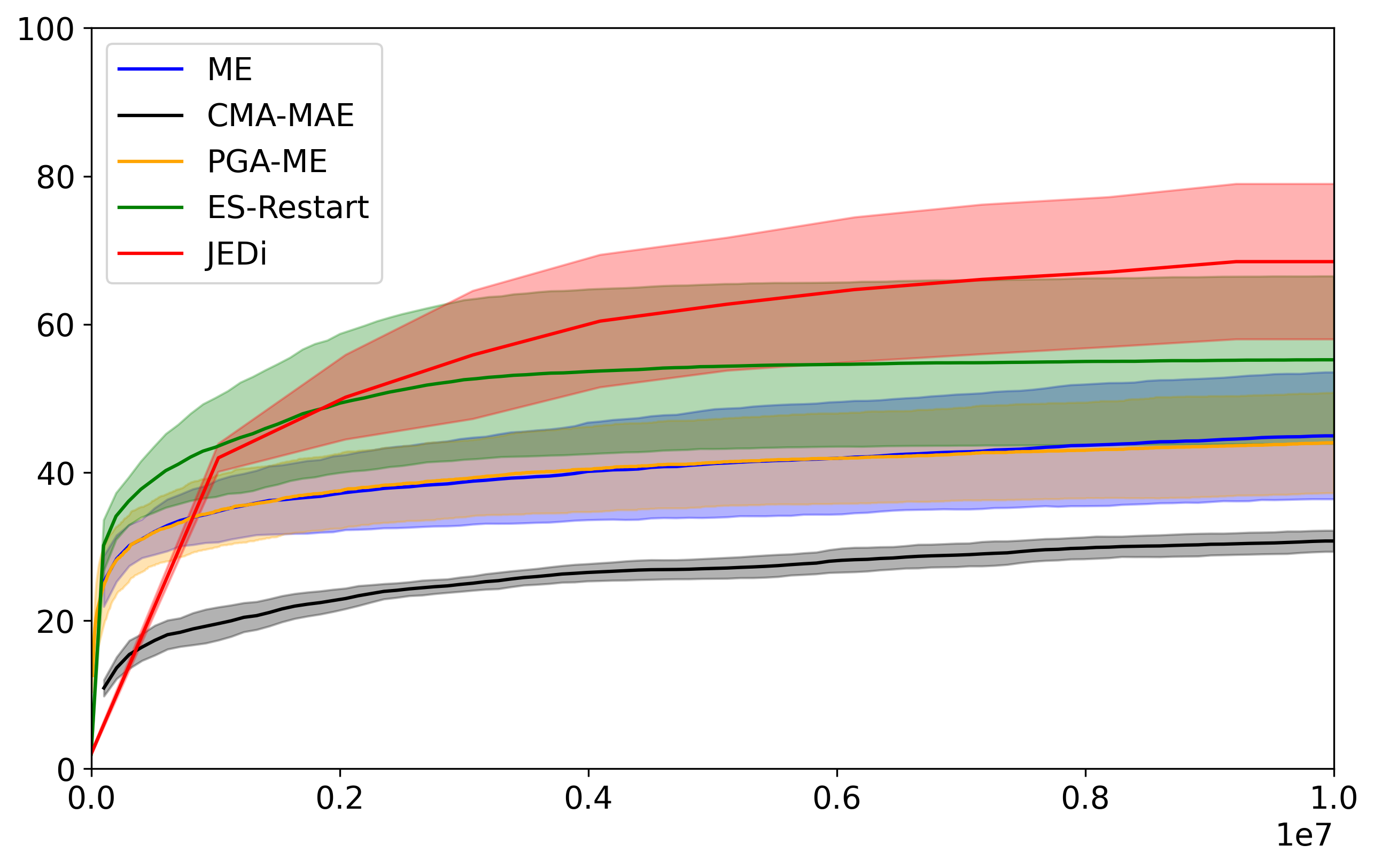}                                                                \\

		                                                                                         & \xleg          & \xleg     & \xleg
	\end{tabular}
	\captionsetup{type=figure}
	\caption{Coverage results for maze simple exploration (row 1) and robotics control tasks (row 2). }
	\label{plot:coverage}
\end{table*}

\subsection{Ablation studies}

We conduct ablation studies by changing parameters in isolation to better understand how some components of \jedi{} impact results. Complete results are available in \Cref{app:ablation}.

\def\figwidth{0.3\linewidth}
\def\xleg{Evaluations}
\def\yleg{Max fitness}
\begin{table*}
		\centering

		\begin{tabular}{cccc}
			                                                                                         & Maze B & Maze Quad B & \halfcheetah{} \\
			\raisebox{3\normalbaselineskip}[1cm][0cm]{\rotatebox[origin=c]{90}{\vspace{1cm}\yleg{}}} &
			\includegraphics[width=\figwidth]{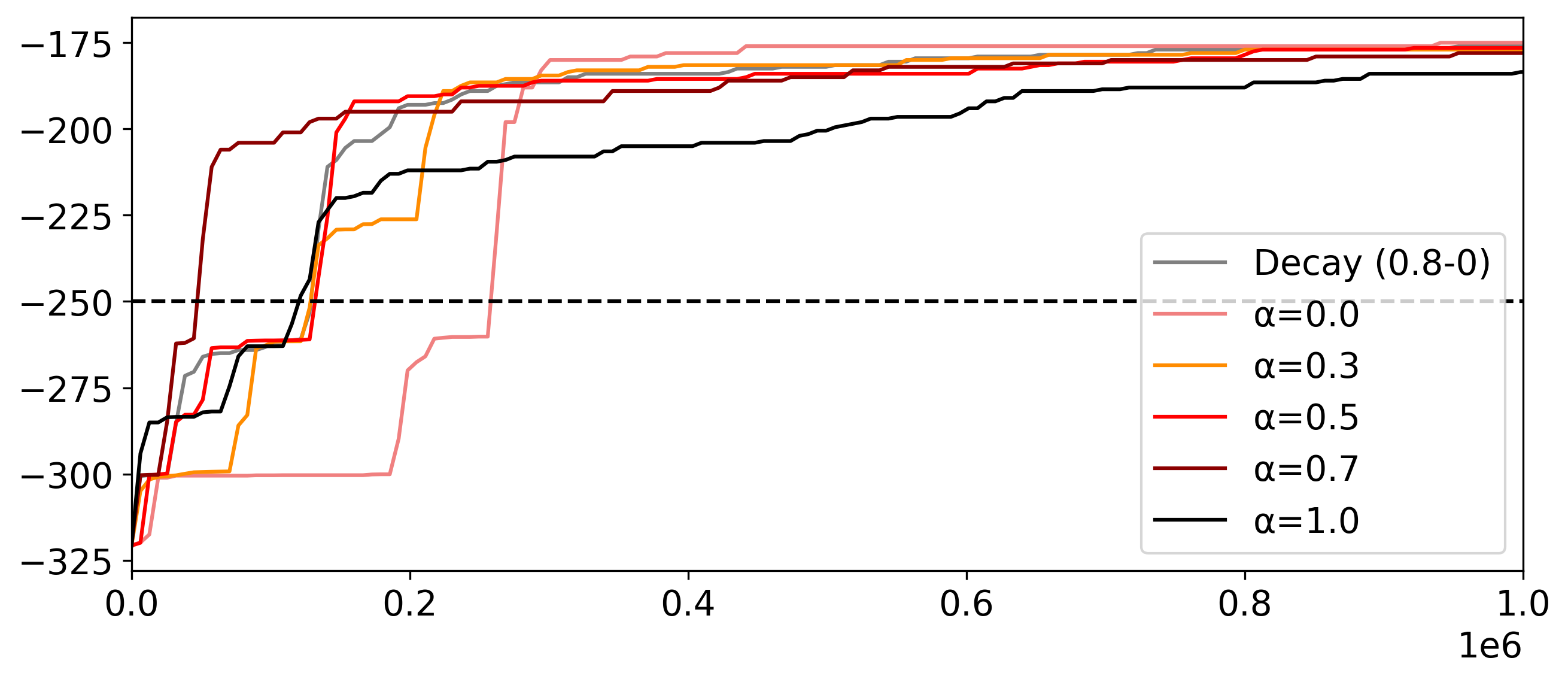}                         &
			\includegraphics[width=\figwidth]{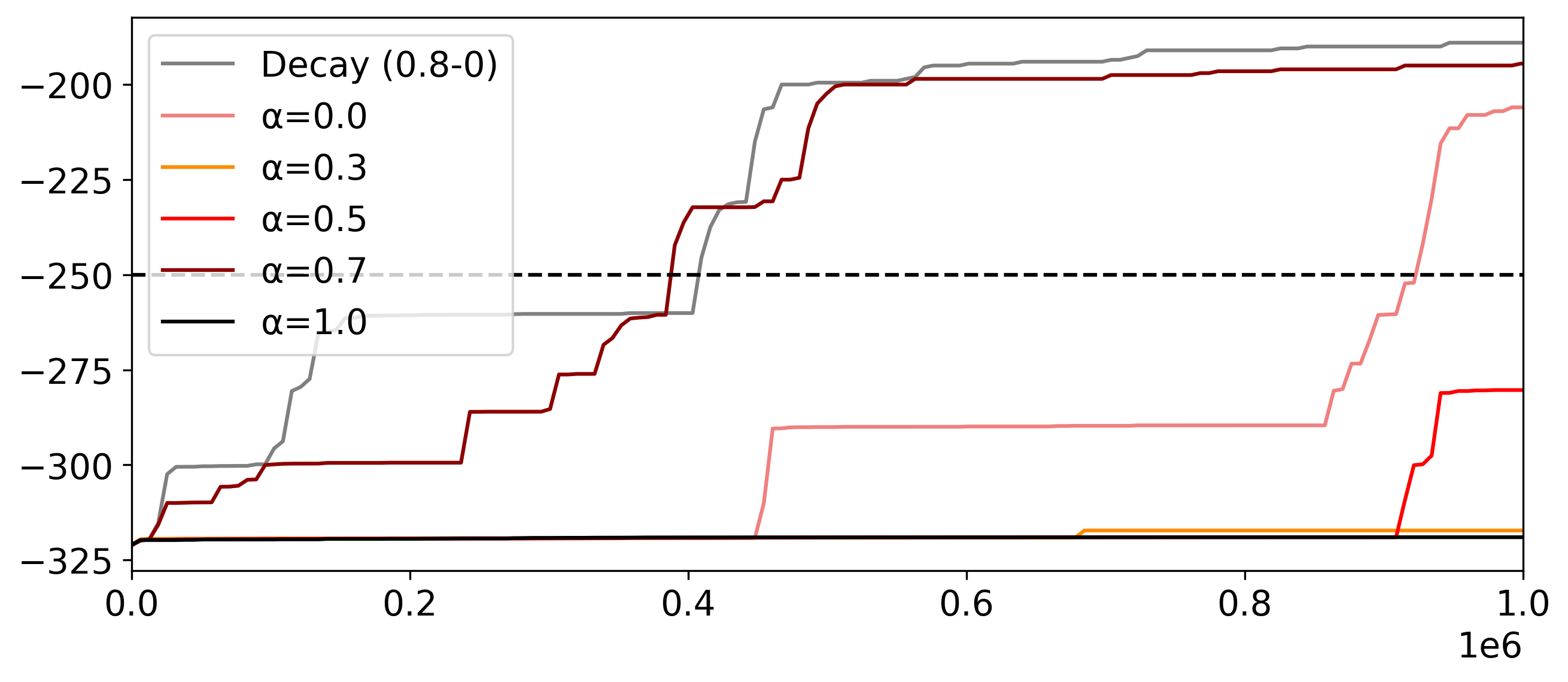}                    &
			\includegraphics[width=\figwidth]{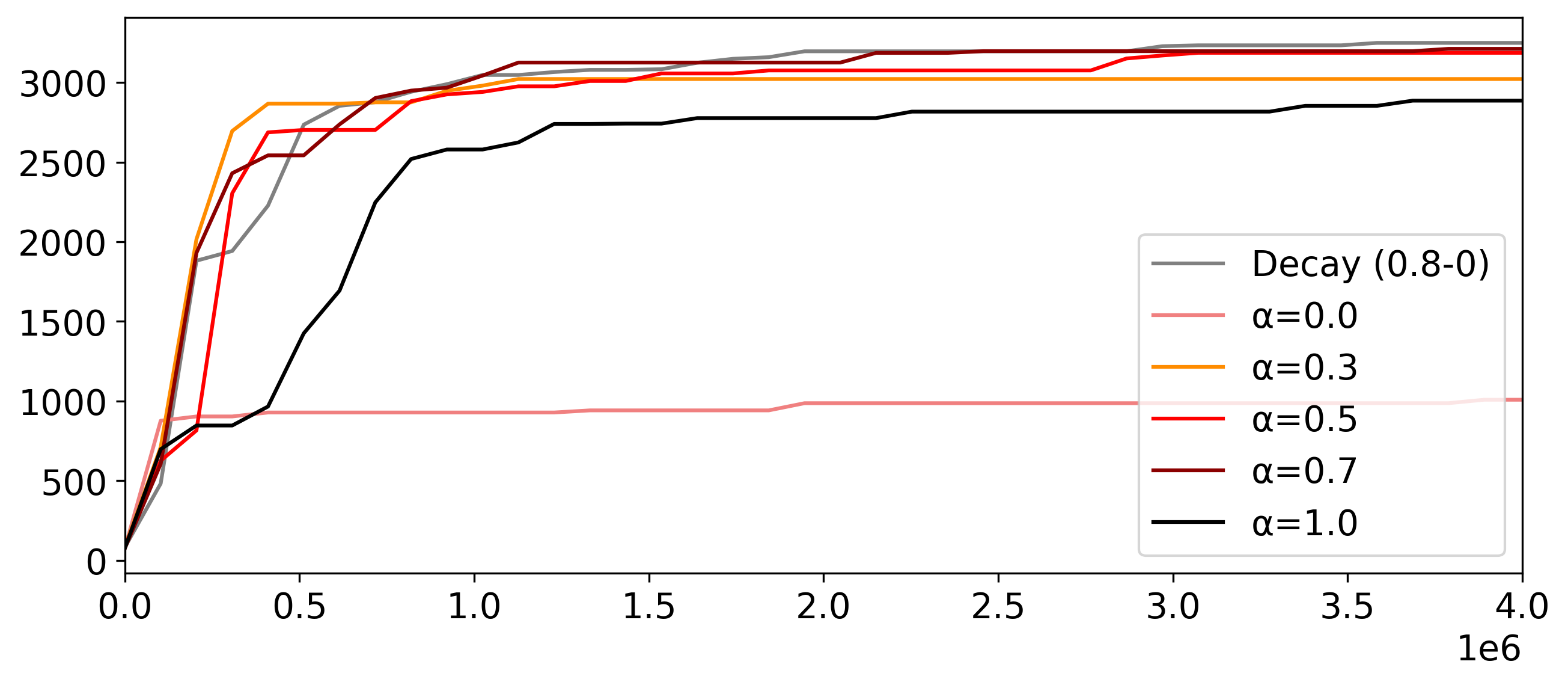}                                                       \\

			                                                                                         & \xleg  & \xleg       & \xleg \\

			                                                                                         & Maze A & Maze C & \antmaze{} \\
			\raisebox{5\normalbaselineskip}[1cm][0cm]{\rotatebox[origin=c]{90}{\vspace{1cm}\yleg{}}} &
			\includegraphics[width=\figwidth]{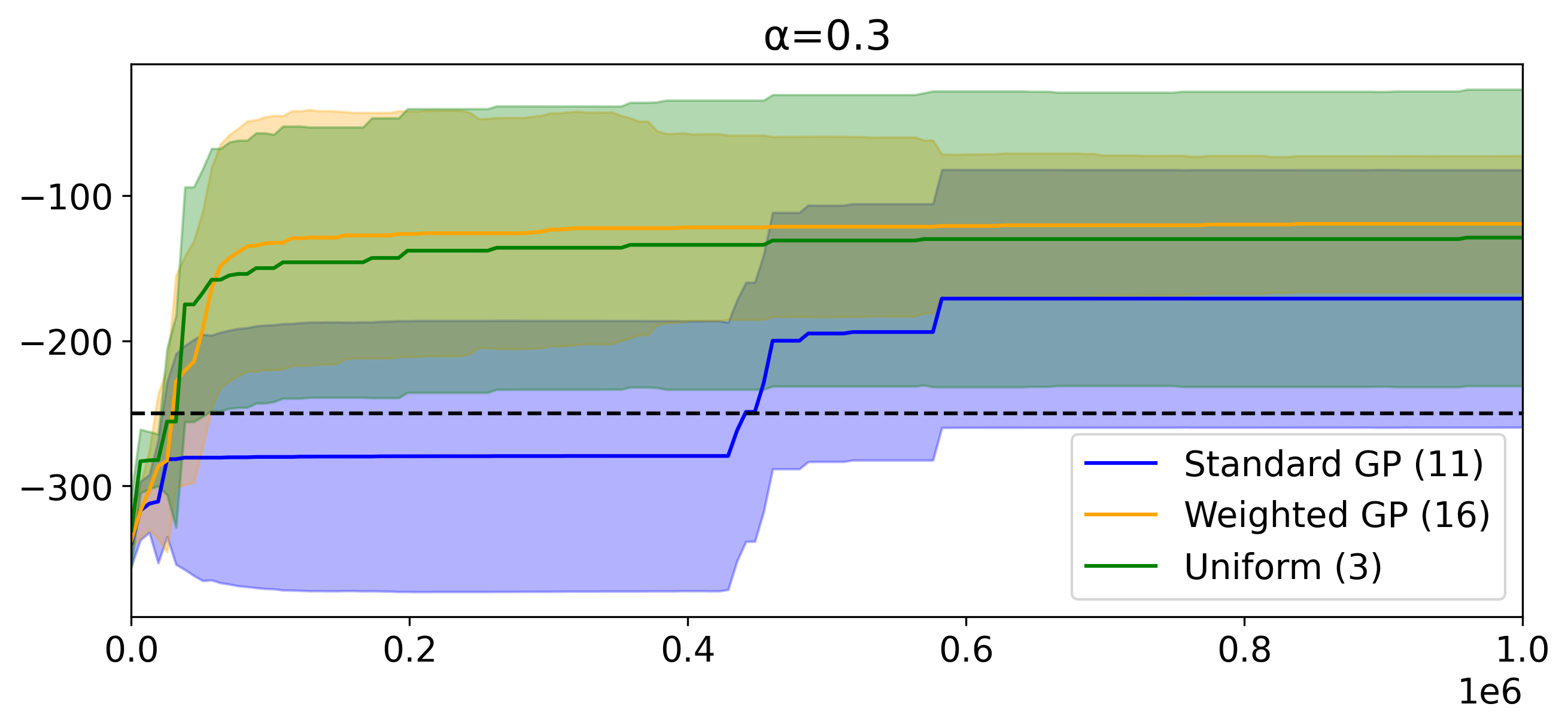}                 &
			\includegraphics[width=\figwidth]{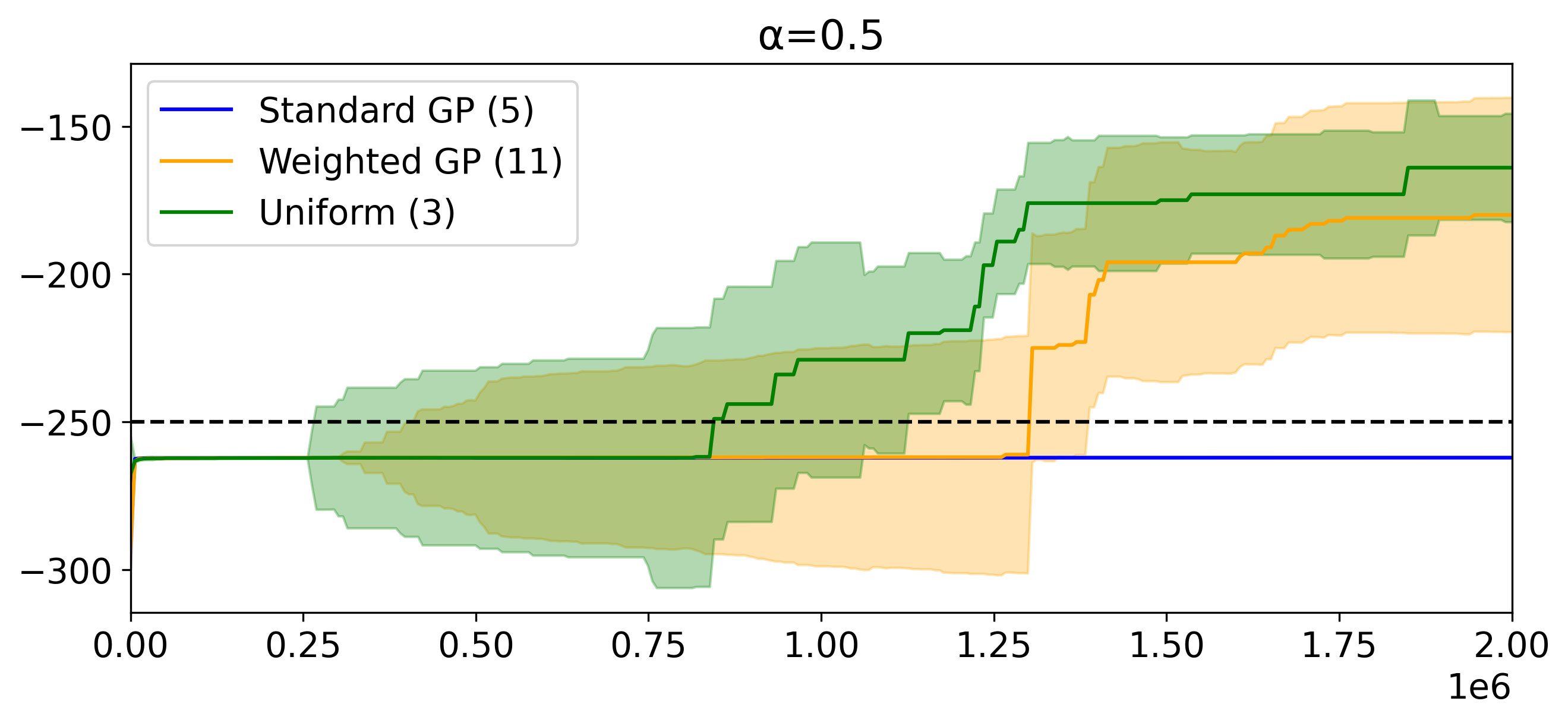}                     &
			\includegraphics[width=\figwidth]{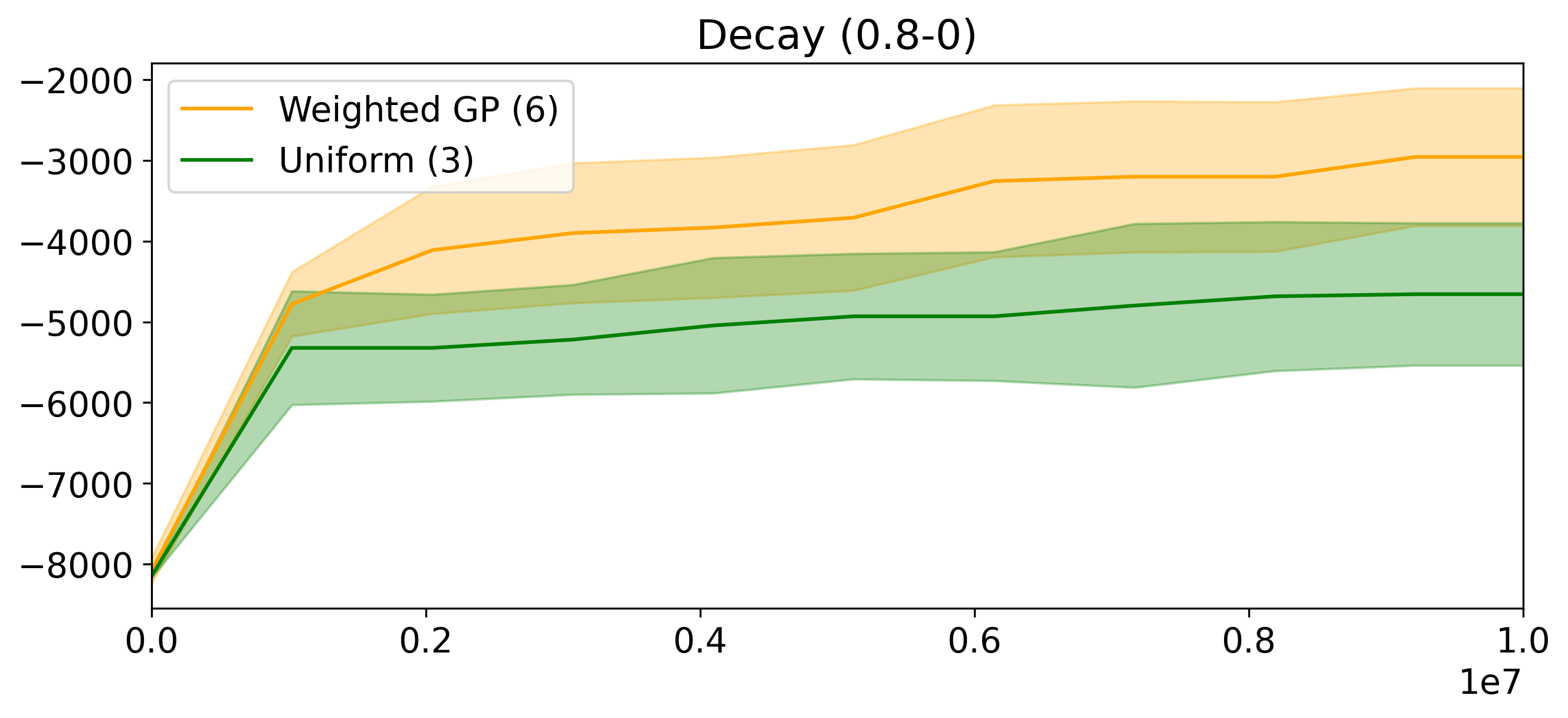}                                        \\

			                                                                                         & \xleg  & \xleg  & \xleg
		\end{tabular}

	\captionsetup{type=figure}
	\caption{Results from ablation studies on selected environments for \jedi{}:  $\alpha$ values (row 1) and weighted GP (row 2). Median fitness values with $\pm 1$ standard deviation (omitted in row 1 for clarity).}
	\label{plot:ablation}
\end{table*}

In \Cref{plot:ablation} we study the impact of the $\alpha$ parameter, which controls the importance of behavior targets in the \longwtf{}: A high $\alpha$ will focus on reaching the target and a low $\alpha$ will follow fitness gradients. On simpler mazes like Maze B all values of $\alpha$ allow to reach the target and the difference is small, but on harder exploration tasks like Maze Quad B the importance of reaching target behaviors grows and low values of $\alpha$ get stuck giving too much importance to fitness gradients. On less deceptive control tasks like \halfcheetah{}, most values can find high performing policies, except for $\alpha=0$ that struggles like the vanilla ES which shows behavior information does help in that case. 
This ablation highlights that \jedi{} is quite resilient to the choice of its parameter $\alpha$, although hard exploration tasks will require finer tuning. The other takeaway is that using a linearly decaying schedule for $\alpha$ performs well on all tasks,  providing a version of \jedi{} free from the $\alpha$ parameter. It even outperforming fixed values of $\alpha$ as seen in \Cref{tab:u_test}.

\jedi{} uses a Weighted Gaussian Process (WGP) (\Cref{sec:wgp}) to tackle the limitations of the standard GP and account for the exploration done in each behavior cell when sampling new targets. 
\Cref{plot:ablation} shows the WGP outperforms the non-weighted approach, as the information from the budget already used in an area helps pick better targets. The uniform selection method (no GP) performs well on some hard exploration tasks like Maze C, where a GP will tend to select targets far from explored areas and make the task harder for the \jedi{} ES. Uniformly sampling targets may provide targets easier to reach, hence better stepping stones in deceptive behavior landscapes. However on tasks like \antmaze{} less reliant on exploration (where ES are better than QD methods), WGP shows that using a GP drives the search to better performing policies.

\section{Conclusion and Future Work}
\label{sec:conclusion}

We introduce the framework of \longjedi{} to use the behavior information available in tasks to help optimize for maximum fitness like an ES. \jedi{} finds policies with higher fitness values than ES and QD methods on most tasks, on both hard exploration tasks like mazes and complex control with large policies.

Although \jedi{} is inspired from and compared to QD methods, it does not aim to replace them. \jedi{} only optimizes for the best policy and not for the quality of the final archive which can then be used for other purposes like adaptation. Compared to existing QD methods, \jedi{} can be used to get higher max fitness values, especially in gradient-free cases where methods with RL are not possible. Compared to ES methods, any environment where a behavior descriptor can be defined may benefit from \jedi{}, especially in deceptive fitness cases where new behaviors can unlock the ES.

Furthermore, the main features of CMA-MAE and \pgame{}, with the soft archive and the use of gradient-informed mutations respectively, could be used in tandem with the target selection and behavior mapping GP of \jedi{}. We leave the study of these combinations as future work. Furthermore, the target selection method could be adapted for the case of hard exploration, e.g. larger maze environments.

\begin{acks}
  This work was granted access to the HPC resources of CALMIP supercomputing center under the allocation P21001.
\end{acks}

\bibliographystyle{ACM-Reference-Format}
\bibliography{references_cleaned}

\appendix
\section{Implementation}
\label{app:implem}
All methods were implemented using the QDax framework \citep{lim2022accelerated} with all ES taken from EvoSax \citep{ evosax2022github} for standard implementations (including in \jedi{} and \cmame{}). ES used are \sepcmaes{} \citep{rosSimpleModificationCMAES2008} for maze tasks and \lmmaes{} \citep{loshchilov2017limited} for Brax tasks due to the quadratic memory cost of \sepcmaes{} in EvoSax reaching GPU memory limits with large genomes. Each policy is evaluated only once. In all problems we keep the stochasticity of the environment to a minimum by setting fixed initial states specific to each run. While ES can be robust to fitness noise, QD is sensitive to uncertainties on both fitness and behavior \citep{flageatFastStableMAPElites2020, flageatUncertainQualityDiversityEvaluation2023} and we keep the analysis of \jedi{} in uncertain problems to future work.

Gaussian processes use a Radial Basis Function (RBF) kernel re-implemented to include weighted GPs. Research code will be made public after review.

Hyperparameters for all experiments are shown in Tables \Cref{tab:maze_hp} and \Cref{tab:brax_hp}. CMA-MAE uses the same parameters as \cmame{}, with an archive learning rate of 0.1 picked as the best of $[10^{-1}, 10^{-2}, 10^{-3}]$. Values of 0 and 1 were tested too to verify they behaved like ES and \cmame{} as supposed. 

\Cref{tab:jedi_hp} reports which $\alpha$ value was used for each task in \jedi{} and how many generations each \jedi{} ES was run for before sampling a new set of targets.

For results presented in \Cref{tab:u_test} each method was tested on at least 10 random seeds for mazes and 5 for Brax task, with additional runs to improve the statistical significance of our results. The number of runs is not exactly the same for all methods due to a memory leak from calling the  \texttt{jax.jit} function many times, which lead some experiments to fail.

\begin{table}[h]
	\centering
	\begin{tabular}{ccc}
\toprule
Task &$\alpha$ &  ES steps \\
\midrule
Maze A         &   0.3 &       100 \\
Maze B         &   0.3 &       100 \\
Maze C         &   0.5 &       100 \\
Maze Quad B    &   0.7 &       100 \\
\halfcheetah{} &   0.5 &       100 \\
\walker{}      &   0.3 &      1000 \\
\antmaze{}     &   0.0 &      1000 \\
\bottomrule
\end{tabular}

	\caption{\jedi{} hyperparameters for maze and Brax tasks.}
	\label{tab:jedi_hp}
\end{table}

\begin{table}[h]
	\centering
	\begin{tabular}{c|c|c}
    \toprule
        & Mazes & Brax \\
    \midrule

    Critic hidden layers & $1 \times 16$ &  $2 \times 256$\\
    Batch size & 64 & 100 \\
    Iso sigma  & 0.2 & 0.005 \\
    Line sigma  & 0 & 0.05 \\   

    \midrule

    PG mutation steps & \multicolumn{2}{c}{10} \\
    Critic steps & \multicolumn{2}{c}{300} \\
    
    PG learning rate & \multicolumn{2}{c}{$10^{-3}$} \\
    Critic learning rate & \multicolumn{2}{c}{$3.10^{-4}$} \\

    Discount factor & \multicolumn{2}{c}{0.99} \\
    Soft update $\tau$ & \multicolumn{2}{c}{0.005} \\

    GA mutation ratio & \multicolumn{2}{c}{0.5} \\
    Buffer size & \multicolumn{2}{c}{$10^6$} \\

    \bottomrule
    \end{tabular}

	\caption{Hyperparameters for \pgame{} in maze and Brax tasks, based on \citep{tjanakaApproximatingGradientsDifferentiable2022}.}
\end{table}

\section{Weighted Gaussian Process}
\label{app:wgp}

Given the evaluation budget $n_i$ of a cell with behavior $x_i$, its $\textbf{ii}$ weight in the diagonal weight matrix $W$ is defined as the inverse of its evaluations and then used in the GP prediction for the fitness of a behavior $x$ with mean $\mu(x)$ and variance $\sigma^2(x)$:

\begin{align}
    W_{ii} &= \frac{1}{n_i} \\
    \mu(x) &= k(x, X) (K + \sigma_n^2 W)^{-1} y \\
    \sigma^2(x) &= k(x, x) - k(x, X) (K + \sigma_n^2 W)^{-1} k(X, x)
\end{align}

with $k$ the kernel function, $X$ the set of all evaluated behaviors, $K$ the covariance matrix of $X$ and $y$ the vector of fitness values of $X$. $\sigma_n$ is the noise parameter of the GP, which is optimized with the kernel parameters.

\section{Additional baselines}
\label{app:sec:additional}

Additional baselines are presented in \Cref{tab:additional_u_test}: \cmame{} with two emitters, ES with and without restart, and CMA-MAE for reference in convergence plots (\Cref{app:plot:additional_results}). 

\begin{table}[h]
	\centering
	\begin{tabular}{l|cccc}
\toprule
          Task & \makecell{CMA-ME\\Imp} & \makecell{CMA-ME\\Opt} &    ES & \makecell{ES\\Restart} \\
\midrule
        Maze A &         -138 &         -132 &  -281 &       -280 \\
        Maze B &         -190 &         -197 &  -319 &       -320 \\
        Maze C &         -236 &         -262 &  -262 &       -262 \\
   Maze Quad B &         -219 &            - &  -319 &       -319 \\
\halfcheetah{} &         1382 &         1260 &  1239 &       1237 \\
     \walker{} &         2159 &         2098 &  3967 &       3995 \\
    \antmaze{} &        -5145 &        -5280 & -3575 &      -2892 \\
\bottomrule
\end{tabular}

	\caption{Median final max fitness for additional baselines.}
	\label{tab:additional_u_test}
\end{table}

\begin{table*}[h]
	\begin{subtable}{0.45\textwidth}
		\centering
		\begin{tabular}{c|c|c|c|c}
    \toprule
        & \jedi{}  & ES  &      CMA-ME & \mapelites{}\\
    \midrule
    Hidden layers  & \multicolumn{4}{c}{$1 \times 8$} \\
    Hidden activation  & \multicolumn{4}{c}{relu} \\
    Final activation  & \multicolumn{4}{c}{tanh} \\
    Centroids  & \multicolumn{4}{c}{1024} \\
    
    \midrule
    
    Iso sigma  & \multicolumn{3}{c|}{-} & 0.2 \\
    Line sigma  & \multicolumn{3}{c|}{-} & 0 \\

    ES  & \multicolumn{3}{c|}{\sepcmaes} & - \\
    ES $\sigma$ init & \multicolumn{3}{c|}{0.05} & - \\
    ES elite ratio & \multicolumn{3}{c|}{0.5} & - \\
    
    Population per ES  & 16 & 64 &  16 & - \\
    ES number  & 4 & 1  & 4 & -\\
    Total batch size  & 64 & 64 & 64 & 64 \\
    
    \bottomrule
    \end{tabular}

		\caption{Maze tasks}
		\label{tab:maze_hp}
	\end{subtable}
	\hfill
	\begin{subtable}{0.45\textwidth}
		\centering
		\begin{tabular}{c|c|c|c|c}
\toprule
    & \jedi{}  & ES  &      CMA-ME & \mapelites{}\\
\midrule
Hidden layers  & \multicolumn{4}{c}{$2 \times 256$} \\
Hidden activation  & \multicolumn{4}{c}{tanh} \\
Final activation  & \multicolumn{4}{c}{tanh} \\
Centroids  & \multicolumn{4}{c}{1024} \\

\midrule
Iso sigma  & \multicolumn{3}{c|}{-} & 0.005 \\
Line sigma  & \multicolumn{3}{c|}{-} & 0.05 \\

ES  & \multicolumn{3}{c|}{\lmmaes} & - \\
ES $\sigma$ init & \multicolumn{3}{c|}{0.05} & - \\
ES elite ratio & \multicolumn{3}{c|}{0.5} & - \\

Population per ES  & 256 & 256 &  256 & - \\
ES number  & 4 & 1  & 4 & -\\
Total batch size  & 1024 & 256 & 1024 & 1024 \\

\bottomrule
\end{tabular}

		\caption{Brax tasks}
		\label{tab:brax_hp}
	\end{subtable}

	\caption{Global hyperparameters for maze and Brax tasks.}
\end{table*}

\def\figwidth{0.3\linewidth}
\def\xleg{Evaluations}
\def\yleg{Max fitness}
\begin{table*}[t]
	\centering
	\begin{tabular}{cccc}
		                                                                                         & Maze A         & Maze B      & Maze C      \\
		\raisebox{5\normalbaselineskip}[1cm][0cm]{\rotatebox[origin=c]{90}{\vspace{1cm}\yleg{}}} &
		\includegraphics[width=\figwidth]{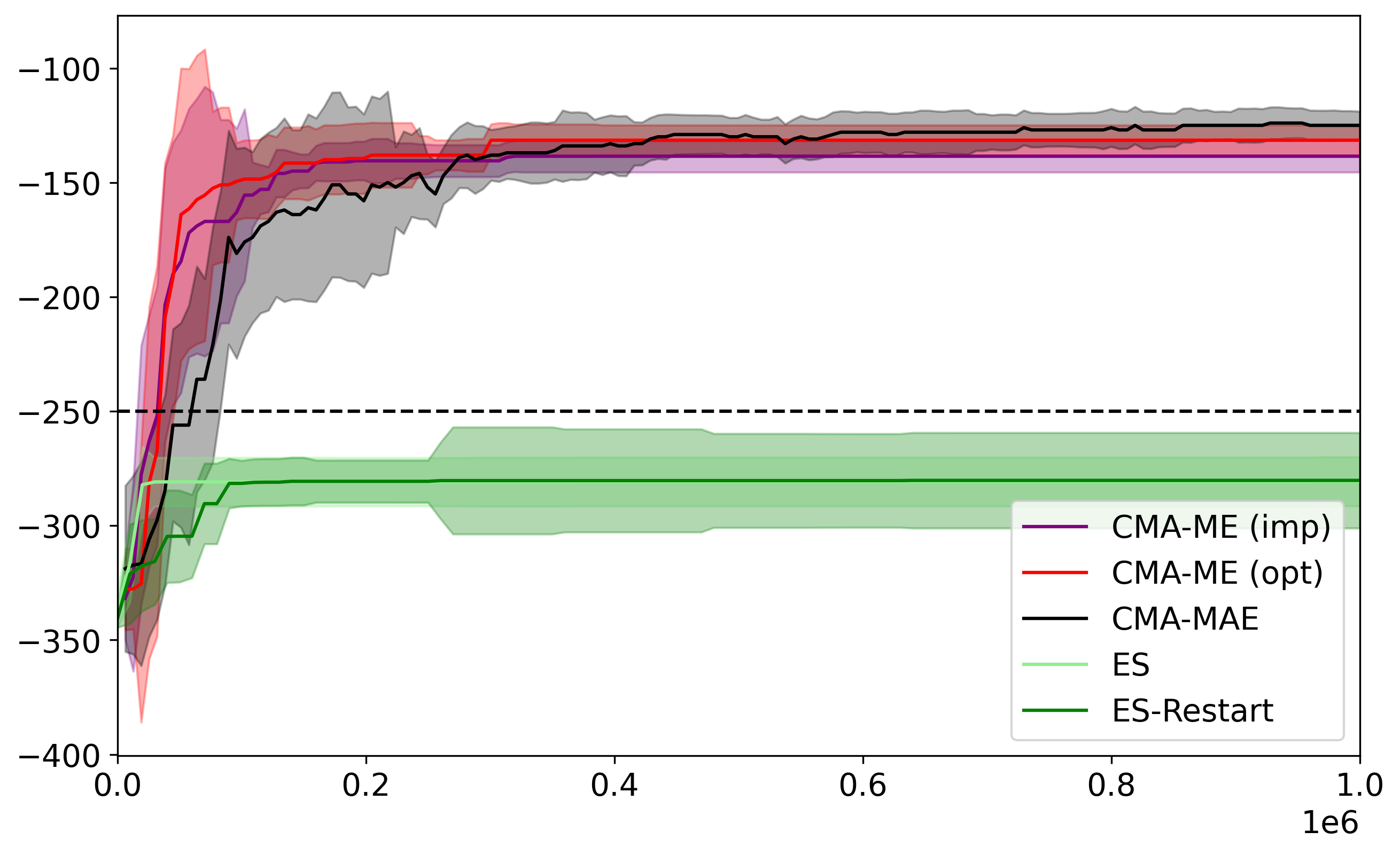}               &
		\includegraphics[width=\figwidth]{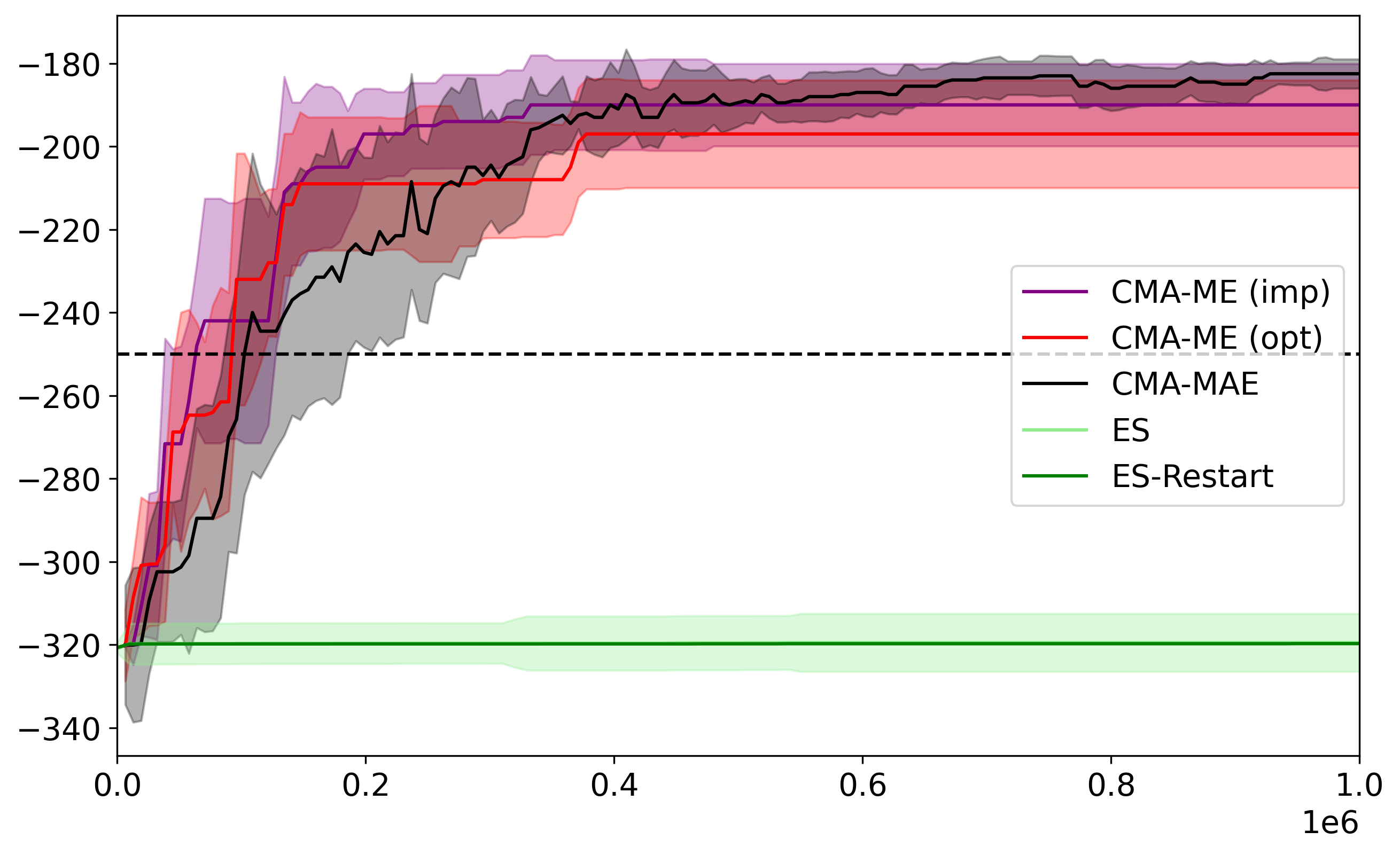}                   &
		\includegraphics[width=\figwidth]{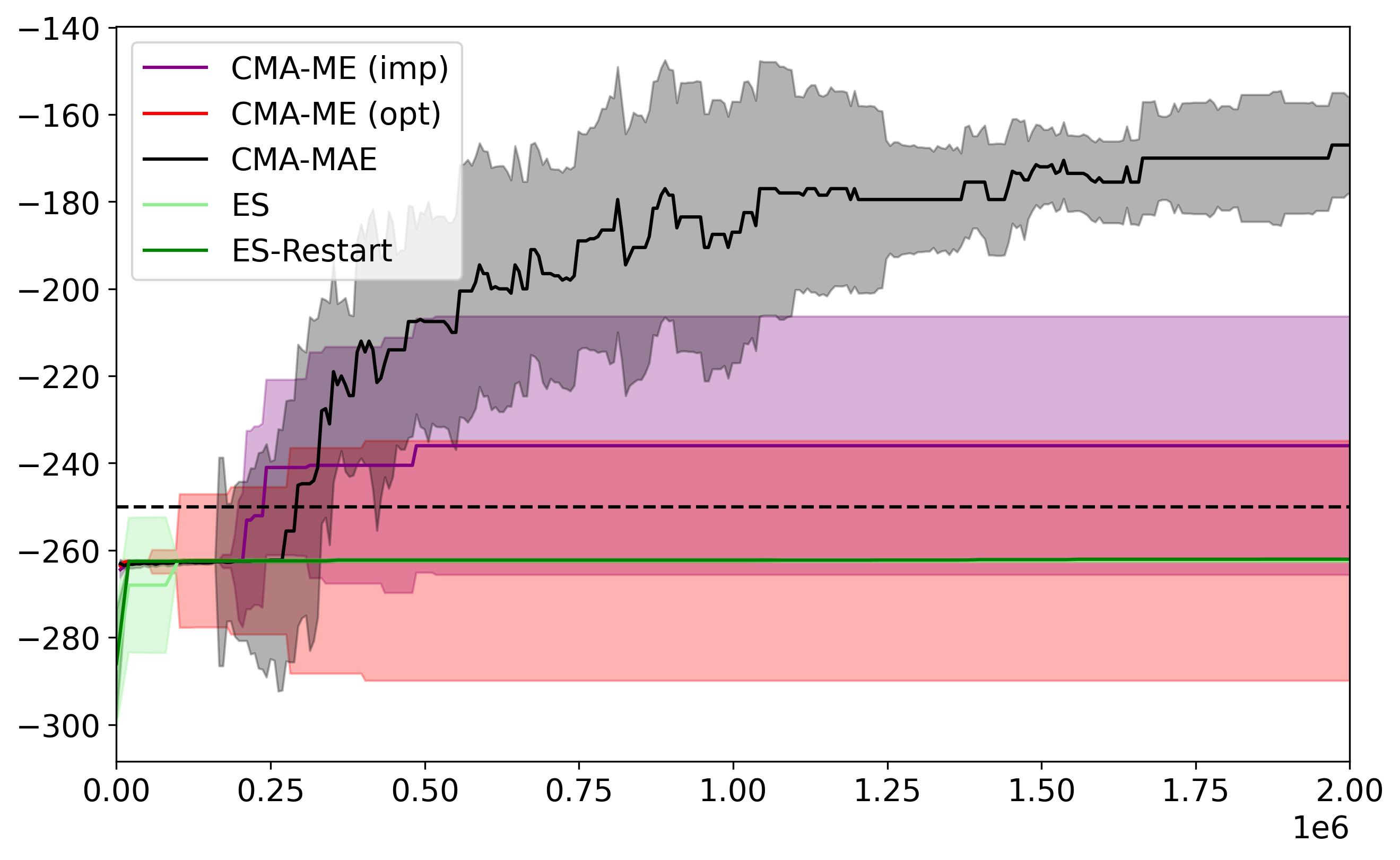}                                                             \\

		                                                                                         & \halfcheetah{} & \walker{}   & \antmaze{}  \\
		\raisebox{5\normalbaselineskip}[1cm][0cm]{\rotatebox[origin=c]{90}{\vspace{1cm}\yleg{}}} &
		\includegraphics[width=\figwidth]{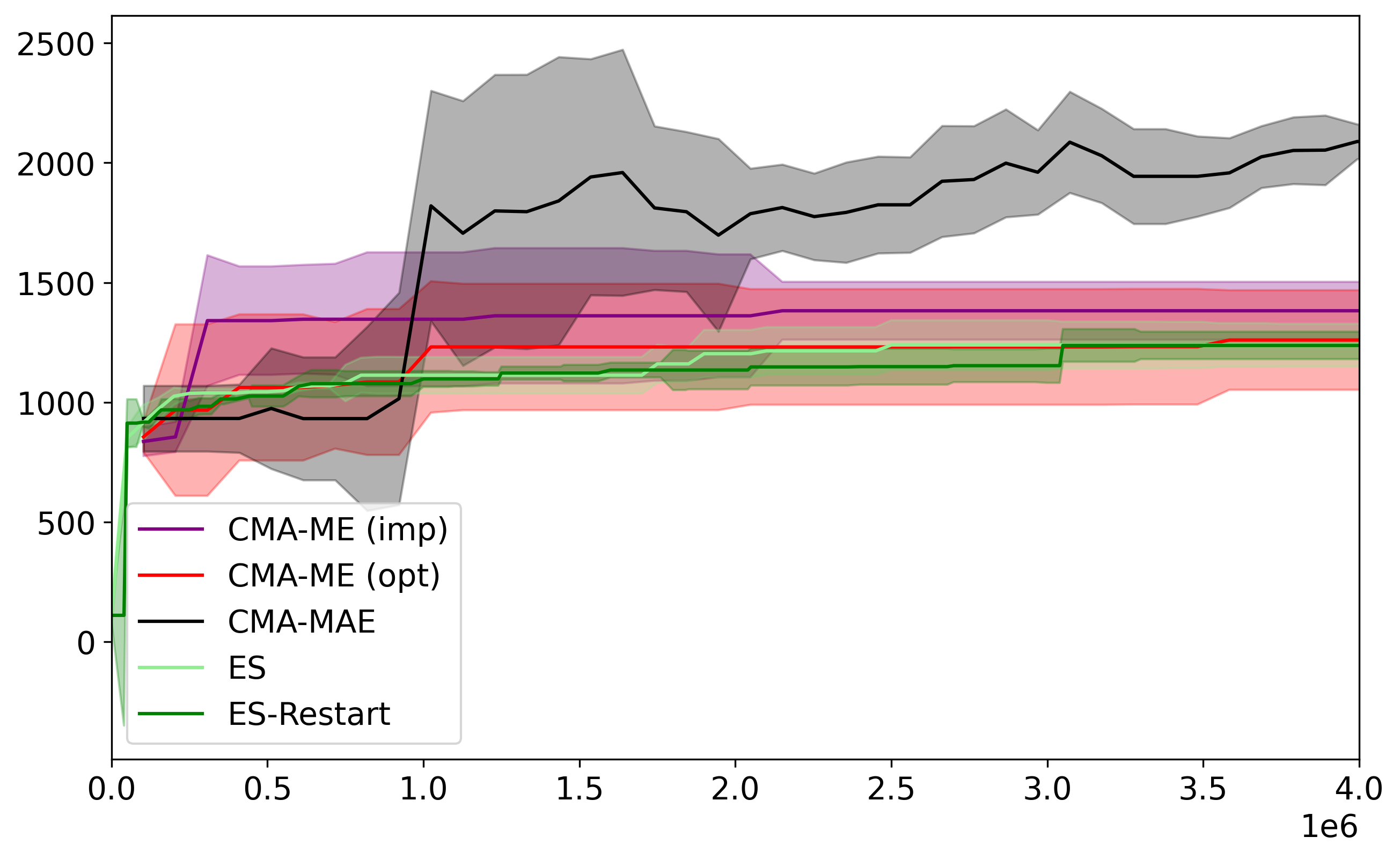}            &
		\includegraphics[width=\figwidth]{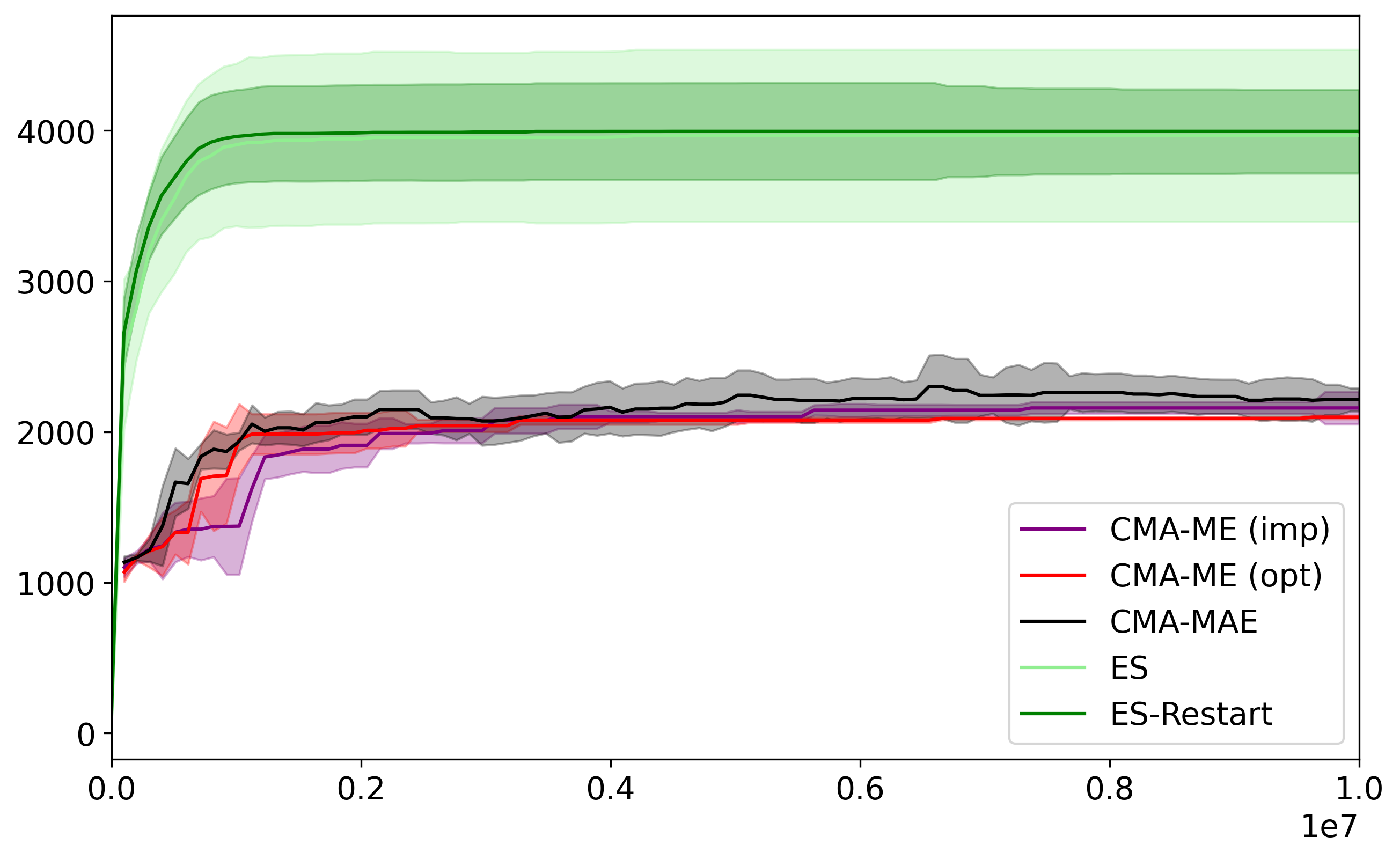}              &
		\includegraphics[width=\figwidth]{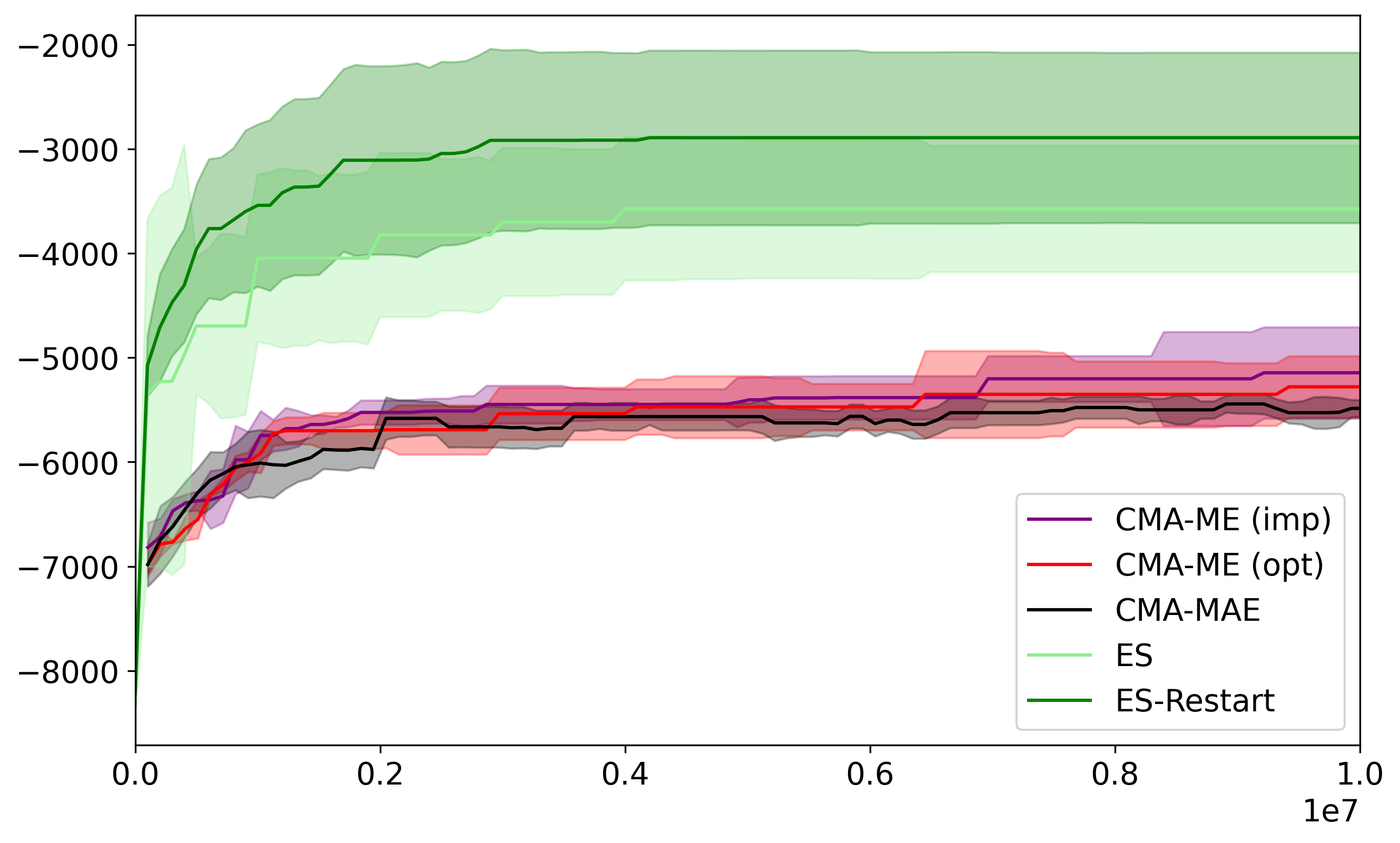}                                                                 \\

		                                                                                         & \xleg          & \xleg       & \xleg
	\end{tabular}
	\captionsetup{type=figure}
	\caption{Additional fitness results for maze simple exploration (row 1) and robotics control tasks (row 2). Reaching the dotted line at -250 in a maze means an agent has reached the target. Strong line is the median over all runs, with colored areas showing $\pm 1$ standard deviation over and under the average.}
	\label{app:plot:additional_results}
\end{table*}

\section{Ablation results}
\label{app:ablation}

This appendix section presents complete ablation results: $\alpha$ values in \Cref{app:plot:jedi_alpha}, Gaussian process used for sampling (\Cref{app:plot:jedi_wgp}) and an additional ablation on the number of parallel ES in \jedi{} (\Cref{app:plot:jedi_gens}) showing the method benefits from parallelized ES to reach multiple targets at the same time as it reaches higher scores.

\def\figwidth{0.3\linewidth}
\def\xleg{Evaluations}
\def\yleg{Max fitness}
\begin{table*}[t]
	\centering
	\begin{tabular}{cccc}
		                                                                                               & Maze A         & Maze B      & Maze C      \\
		\raisebox{3\normalbaselineskip}[1cm][0cm]{\rotatebox[origin=c]{90}{\vspace{1cm}\yleg{}}}       &
		\includegraphics[width=\figwidth]{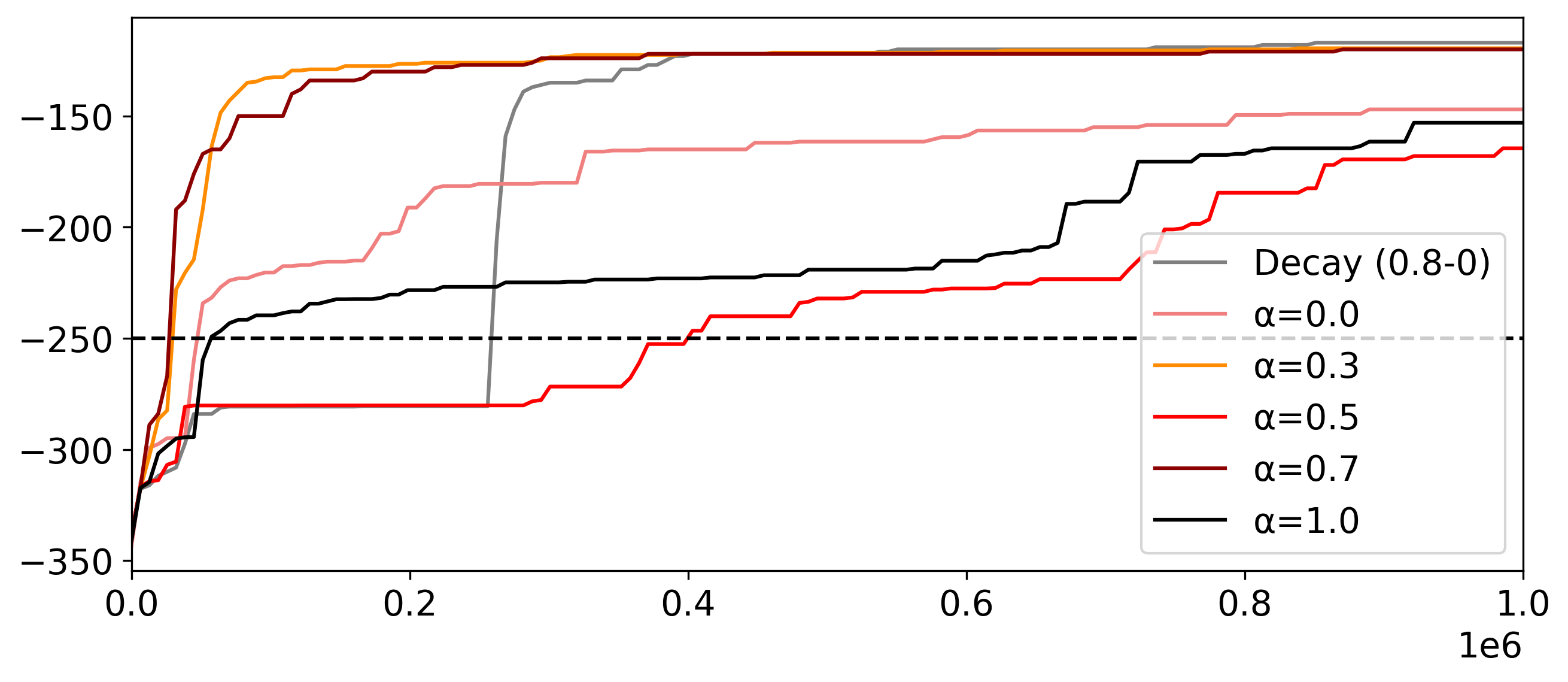}                           &
		\includegraphics[width=\figwidth]{plots/jedi_KH-snake-250-D.png}                               &
		\includegraphics[width=\figwidth]{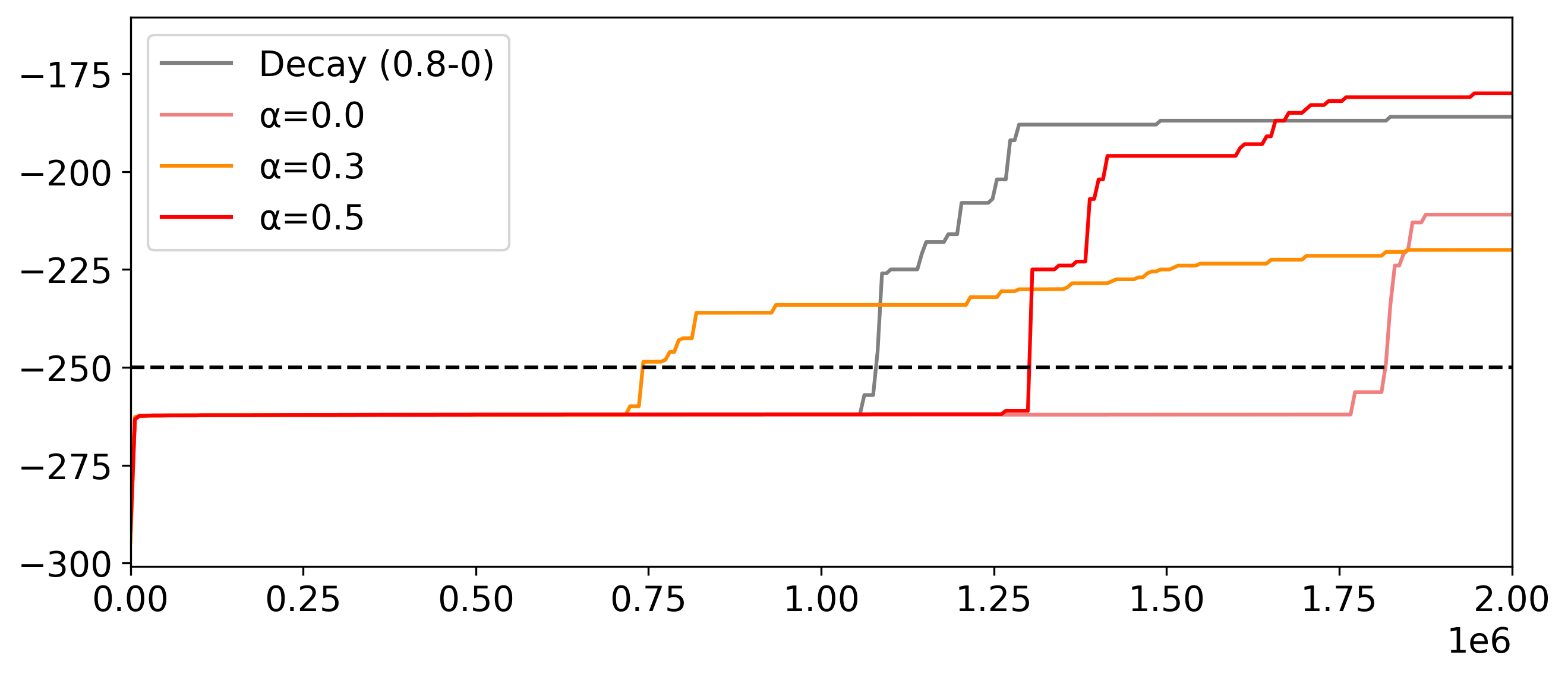}                                                                         \\

		\raisebox{3\normalbaselineskip}[1cm][0cm]{\rotatebox[origin=c]{90}{\vspace{1cm}Coverage (\%)}} &
		\includegraphics[width=\figwidth]{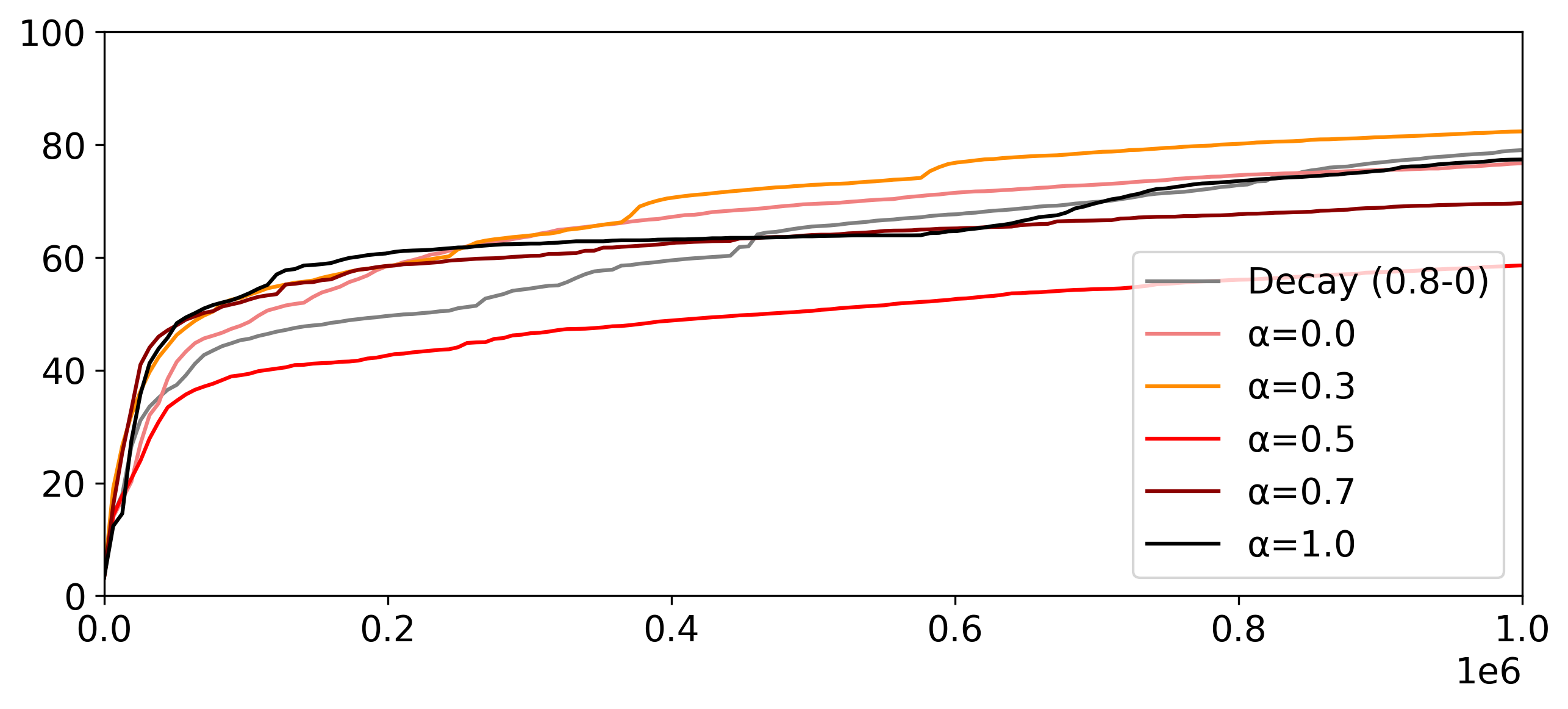}                  &
		\includegraphics[width=\figwidth]{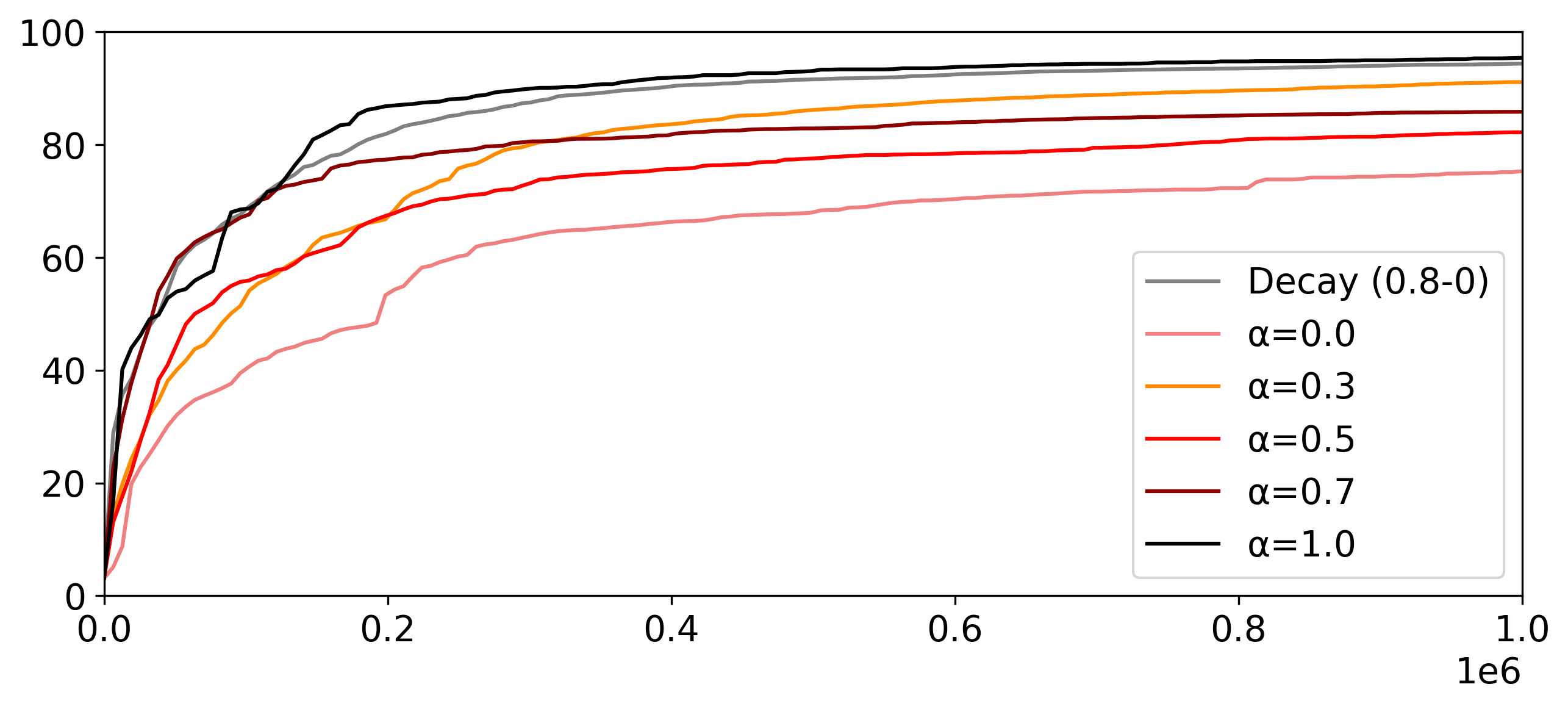}                      &
		\includegraphics[width=\figwidth]{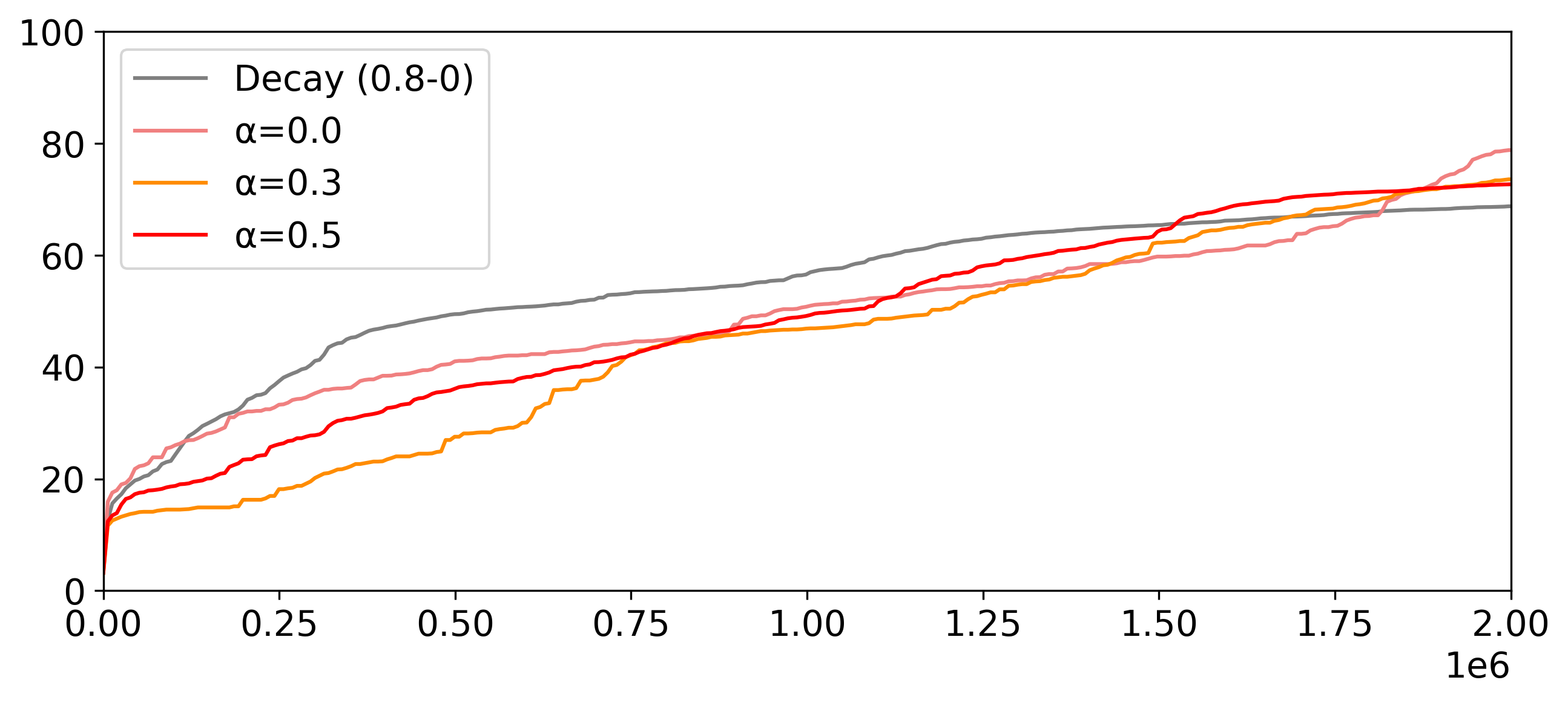}                                                                \\

		\raisebox{3\normalbaselineskip}[1cm][0cm]{\rotatebox[origin=c]{90}{\vspace{1cm}Final fitness}} &
		\includegraphics[width=\figwidth]{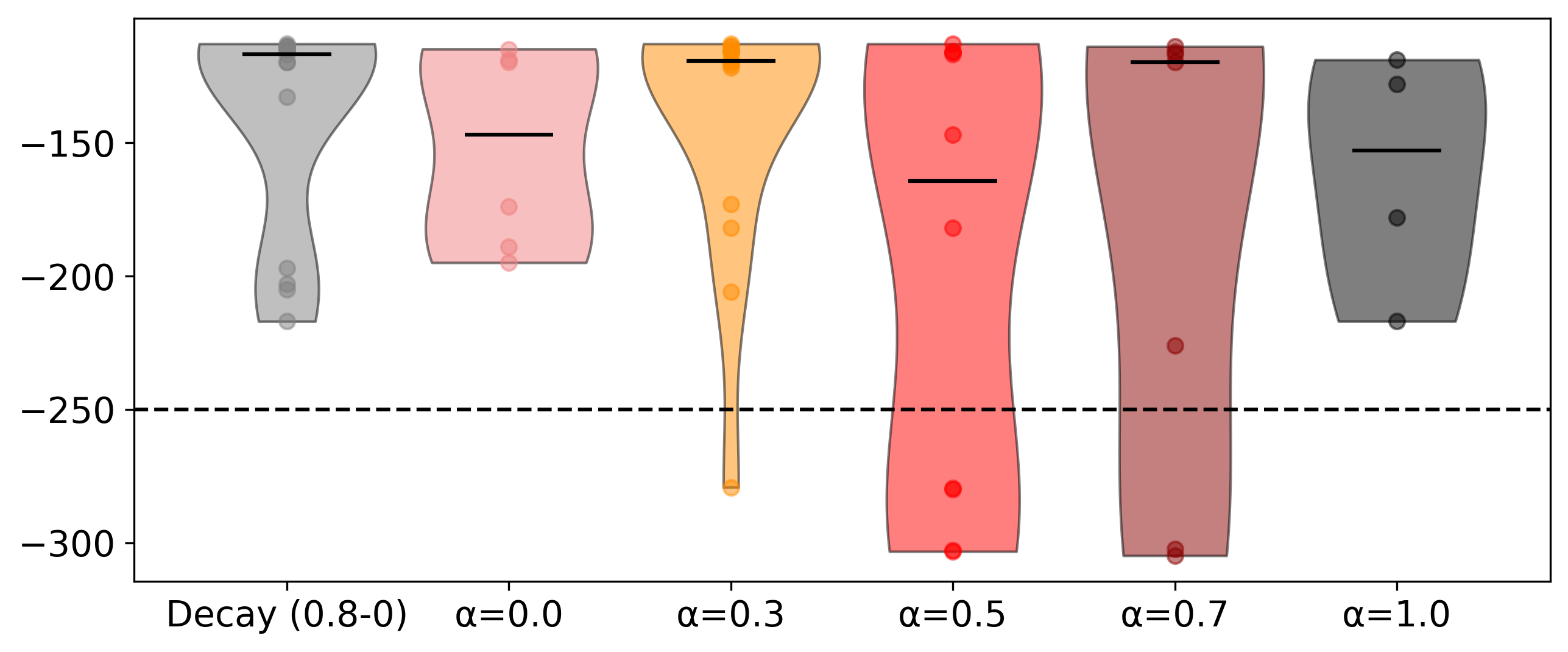}              &
		\includegraphics[width=\figwidth]{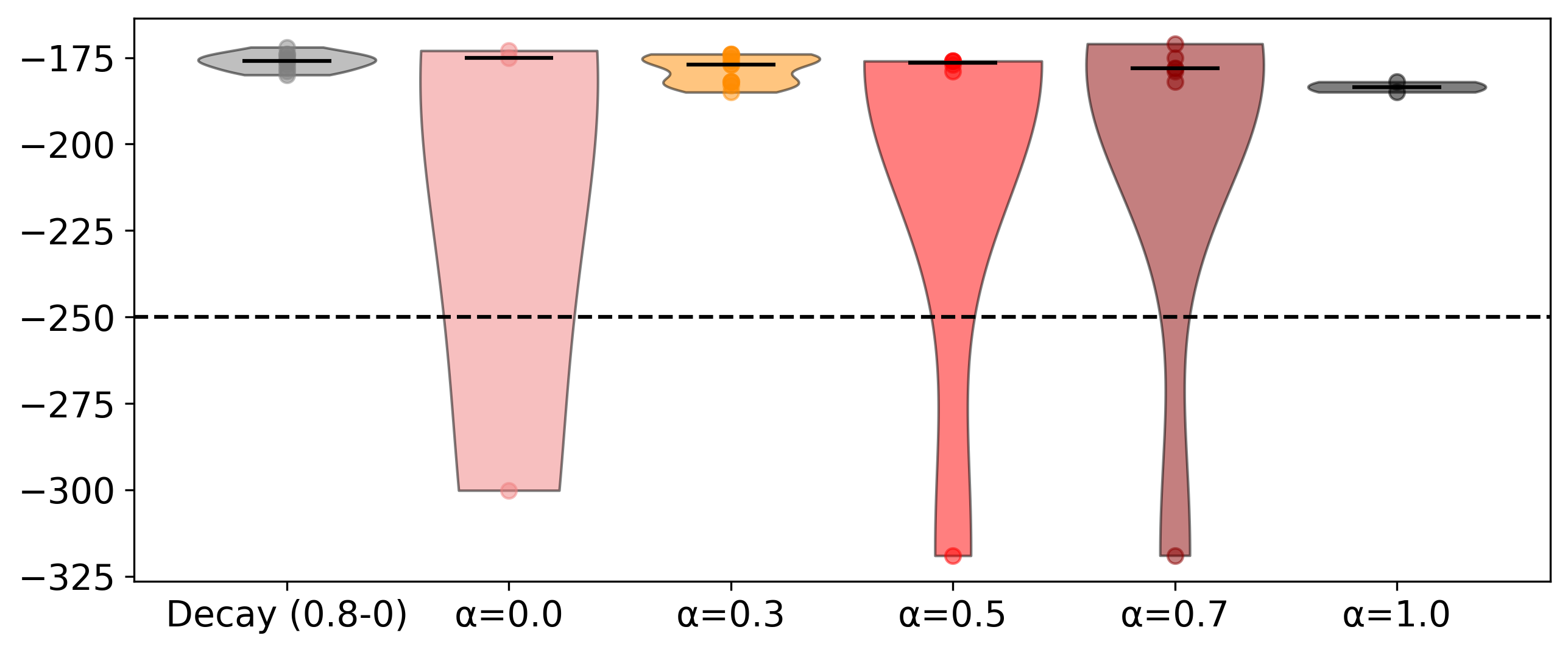}                  &
		\includegraphics[width=\figwidth]{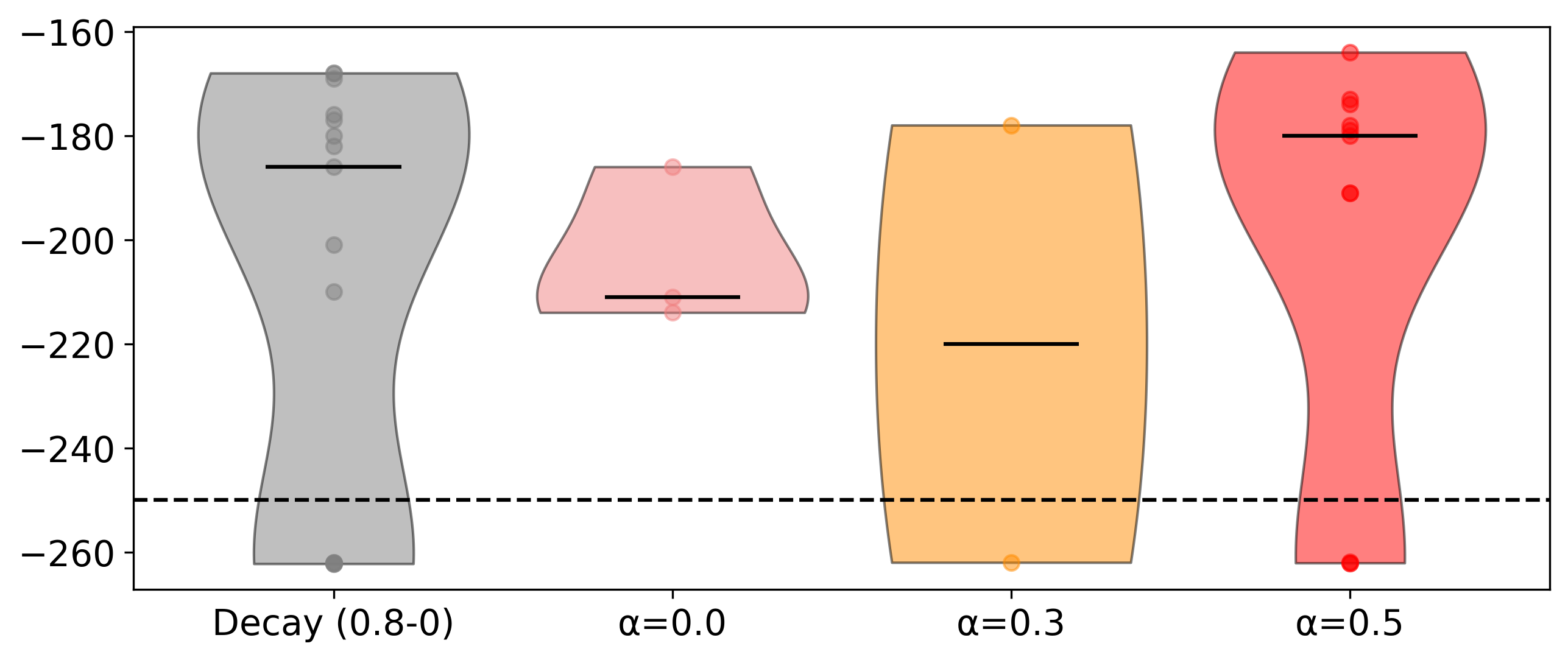}                                                            \\

  \\
		                                                                                               & \halfcheetah{} & \walker{}   & \antmaze{}  \\
		\raisebox{3\normalbaselineskip}[1cm][0cm]{\rotatebox[origin=c]{90}{\vspace{1cm}\yleg{}}}       &
		\includegraphics[width=\figwidth]{plots/jedi_halfcheetah_uni-500-D.png}                        &
		\includegraphics[width=\figwidth]{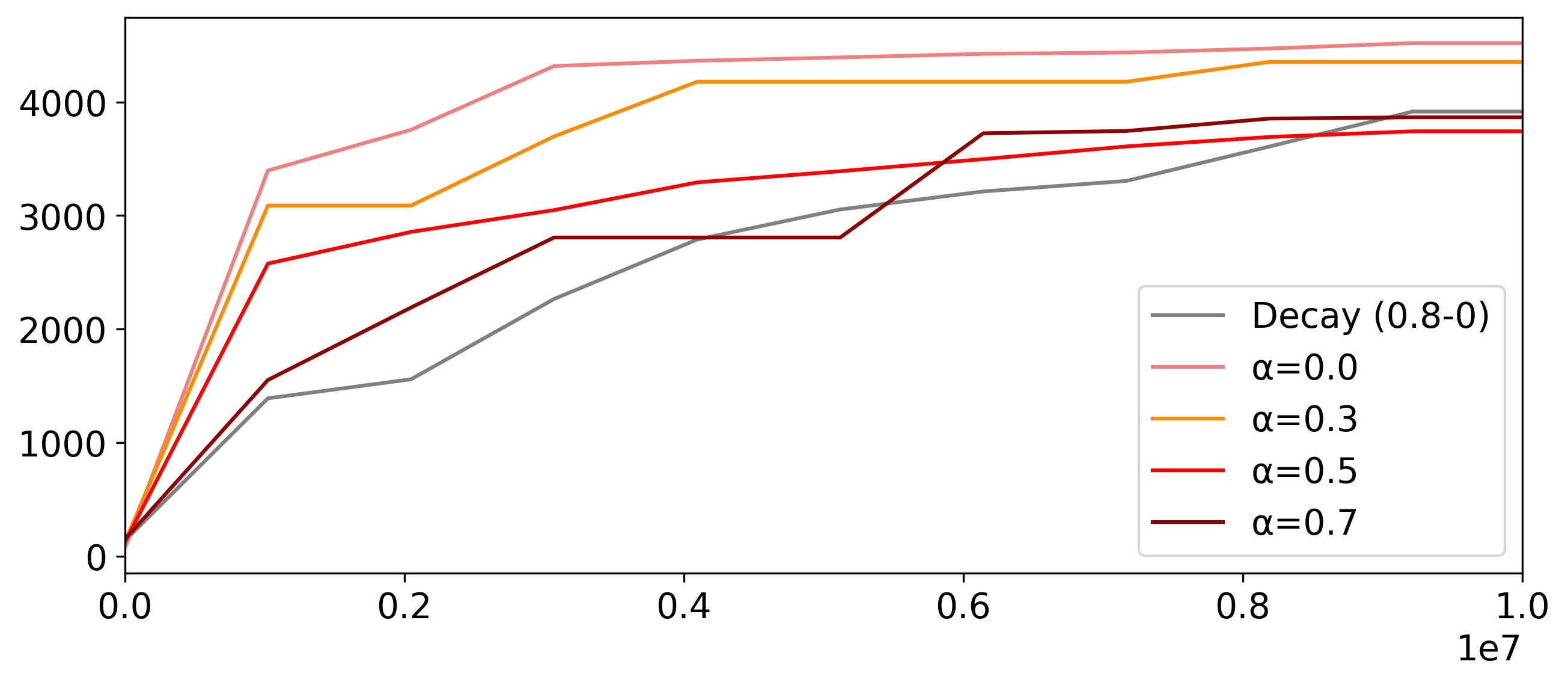}                          &
		\includegraphics[width=\figwidth]{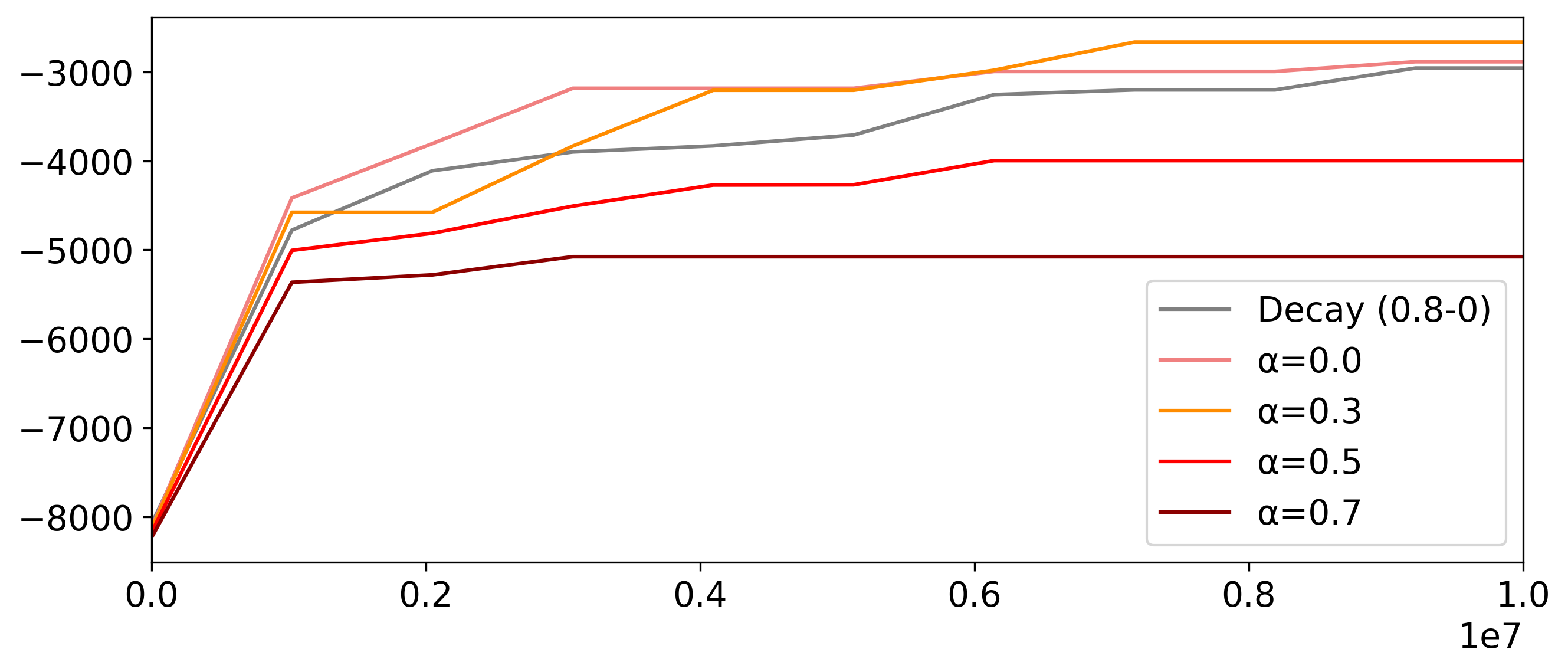}                                                                             \\

		\raisebox{3\normalbaselineskip}[1cm][0cm]{\rotatebox[origin=c]{90}{\vspace{1cm}Coverage (\%)}} &
		\includegraphics[width=\figwidth]{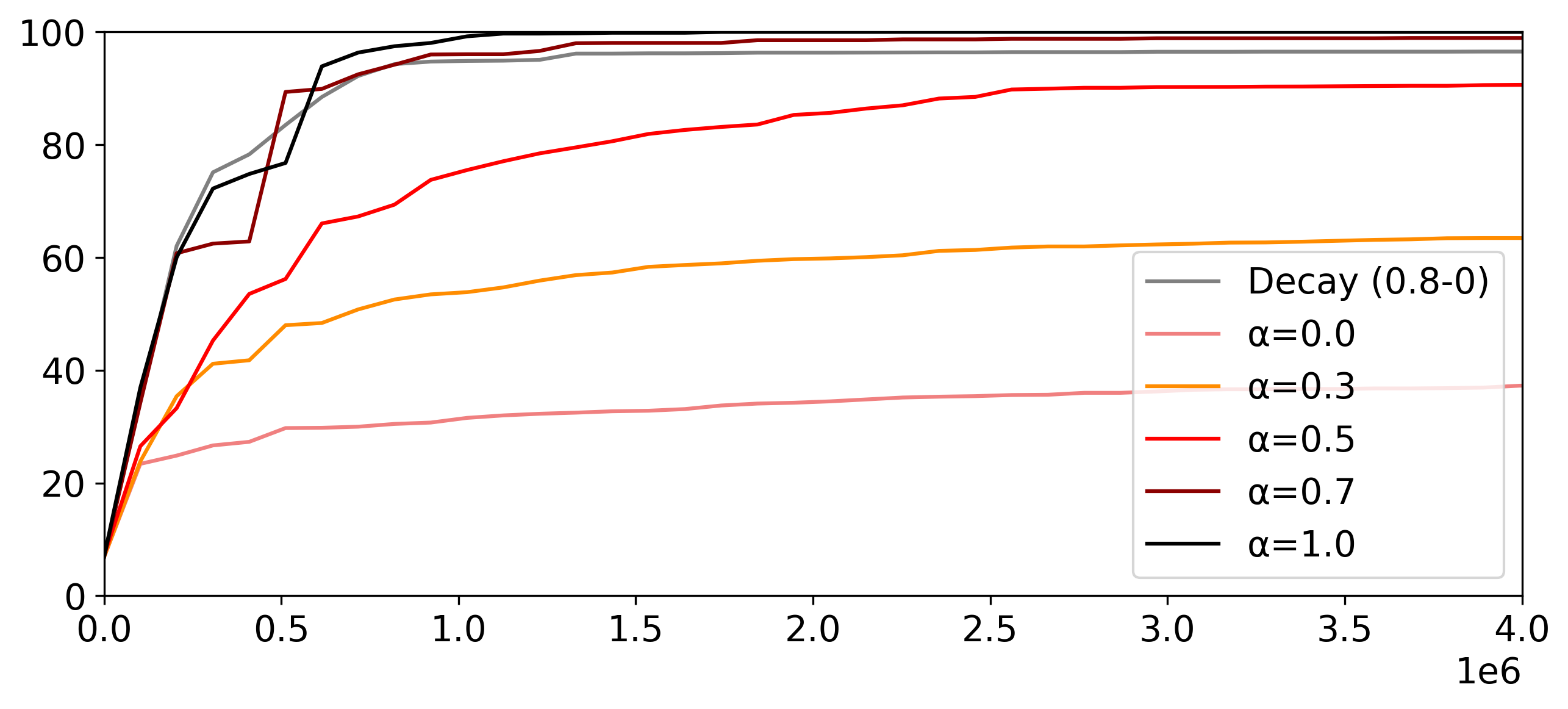}               &
		\includegraphics[width=\figwidth]{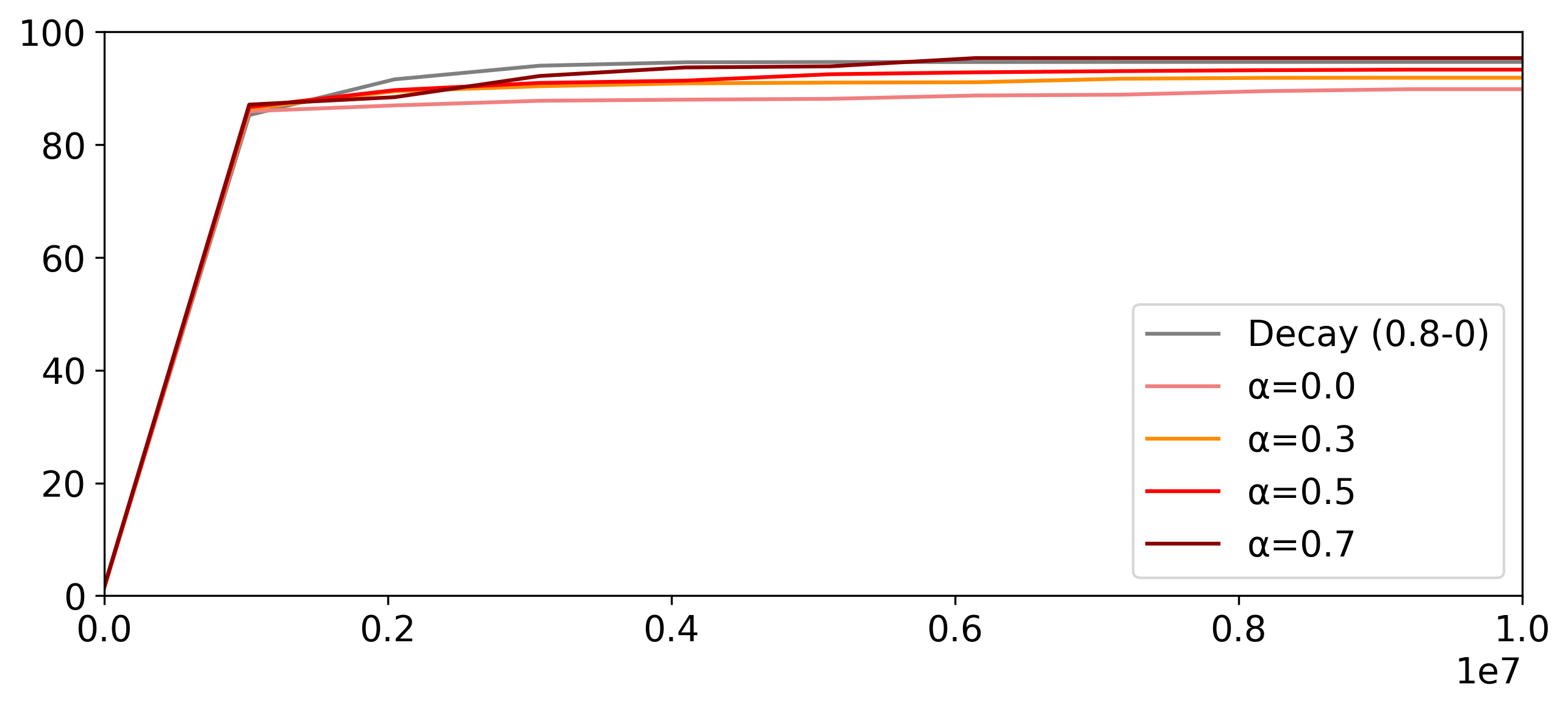}                 &
		\includegraphics[width=\figwidth]{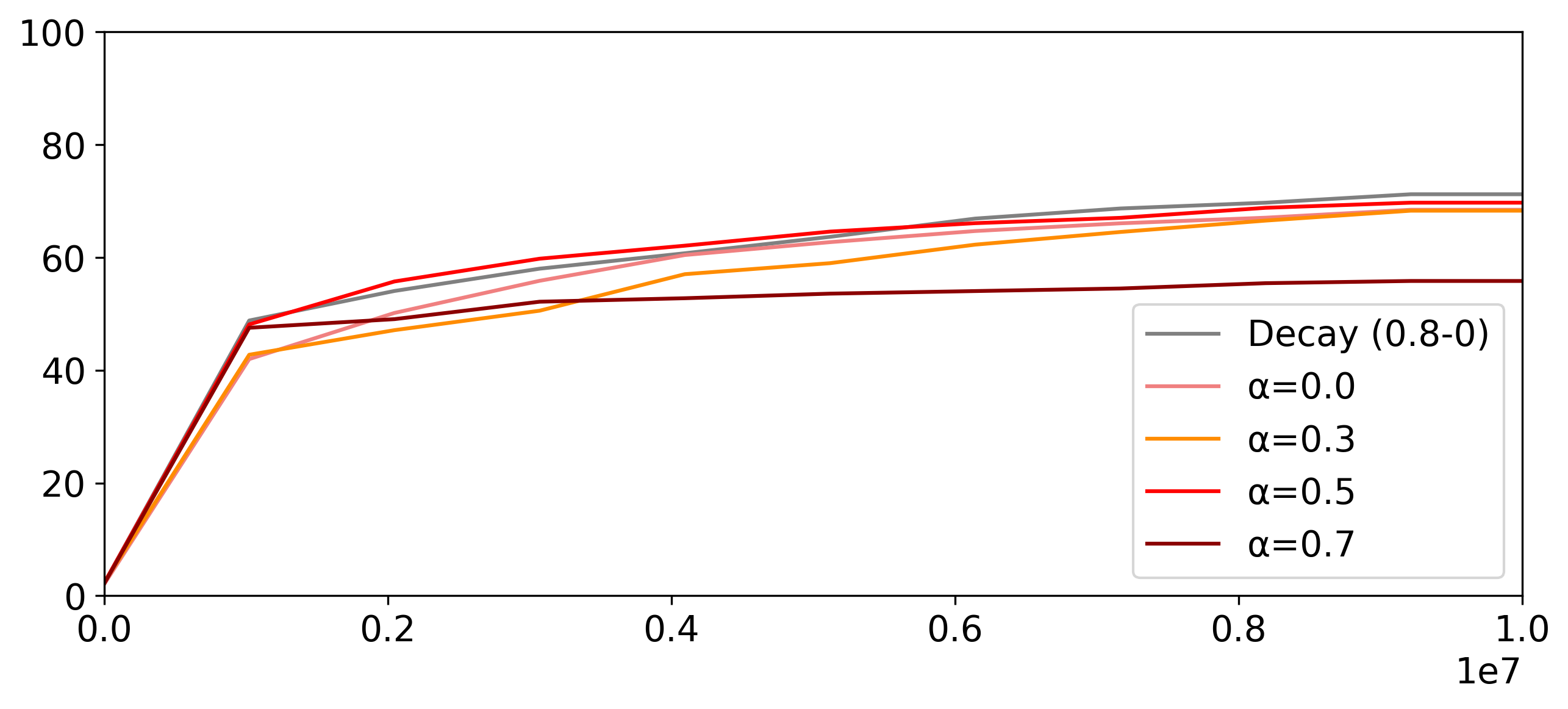}                                                                    \\
  		\raisebox{3\normalbaselineskip}[1cm][0cm]{\rotatebox[origin=c]{90}{\vspace{1cm}Final fitness}} &
		\includegraphics[width=\figwidth]{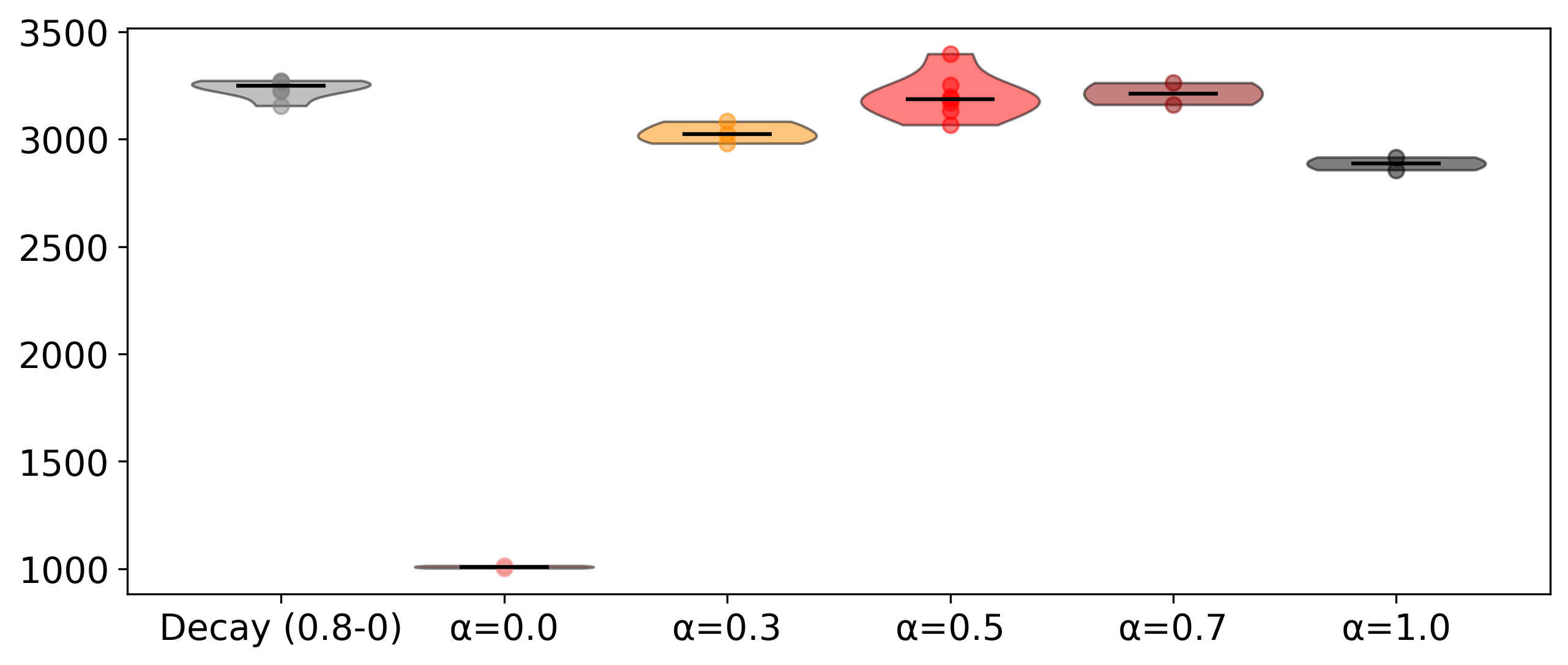}           &
		\includegraphics[width=\figwidth]{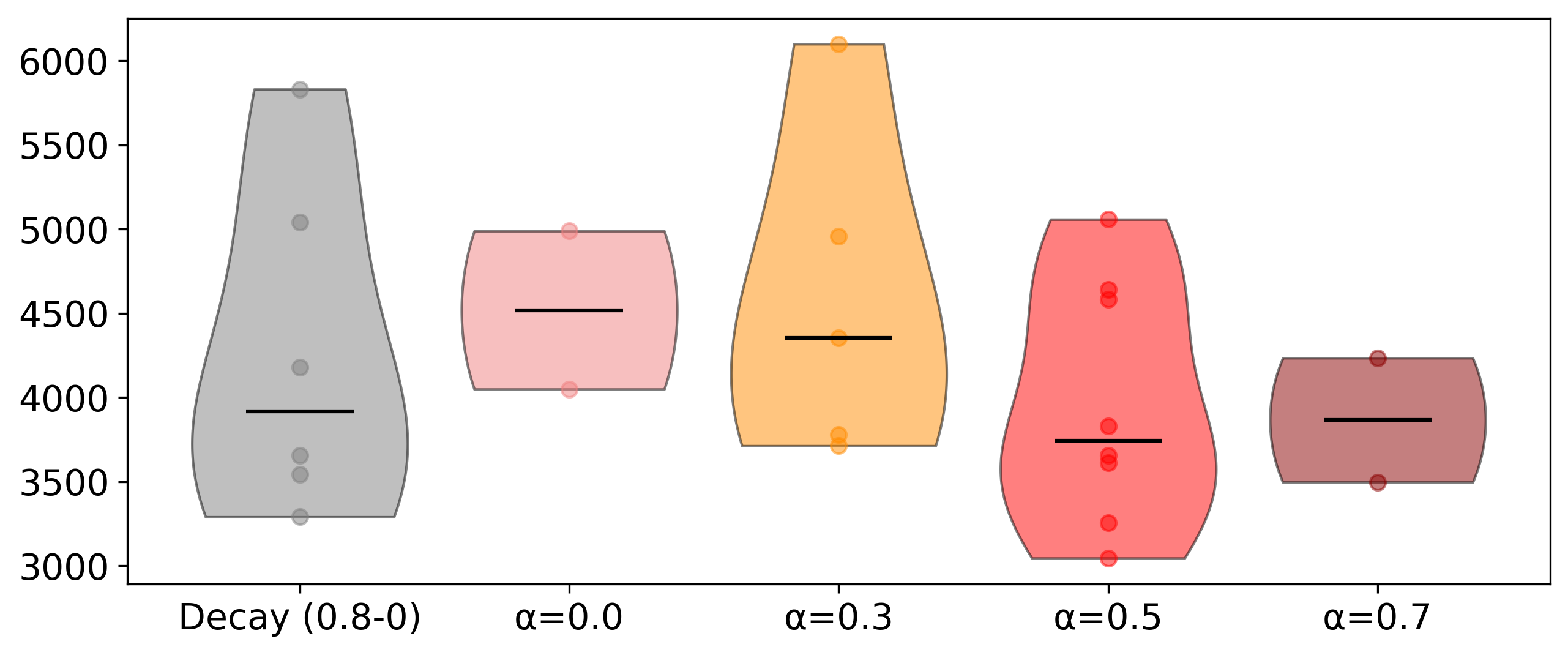}             &
		\includegraphics[width=\figwidth]{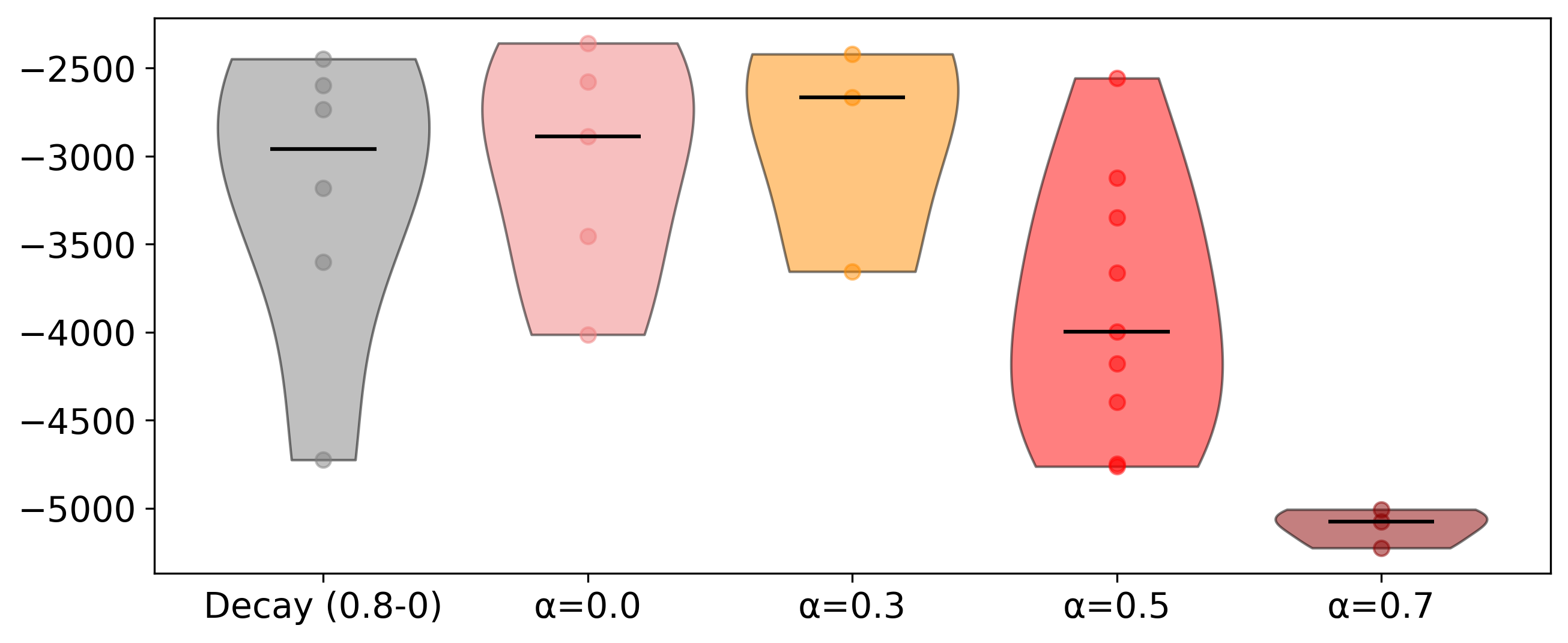}                                                                \\
		                                                                                               & \xleg          & \xleg       & \xleg
	\end{tabular}
	\captionsetup{type=figure}
	\caption{Ablation on \jedi{} with different \wtf{} weights schedules. Figures show median max fitness (row 1), coverage (row 2) and final fitness distribution (row 3). }
	\label{app:plot:jedi_alpha}
\end{table*}

\def\figwidth{0.3\linewidth}
\def\xleg{Evaluations}
\def\yleg{Max fitness}
\begin{table*}[t]
	\centering
	\begin{tabular}{cccc}
		                                                                                         & Maze A         & Maze B      & Maze C      \\
		\raisebox{3\normalbaselineskip}[1cm][0cm]{\rotatebox[origin=c]{90}{\vspace{1cm}\yleg{}}} &
		\includegraphics[width=\figwidth]{plots/weighted_KH-pointmaze-250-D.png}                 &
		\includegraphics[width=\figwidth]{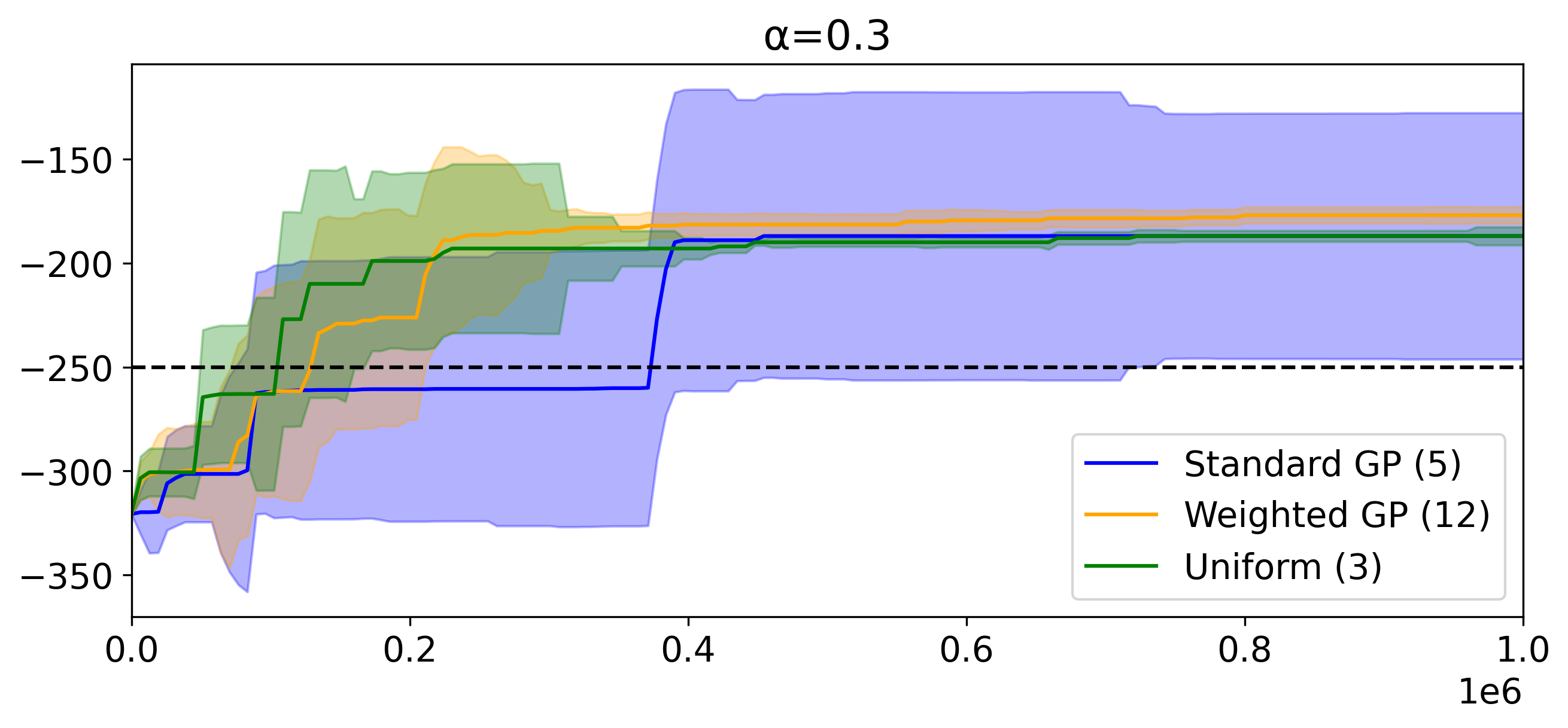}                     &
		\includegraphics[width=\figwidth]{plots/weighted_KH-standard-250-D.png}                                                               \\

		                                                                                         & \halfcheetah{} & \walker{}   & \antmaze{}  \\
		\raisebox{3\normalbaselineskip}[1cm][0cm]{\rotatebox[origin=c]{90}{\vspace{1cm}\yleg{}}} &
		\includegraphics[width=\figwidth]{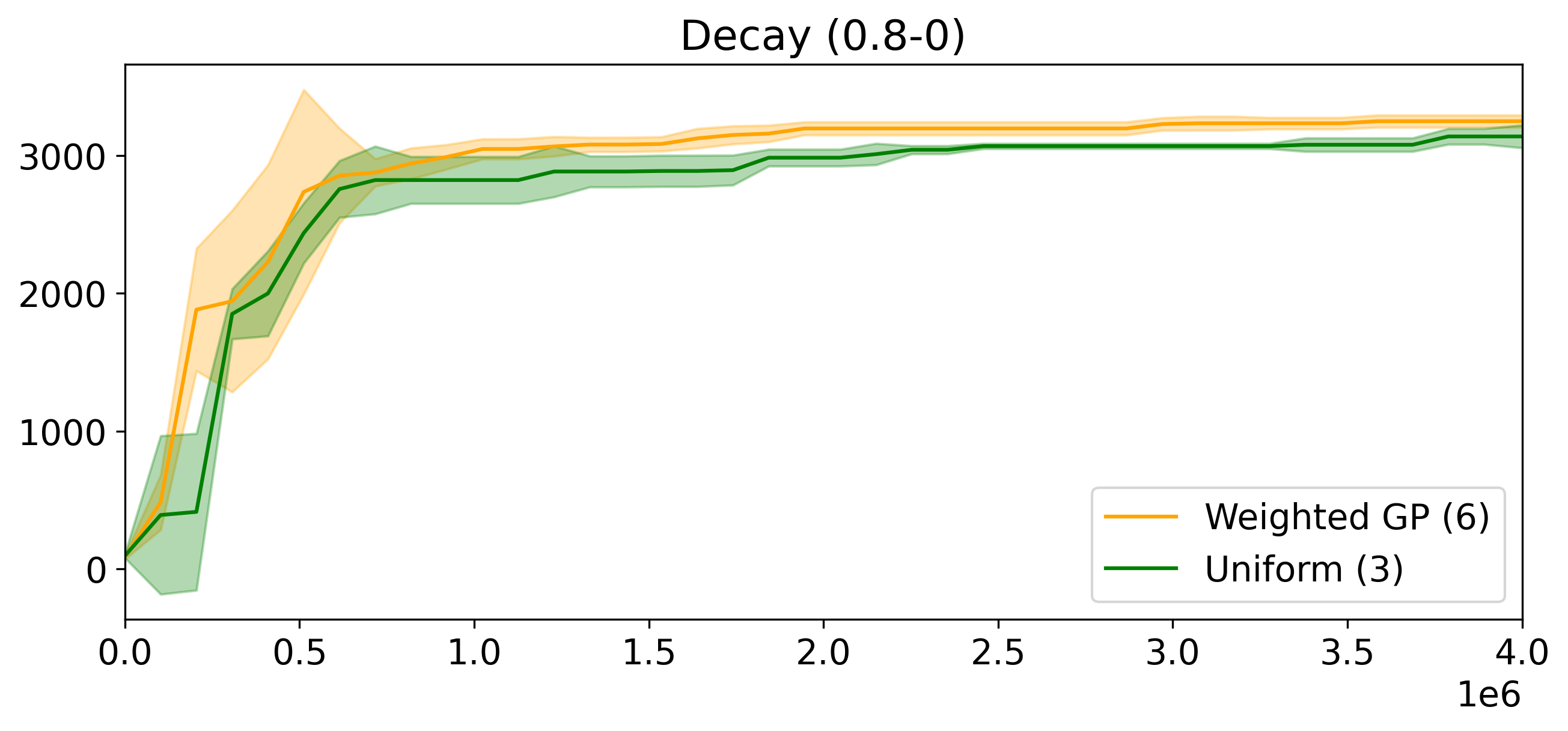}              &
		\includegraphics[width=\figwidth]{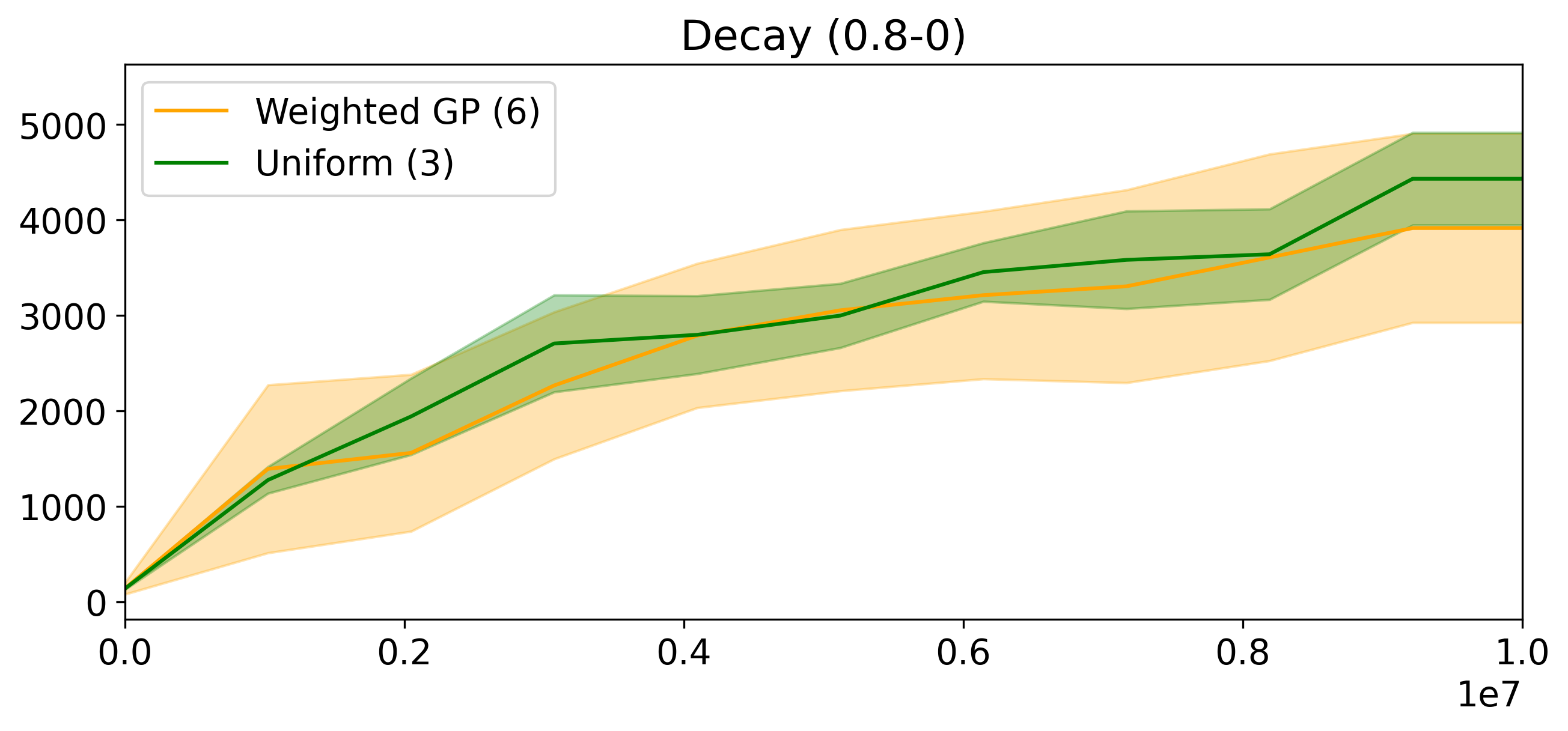}                &
		\includegraphics[width=\figwidth]{plots/weighted_antmaze-250-D.png}                                                                   \\

		                                                                                         & \xleg          & \xleg       & \xleg
	\end{tabular}
	\captionsetup{type=figure}
	\caption{Median fitness results for \jedi{} with standard Gaussian Process, weighted GP and uniform sampling. Reaching the dotted line at -250 fitness means an agent has reached the target.}
	\label{app:plot:jedi_wgp}
\end{table*}

\def\figwidth{0.3\linewidth}
\def\xleg{Evaluations}
\def\yleg{Max fitness}
\begin{table*}[t]
	\centering
	\begin{tabular}{cccc}
		                                                                                         & Maze A         & Maze B      & Maze C      \\
		\raisebox{5\normalbaselineskip}[1cm][0cm]{\rotatebox[origin=c]{90}{\vspace{1cm}\yleg{}}} &
		\includegraphics[width=\figwidth]{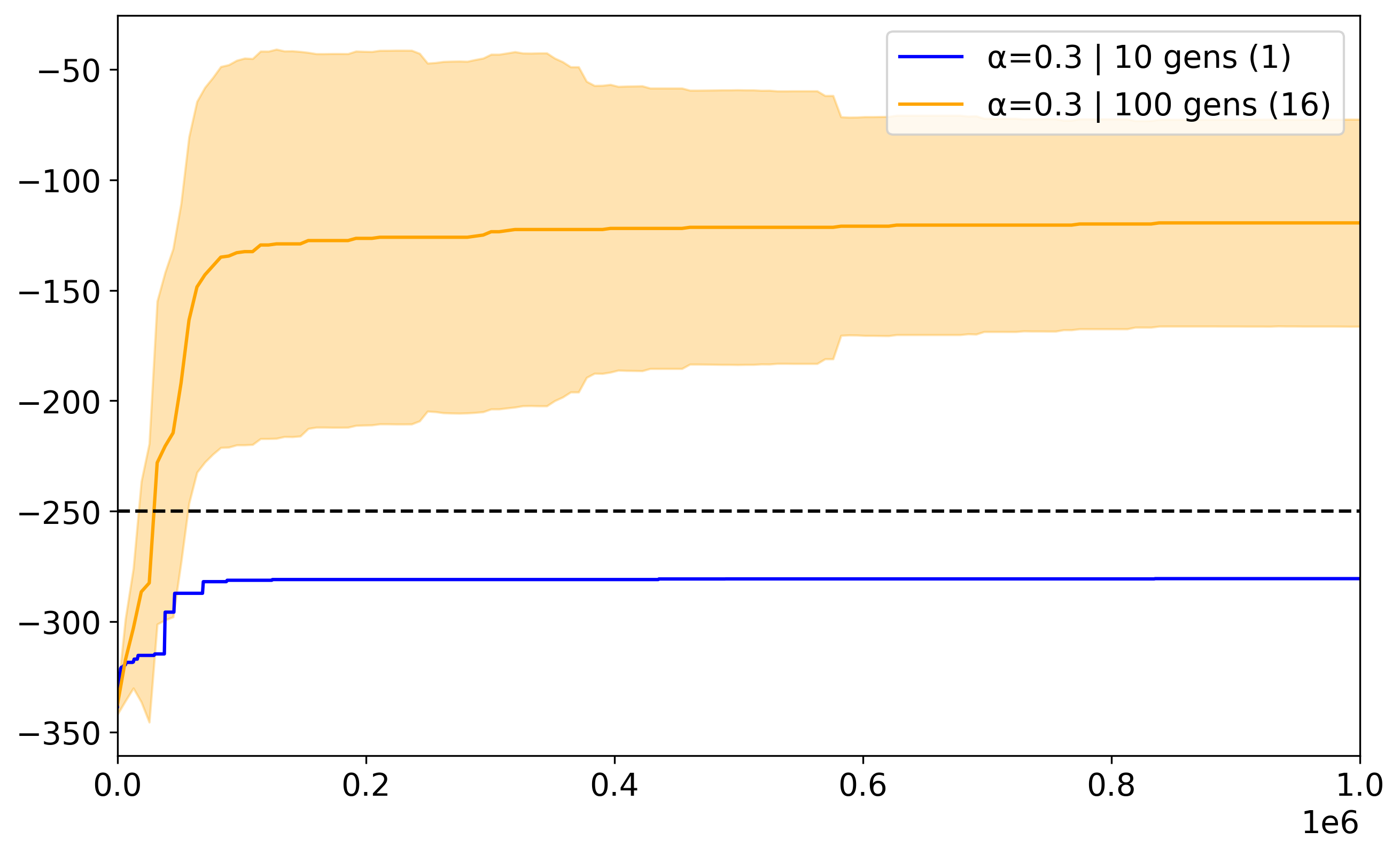}                  &
		\includegraphics[width=\figwidth]{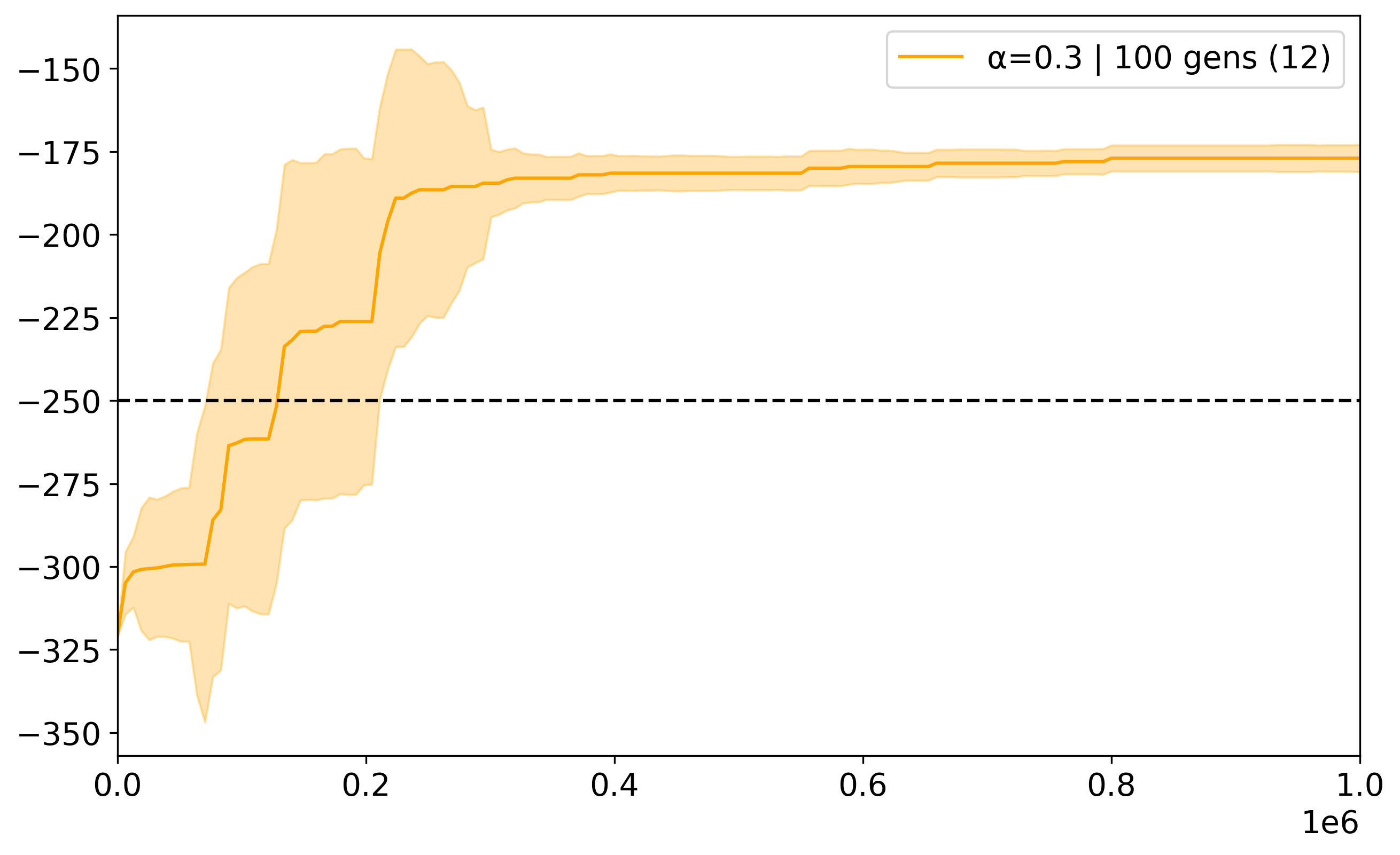}                      &
		\includegraphics[width=\figwidth]{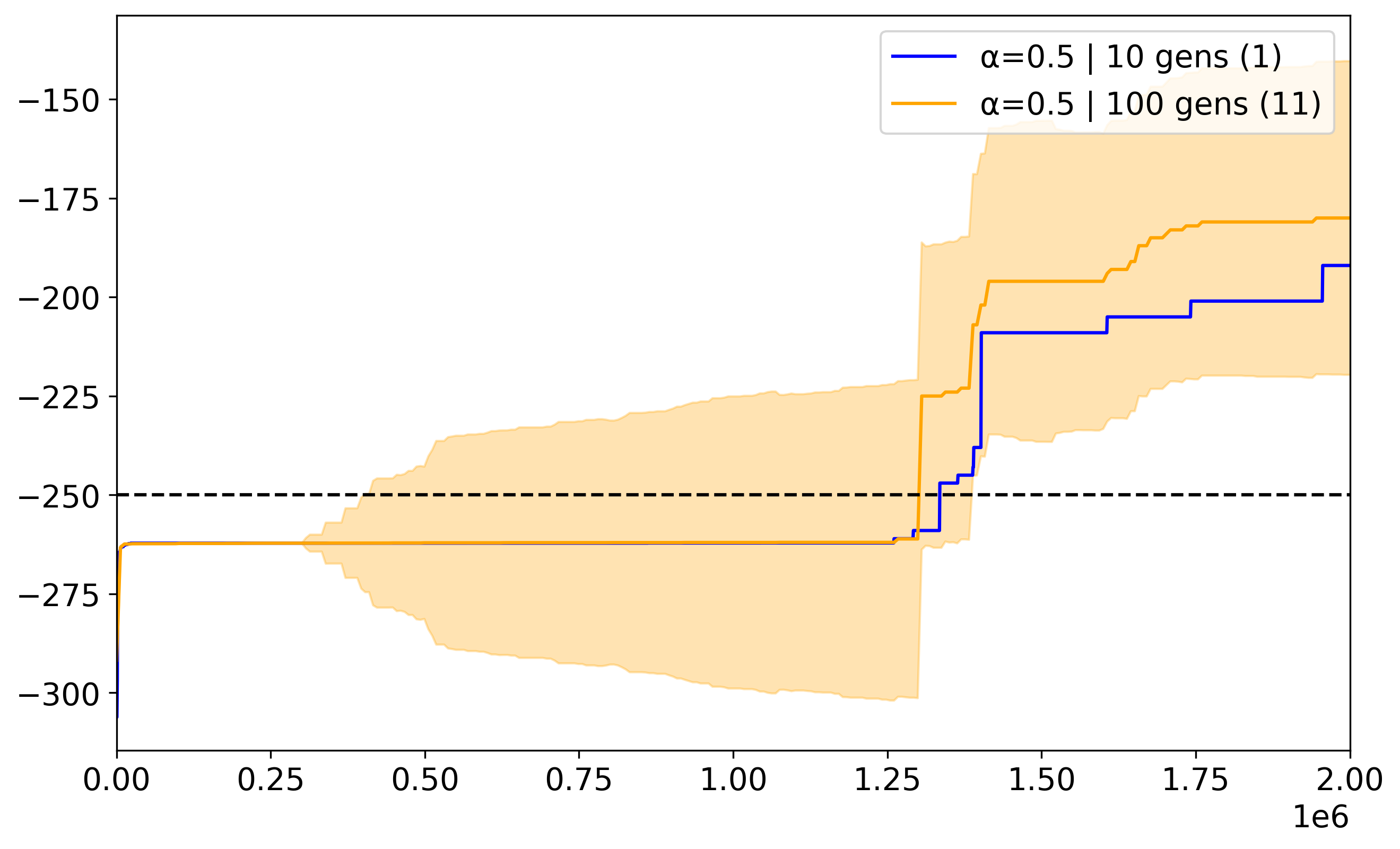}                                                                \\

		                                                                                         & \xleg          & \xleg       & \xleg
	\end{tabular}
	\captionsetup{type=figure}
	\caption{Fitness results for maze exploration for \jedi{} with different \jedi{} ES generations. Reaching the dotted line at -250 fitness means an agent has reached the target.}
	\label{app:plot:jedi_gens}
\end{table*}

\end{document}